\documentclass[sn-mathphys,Numbered]{sn-jnl}

\usepackage{graphicx}%
\usepackage{subfigure}
\usepackage{multirow}%
\usepackage{amsmath,amssymb,amsfonts}%
\usepackage{amsthm}%
\usepackage{mathrsfs}%
\usepackage[title]{appendix}%
\usepackage{xcolor}%
\usepackage{textcomp}%
\usepackage{manyfoot}%
\usepackage{booktabs}%
\usepackage{algorithm}%
\usepackage{algorithmicx}%
\usepackage{algpseudocode}%
\usepackage{listings}%
\usepackage{makecell}

\theoremstyle{thmstyleone}%
\theoremstyle{thmstyletwo}%

\theoremstyle{thmstylethree}%

\begin{document}

\title[Article Title]{Affective Video Content Analysis: Decade Review and New Perspectives}


\author[1,2]{\fnm{Junxiao} \sur{Xue}}\email{xuejx@zhejianglab.com}

\author*[2]{\fnm{Jie} \sur{Wang}}\email{w202112332015479@gs.zzu.edu.cn}

\author[3]{\fnm{Xuecheng} \sur{Wu}}\email{wuxc3@stu.xjtu.edu.cn}

\author[2]{\fnm{Qian} \sur{Zhang}}

\affil*[1]{\orgdiv{Research Institute of Artificial Intelligence}, \orgname{Zhejiang Lab}, \orgaddress{\city{Hanghzou}, \postcode{310000}, \state{Zhejiang}, \country{China}}}

\affil[2]{\orgdiv{School of Cyber Science and Engineering}, \orgname{Zhengzhou University}, \orgaddress{\city{Zhengzhou}, \postcode{450002}, \state{Henan}, \country{China}}}

\affil[3]{\orgdiv{School of Computer Science and Technology}, \orgname{Xi'an Jiaotong University},
\orgaddress{\city{Xi'an}, \postcode{710049}, \state{Shaanxi}, \country{China}}}


\abstract{Video content is rich in semantics and has the ability to evoke various emotions in viewers. In recent years, with the rapid development of affective computing and the explosive growth of visual data, affective video content analysis (AVCA) as an essential branch of affective computing has become a widely researched topic. In this study, we comprehensively review the development of AVCA over the past decade, particularly focusing on the most advanced methods adopted to address the three major challenges of video feature extraction, expression subjectivity, and multimodal feature fusion. We first introduce the widely used emotion representation models in AVCA and describe commonly used datasets. We summarize and compare representative methods in the following aspects: (1) unimodal AVCA models, including facial expression recognition and posture emotion recognition; (2) multimodal AVCA models, including feature fusion, decision fusion, and attention-based multimodal models; (3) model performance evaluation standards. Finally, we discuss future challenges and promising research directions, such as emotion recognition and public opinion analysis, human-computer interaction, and emotional intelligence.}

\keywords{Affective computing, Video emotion, Video feature extraction, Machine learning, Emotional intelligence}



\maketitle

\section{Introduction}\label{intro}

The ability of machines to possess emotions is a crucial feature of intelligent machines, and addressing emotional issues is a prerequisite for machines to have emotions. Despite the significant role emotions play in machine and artificial intelligence, much less attention has been given to affective computing than objective semantic understanding, such as object classification in computer vision. The rapid development of artificial intelligence has achieved significant advancements in semantic understanding, which has placed higher demands on emotional interaction. For example, a mobile assistant capable of recognizing and expressing emotions could provide services that better meet users' needs, especially the visually impaired. In order to achieve human-like emotions, machines should first understand how humans express emotions through various channels, such as voice, facial expressions, body posture, and physiological signals. Although physiological signals, unaffected by human will, can provide more reliable information, capturing accurate physiological signals is quite challenging and requires special wearable sensors. On the other hand, the widespread popularity of short video social platforms (such as TikTok, Instagram, and Douyin) allows users to share their daily lives and express their opinions through audio and video. Recognizing the emotional content in this vast amount of video data provides a means to understand users' behavior and emotions.

 As we know, "a picture is worth a thousand words", which emphasizes the rich information an image can convey. A video is composed of a sequence of continuous images, making it capable of conveying even richer information than a single image. In addition to images, videos also include audio signals, and sound is the most commonly used way for humans to express emotions \cite{tajadura2008embodied, asutay2012emoacoustics, globerson2013psychoacoustic}. In contrast to existing research, analyzing video data faces the challenge of extracting video features, making effective video architecture a key focus of Affective Video Content Analysis (AVCA). Compared to objective semantic understanding, emotion recognition focuses on a higher level - the cognitive level, i.e., understanding how videos can evoke emotions in viewers, which is more challenging. Using AVCA to assess human emotional states automatically can help evaluate users' mental health, detect emotional abnormalities, and assist recommendation algorithms in recommending suitable videos to adjust emotions, thereby preventing the deepening of mental issues for users or even the occurrence of extreme behavior in society as a whole \cite{gross2002emotion, fischer2016social}.

\begin{figure}[t]
    \centering
    \subfigure[sadness]{
        \begin{minipage}{0.22\linewidth}
            \centering
            \includegraphics[width=\linewidth]{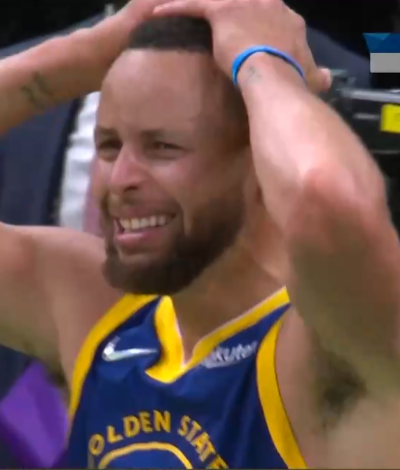}
            \label{fig1a}
        \end{minipage}
    }
    \subfigure[sadness]{
        \begin{minipage}{0.22\linewidth}
            \centering
            \includegraphics[width=\linewidth]{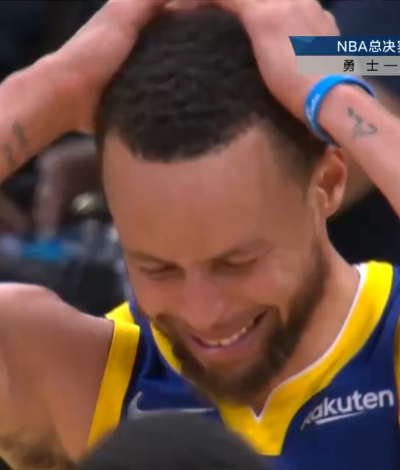}
            \label{fig1b}
        \end{minipage}
    }
    \subfigure[neutral]{
        \begin{minipage}{0.22\linewidth}
            \centering
            \includegraphics[width=\linewidth]{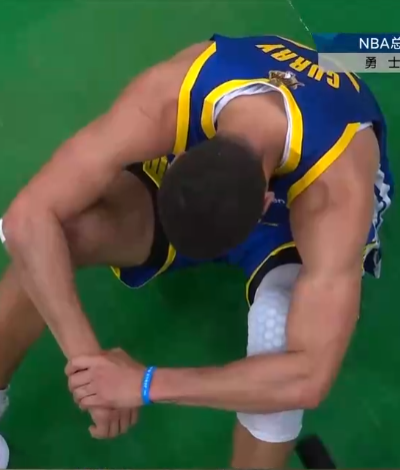}
            \label{fig1c}
        \end{minipage}
    }
    \subfigure[happiness]{
        \begin{minipage}{0.22\linewidth}
            \centering
            \includegraphics[width=\linewidth]{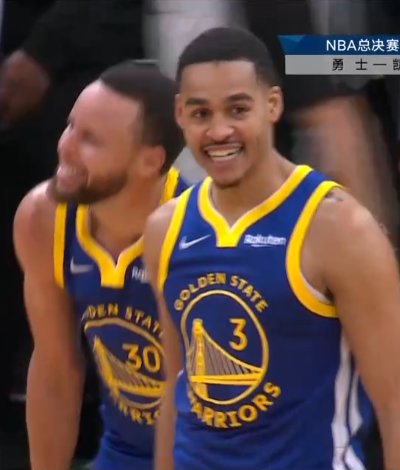}
            \label{fig1d}
        \end{minipage}
    }
    \caption{The temporal nature plays an important role in AVCA. These images are captured from a 5-second consecutive segment from the same video, showcasing Curry's emotions after winning the NBA championship once again. In images (a) and (d), Curry expresses different emotions (sadness vs. happiness).}
    \label{fig1}
\end{figure}

\subsection{Main Goals and Challenges}\label{goal}

\textbf{Main Goals.} Given an input video, AVCA mainly aims to (1) identify the emotions the video creator intends to convey (Based psychology, the emotions can be represented using different models, such as categorical or dimensional models. For further details, please refer to Section \ref{modemo}.), (2) analyze the stimuli within the video that trigger these emotions, such as specific language, actions, or combinations of expressions, and (3) apply the identified emotions to various real-world contexts to enhance emotional intelligence.

\noindent\textbf{Challenges. (1) Video Feature Extraction.} Compared to static image data, video data possesses more complex characteristics. The temporal nature of video provides a richer form of emotional expression. For instance, in Fig. \ref{fig1}, the person in Fig. \ref{fig1a} displays a sense of sadness, while as the video progresses, we can observe the person in Fig. \ref{fig1d} expressing overwhelming happiness. Relying solely on individual frames, it is difficult to accurately discern the emotions conveyed by the creator, whereas the temporal nature of video effectively addresses this issue. Furthermore, the temporal nature of video also presents new challenges for feature extraction.

The method of taking the average of the feature vectors of all frames in a video, although based on the intuitive solution of image convolutional neural network (CNN), cannot effectively utilize the feature information of the video. In contrast, NetVLAD \cite{arandjelovic2016netvlad} uses the clustering concept to obtain multiple cluster centers by clustering the frame features of the video. Then concat the average of the feature vectors in each cluster area to form the feature vector of the entire video. This method effectively utilizes the video's feature information and performs excellently in video feature extraction. In response to the temporal information of the video, researchers have utilized recurrent neural networks (RNN) for feature extraction. RNN can effectively extract feature information from temporal data, with Long Short Term Memory (LSTM) being widely applied in video feature extraction due to its ability to capture long-term dependencies. It has been applied in various domains, such as rainfall text emotion recognition \cite{glenn2023emotion}, video summarization \cite{zhao2018hsa}, and action recognition \cite{donahue2015long}. Expanding two-dimensional convolutions to obtain three-dimensional convolutional kernels suitable for video data is the most natural solution. João et al. \cite{carreira2017quo} expanded the pre-trained weights of two-dimensional convolutions to 3D convolutions and incorporated optical flow modal information to address the issue of motion information loss during frame extraction. To date, the use of three-dimensional convolutional kernels in the design of models for video feature extraction is considered the optimal solution, with the extracted features demonstrating significantly higher effectiveness than other methods.

\begin{figure}[t]
    \centering
    \subfigure[Within images]{
        \begin{minipage}{0.48\linewidth}
            \centering
            \includegraphics[width=\linewidth]{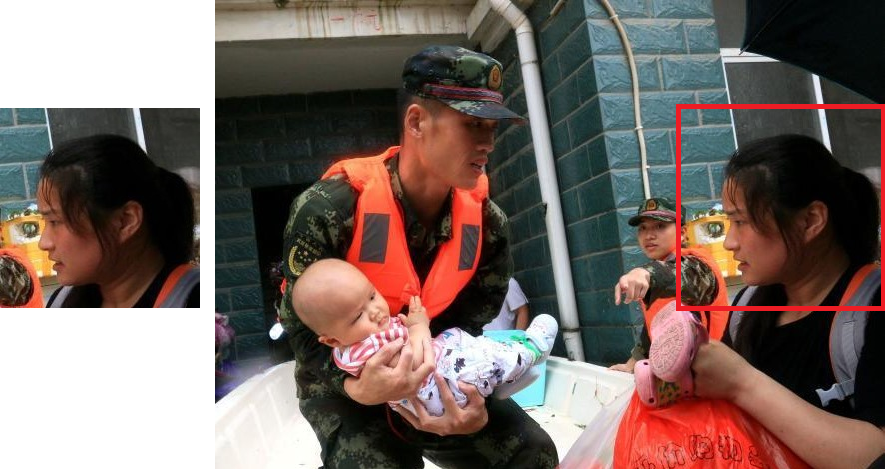}
            \label{fig2a}
        \end{minipage}
    }
    \subfigure[Within voice]{
        \begin{minipage}{0.46\linewidth}
            \centering
            \includegraphics[width=\linewidth]{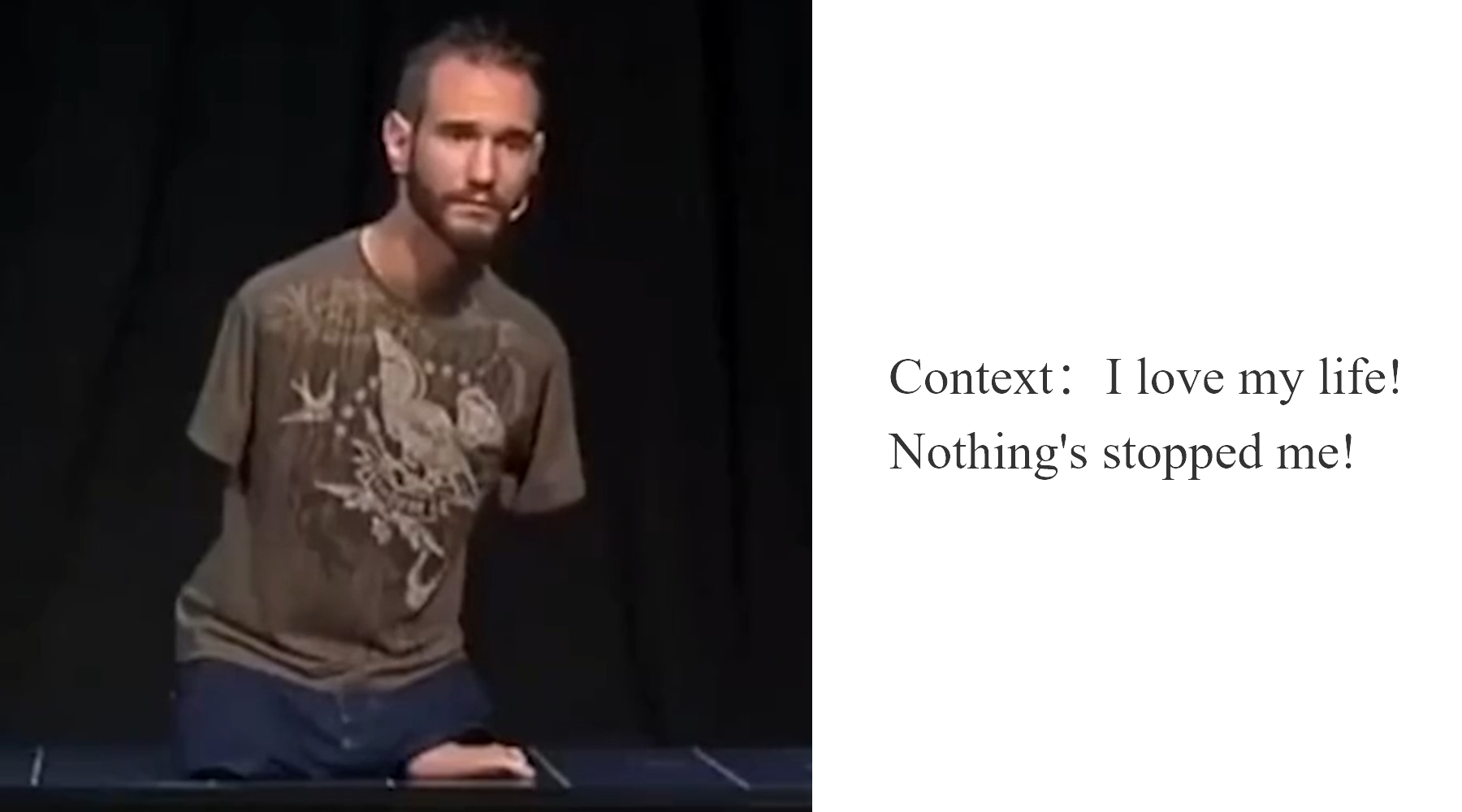}
            \label{fig2b}
        \end{minipage}
    }
    \caption{The contextual information also proves helpful for the AVCA task. (a) The image without and with the detailed scene context expresses different emotions (fear vs. surprise). (b) The voice can also infulence the emtion perception of the same video (sadness vs. positive).}
    \label{fig2}
\end{figure}

\begin{figure}[t]
    \centering
    \subfigure[Thumbs up]{
        \begin{minipage}{0.56\linewidth}
            \centering
            \includegraphics[width=\linewidth]{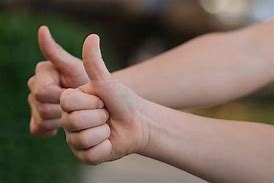}
            \label{fig3a}
        \end{minipage}
    }
    \subfigure[Nod]{
        \begin{minipage}{0.38\linewidth}
            \centering
            \includegraphics[width=\linewidth]{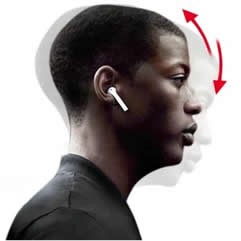}
            \label{fig3b}
        \end{minipage}
    }
    \caption{Illustration of the expression subjectivity. The same gestures or actions may elicit significantly different emotional expressions from content creators in different cultures. In most countries, people may express agreement and approval with images (a) and (b), individuals in the Mediterranean region might convey disdain and negation instead.}
    \label{fig3}
\end{figure}

In addition to video features, incorporating available contextual information also proves helpful for the AVCA task. As illustrated in Fig. \ref{fig2}, the same video may evoke different emotions under different narratives and voice-overs. For instance, in Fig. \ref{fig2a}, if we only see the mother, we might infer fear based on her expression; however, witnessing the firefighter rescuing a child from a flood is more likely to lead to an inference of surprise. In Fig. \ref{fig2b}, upon seeing the disabled person, we might infer profound sadness; yet upon hearing his words, "I love my life! Nothing's stopped me!", we are more likely to infer an extremely positive outlook.

\noindent\textbf{(2) Expression Subjectivity.} As a result of the differing personal backgrounds of users, including cultural, social, and individual personality factors \cite{shaver1992cross, fischer2003social}, the same emotion may be expressed in completely opposite ways. For example, the gesture in Fig. \ref{fig3a} is commonly used to express affirmation and support in most countries; however, in some countries, this gesture is used to convey offensive behavior, such as in Iran, Greece, and the island of Sardinia in Italy. Additionally, other actions may have different meanings in different regions, such as nodding indicating disagreement in the Mediterranean region. This fact leads to expression subjectivity issues, thus requiring accurate annotation of emotional content by considering the overall background of the content creator. These background differences may impact the recognition of emotions and the interpretation of emotional content, and it is essential to consider these background factors when conducting emotional analysis and annotating emotional content.

To address the subjectivity of emotional expression, it is crucial to accurately consider the cultural and social backgrounds of content creators during the dataset construction process to achieve precise data annotation. Accurately describing the emotions expressed by users is the primary concern in building the dataset. In psychological research, different models are used to represent emotions, primarily categorized as categorical emotion states (CES) \cite{plutchik1982psychoevolutionary, ekman1992argument, mikels2005emotional} and dimensional emotion space (DES) \cite{schlosberg1954three, lee2011fuzzy}. These methods of representing emotions have been widely adopted in other areas of affective computing research, and existing video emotion content analysis datasets also employ the same emotional representation methods.

\noindent\textbf{(3) Multimodal Feature Fusion.} Videos contain multimodal information such as images, audio, and text. Effectively utilizing the rich information in videos is an important method for improving model performance. In multimodal research, fusion methods are categorized based on different fusion times into feature-level fusion (early fusion), decision-level fusion (late fusion), and hybrid fusion. Feature-level fusion involves using multiple feature extractors to extract features from different modalities and then fusing these features into input features according to a fusion strategy before entering the classifier. This method can fully utilize the feature information from different modalities, but it is also important to consider how to effectively fuse different modality features while maintaining the integrity of the information. In contrast, decision-level fusion involves assigning a separate classifier to each feature extractor to obtain emotional decisions for different modalities and then fusing these decisions into a final decision based on a fusion strategy. This method can independently handle the features and decisions of different modalities, but it also requires careful consideration of how to sensibly fuse the decisions from multiple modalities to obtain the final result. Hybrid fusion combines the advantages of the previous two fusion methods. It involves fusing similar modalities at the feature level, such as images and optical flow, and fusing dissimilar modalities at the decision level, such as images and text. This fusion method simplifies the feature information of similar modalities and effectively utilizes the feature information from different modalities, thereby achieving better performance in multimodal emotion recognition.

Fusion strategy is indeed the key factor in determining the performance of multimodal models across the three fusion methods. Taking the average of all modality features or decisions is the most intuitive solution and has been widely used in many models. For example, I3D \cite{carreira2017quo} uses this decision fusion for the image and optical flow modalities and achieves excellent performance. However, while averaging is intuitive, it contradicts the logic of emotional decision-making. When dealing with information from different modalities, people assign different weights. For example, phrases such as "a picture is worth a thousand words," "seeing is believing," and "actions speak louder than words" indicate the differential weight assigned to different modalities. Therefore, in some studies \cite{cai2019feature, sahoo2016emotion}, manually designed multimodal fusion weights have resulted in superior performance. In other studies \cite{lian2021ctnet}, researchers have optimized fusion weights as model parameters during the training process, leading to even better performance and higher efficiency compared to manually designed forms.

\begin{figure}[t]
    \centering
    \subfigure[Without credibility]{
        \begin{minipage}{0.85\linewidth}
            \centering
            \includegraphics[width=\linewidth]{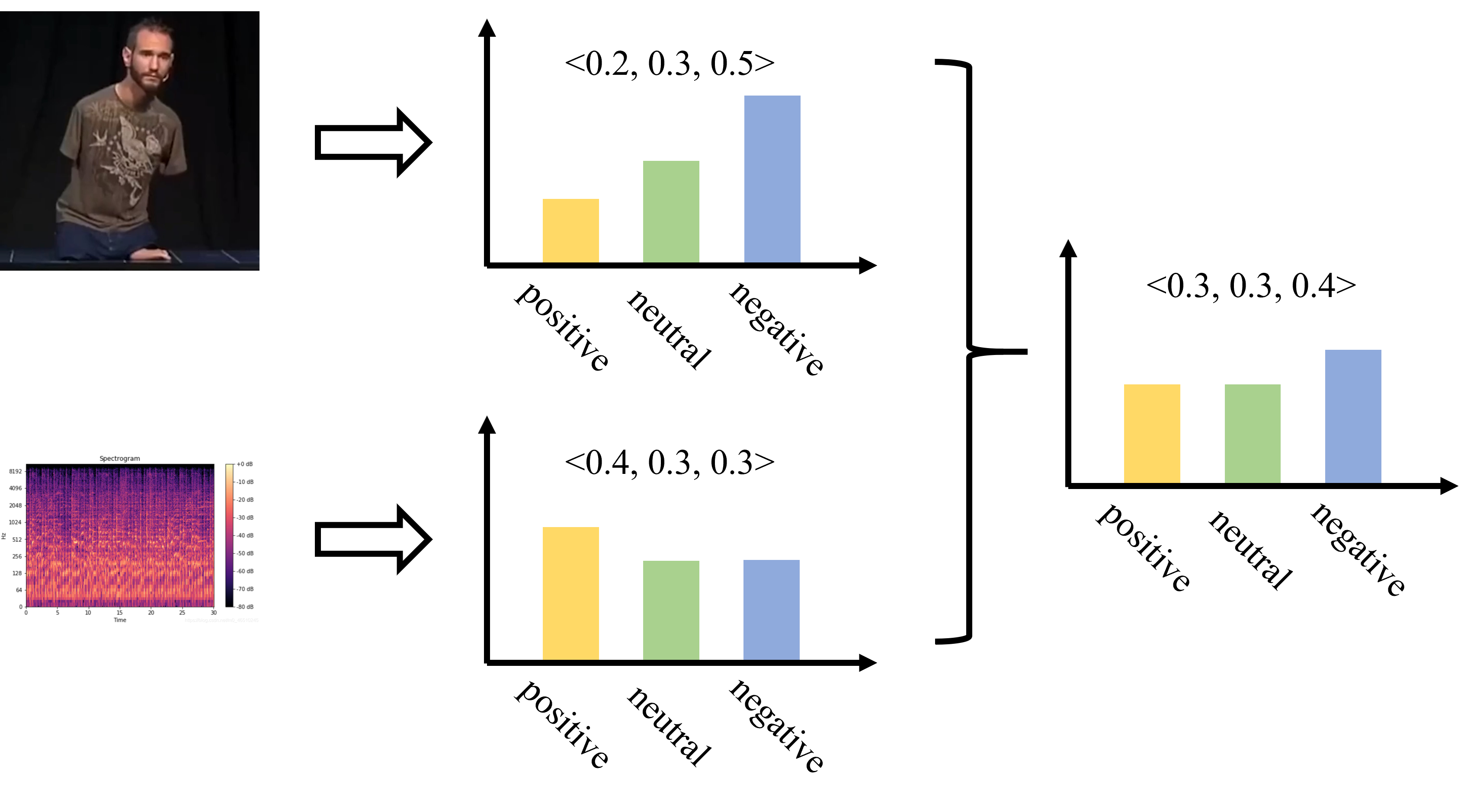}
            \label{fig4a}
        \end{minipage}
    }
    \subfigure[Within credibility]{
        \begin{minipage}{0.85\linewidth}
            \centering
            \includegraphics[width=\linewidth]{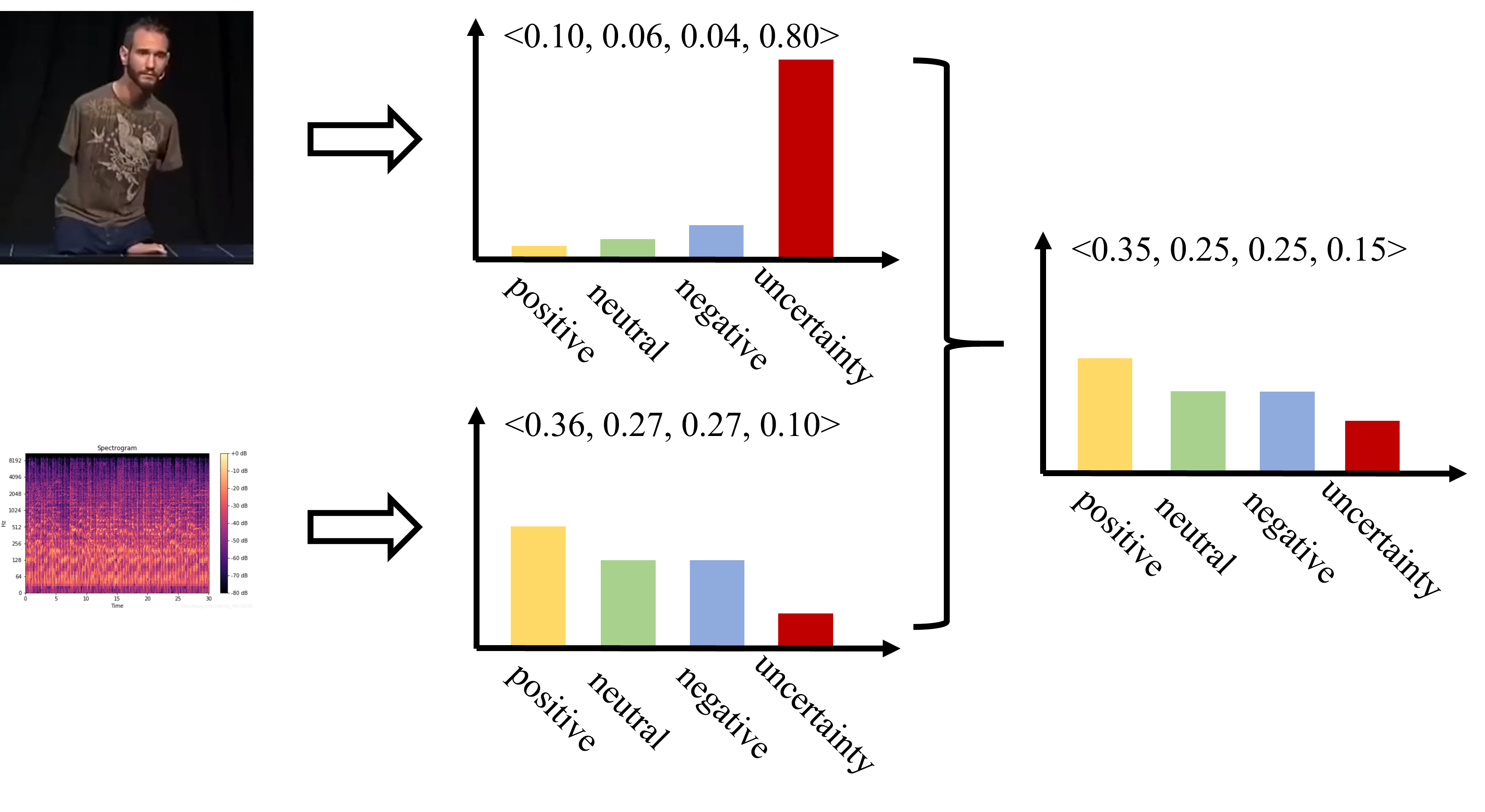}
            \label{fig4b}
        \end{minipage}
    }
    \caption{The credibility of model classification results plays a vital role in multimodal feature fusion. (a) The results of fusion without credibility in the image are incorrect. (b) Incorporating confidence as fusion weights leads to a correct output.}
    \label{fig4}
\end{figure}

The credibility of model classification results is increasingly receiving attention. In single-modal models, the credibility of the model indicates the probability that the model's classification results are worthy of adoption. In multimodal models, the credibility of different modality classification results may affect the overall credibility of the decision. For instance, in the case of the video depicted in Fig. \ref{fig4}, the multimodal model's classification results, as shown in Fig. \ref{fig4a}, indicate that the model classified the input data into incorrect labels in the video modalities. In contrast, the audio modality correctly classified the input data. Consequently, the final decision resulting from decision fusion could be influenced by the erroneous sub-results, leading to an incorrect output. In a study by Han et al. \cite{han2022trusted}, this issue was addressed by incorporating confidence as fusion weights, as shown in Fig. \ref{fig4b}. The outstanding performance of the credibility of model classification results in multimodal models warrants further research and exploration.

\subsection{Organization of This Survey}

In this survey, we will first focus on reviewing the latest approaches to AVCA and outline research trends. Specifically, we will provide a brief historical overview of AVCA in section \ref{brief} and introduce its comparison with other related topics in section \ref{comp}. Furthermore, we will introduce widely used emotion representation models in section \ref{modemo}. Thirdly, we will summarize the available datasets for AVCA evaluation in section \ref{datas}, comparing the sources of information and the number of information modalities included in the datasets. In section \ref{avca}, based on the primary objectives and challenges outlined in section \ref{goal}, we will summarize and compare video feature extraction methods, the effectiveness of different modality features, and multimodal fusion methods. Fifthly, we will present representative methods for evaluating emotion recognition results in section \ref{eval}. Finally, we will discuss potential research directions in section \ref{futd}. These contents will contribute to a comprehensive understanding of the latest developments and research trends in AVCA.

\subsection{Brief History}\label{brief}

\textbf{Affective Computing.} The origins of affective computing can be traced back to a patent application in 1978 \cite{williamson1979speech}. Subsequently, papers on speech emotion generation \cite{cahn1990generation} and facial expressions recognition \cite{kobayashi1992recognition} by neural networks introduced affective computing into the research domain in 1990 and 1992, respectively.

Since the inception of the concept of affective computing in the context of intelligent machines, research related to emotions, including the definition and recognition of emotions \cite{salovey1990emotional}, has garnered significant attention. In 1997, the concept of affective computing was first proposed, with Picard providing the following definition \cite{picard2000affective}: "affective computing is computing that relates to, arises from, or deliberately influences emotion or other affective phenomena". Subsequently, more researchers began to focus on the field of affective computing and organized important academic conference events. In 2005, the first International Conference on Affective Computing and Intelligent Interaction (ACII) was jointly held by IEEE and AAAI. The foundation of the Association for the Advancement of Affective Computing (AAAC) (originally named HUMAINE Association) was established in 2007. The public 'emotion challenge' began at Interspeech 2009, and the IEEE Transactions on Affective Computing (TAFFC) was first published in 2010. The Emotional and Social Signals in Multimedia area was introduced at the ACM Multimedia 2014.

\noindent\textbf{Affective Image Content Analysis.} AICA focuses on studying the relationship between images and emotions, with the aim of identifying and understanding the emotions evoked by images. Early emotion recognition methods were primarily based on handcrafted features such as Wiccest, Gabor \cite{yanulevskaya2008emotional}, artistic elements \cite{machajdik2010affective}, and Adjective Noun Pairs (ANPs) \cite{borth2013large}. However, with the emergence of CNNs in 2014, their outstanding performance has surpassed that of all handcrafted features. Through pre-training on large-scale data and model transfer learning of parameters, CNNs can extract more representative features, thus overcoming the challenge of manually designed features. In addition, AICA has also begun to focus on the challenge of perception subjectivity, with researchers considering personalized emotional prediction \cite{zhao2016predicting, yang2013user} and emotional distribution learning \cite{peng2015mixed, zhao2015predicting}. To address the challenge of missing labels, AICA models have introduced methods such as domain adaptation \cite{zhao2018emotiongan, zhao2019cycleemotiongan} and zero-shot learning \cite{zhan2019zero}. These methods aim to improve the accuracy and applicability of emotion recognition in order to understand better the emotions evoked by images.

\begin{figure}[t]
    \centering
    \includegraphics[width=\linewidth]{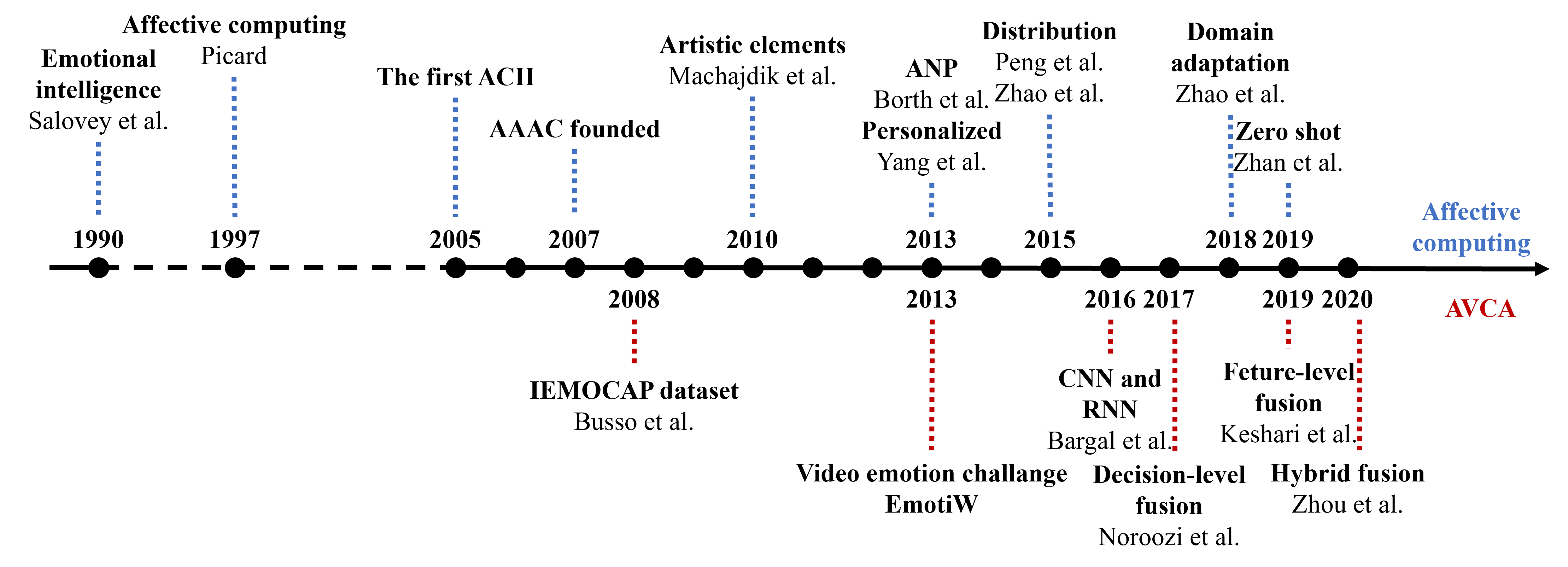}
    \caption{Milestones in both general affective computing (above line, blue) and affective video content analysis (below line, red).}
    \label{fig5}
\end{figure}

\noindent\textbf{Affective Video Content Analysis.} AVCA is an extension of AICA's research. With the iteration and evolution of social platforms, the mainstream mode of information dissemination is gradually shifting towards video format. Therefore, the transformation and optimization of existing networks are imperative. As the earliest public' video emotion challenge', EmotiW was held at the ACM International Conference on Multimodal Interaction (ICMI) in 2013. In early research, researchers achieved video feature extraction by combining CNN and RNN \cite{bargal2016emotion, hu2019video}. 3D CNNs have shown outstanding performance in video feature extraction, successfully addressing this challenge. Researchers have also started to focus on the multimodal information contained in videos. They have considered different feature fusion strategies, such as feature-level fusion \cite{keshari2019emotion, samadiani2022multiple}, decision-level fusion \cite{noroozi2017audio, avots2019audiovisual}, and hybrid fusion \cite{zhou2020hi}. The milestones in emotional computing and AVCA are summarized in Fig. \ref{fig5}.

\subsection{Comparison with Other Related Topics}\label{comp}

\textbf{Comparison with Affective Computing of Other Modalities.} Affective content analysis has been widely studied in other research fields, such as text \cite{giachanou2016like, zhang2018deep}, audio \cite{schuller2012automatic}, music \cite{schuller2010mister, yang2012machine}, images \cite{zhao2021affective}, facial expression \cite{pantic2006dynamics, sariyanidi2014automatic, hassan2019automatic}, and physiological signals \cite{alarcao2017emotions, zhao2019personalized}. Although similar affective models and learning methods are employed, there are significant differences between the affective computing of videos and that of other modalities, especially in the representation of emotional features from multimodal information. Despite the fruitful research conducted on other modalities, the study of AVCA is still not comprehensive enough. Considering the rich emotional information contained in videos, an in-depth analysis of AVCA will propel the development of the field of emotional computation.

\noindent\textbf{Comparison with Computer Vision.} AVCA generally involves three stages: manual annotation, extraction of visual features, and learning the mapping between emotional labels. Although these steps may seem similar to those in computer vision (CV), there are significant differences between AVCA and CV. (1) The subjective nature of emotional expression requires the manual annotation process to consider the personal and social backgrounds of the video presenters. For example, individuals from the Mediterranean region may express opposite emotions with a thumbs-up gesture compared to individuals from other regions. (2) Objects are objective concepts (e.g., gestures such as thumbs-up), while emotions are relatively subjective concepts (related to personal and social backgrounds). (3) Consequently, object classification belongs to the perceptual level of images, while AVCA focuses on the cognitive level. The CV community primarily studies object classification, whereas AVCA is an interdisciplinary task that requires support from psychology, cognitive science, multimedia, and machine learning, among other disciplines.

\section{Emotion Models from Psychology}\label{modemo}

Emotions play a significant role in our daily lives, exerting profound effects on human consciousness. Generally, emotions are spontaneously generated psychological states or feelings, representing an attitude or experience of individuals toward whether objective things satisfy their own needs \cite{minsky1988society}. This experience is subjective, with potential variations among different individuals in response to the same objective stimuli. Simultaneously, it exhibits a certain regularity, and the social nature of humanity results in a consistent tendency among individuals in the same social environment regarding emotions toward the same objective stimuli.

\begin{figure}[t]
    \centering
    \includegraphics[width=0.5\linewidth]{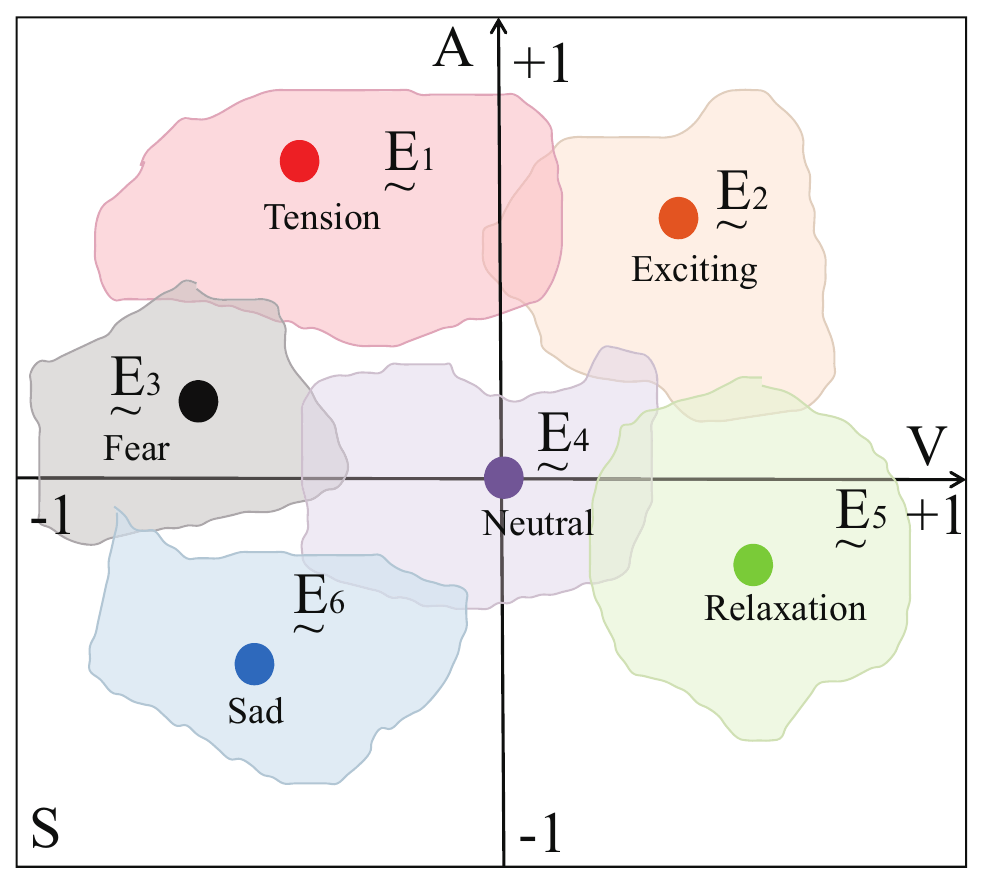}
    \caption{V-A emotional space and typical emotional subspaces (as cited in Plutchik et al., 1980)}
    \label{fig6}
\end{figure}

\begin{figure}[t]
    \centering
    \includegraphics[width=\linewidth]{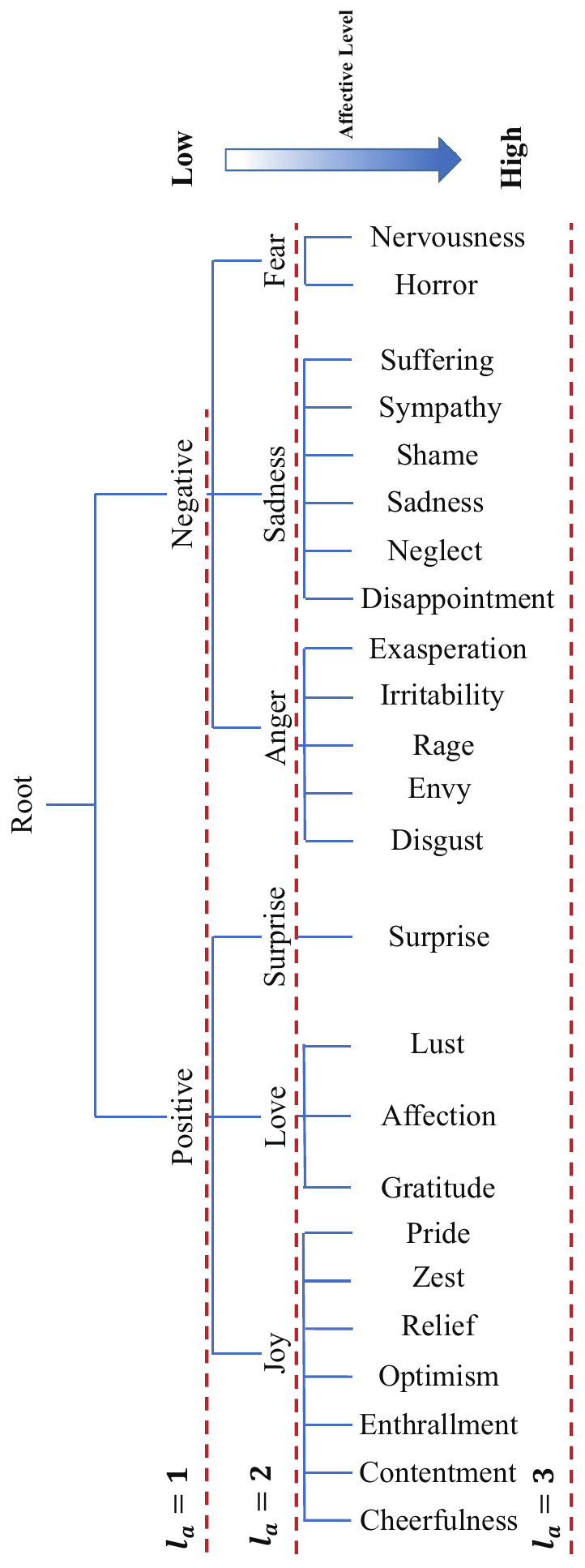}
    \caption{The three-level emotional hierarchy defined by Parrott in psychological research (as cited in Parrott et al., 2001)}
    \label{fig7}
\end{figure}

In affective computing tasks, researchers often use two different methods to describe emotions. One method is the categorical emotion states (CES), which categorizes emotions into discrete categories. In other words, the result of emotion recognition must be selected from a predefined list of word labels, such as the six emotion models proposed by Ekman \cite{ekman1992argument} (anger, disgust, fear, happiness, sadness, surprise) and the eight emotion models by Mikels \cite{mikels2005emotional} (amusement, anger, awe, contentment, disgust, excitement, fear, and sadness). With the development of psychological theories, discrete emotion models have become increasingly diverse and precise. Apart from the eight basic emotion categories, Plutchik \cite{plutchik1982psychoevolutionary} organized each emotion into three intensities, as shown in Fig. \ref{fig6}, thereby providing a richer set. For example, the three intensities of joy and fear are respectively ecstasy, happiness, serenity and terror, fear, unease. Another representative discrete emotion model is the tree hierarchical grouping proposed by Parrott \cite{parrott2001emotions}, which divides emotions into first, second, and third categories. For instance, a three-tier emotion is designed with two basic categories at level 1 (positive and negative), six categories at level 2 (anger, fear, happiness, love, sadness, surprise), and 25 refined emotional categories at level 3, as shown in Fig. \ref{fig7}.

\begin{figure}[t]
    \centering
    \begin{minipage}{0.50\linewidth}
        \centering
        \includegraphics[width=\linewidth]{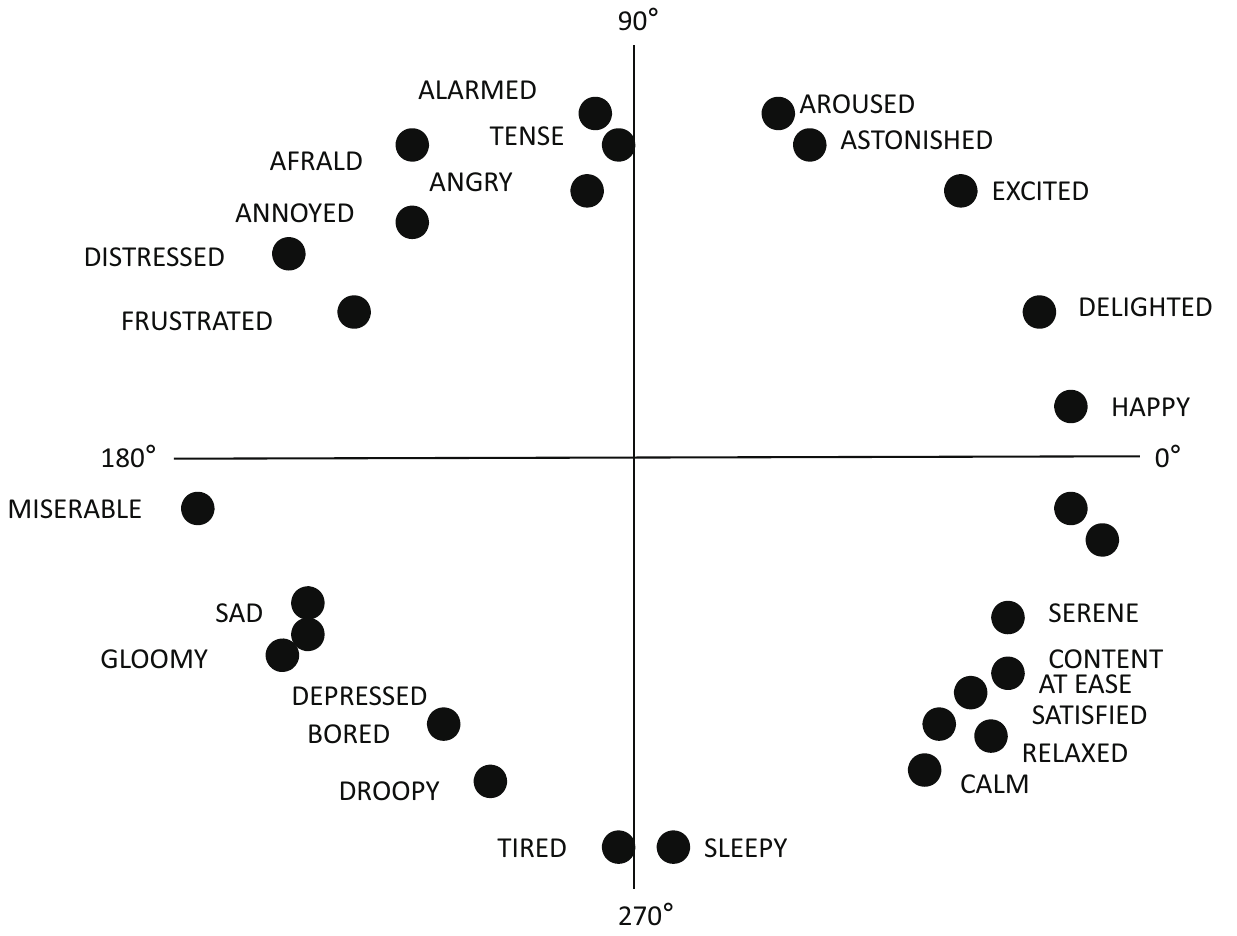} 
        \caption{The Circumflex Model of Russel (as cited in Russell et al., 1980, slightly modified)}
        \label{fig8}
    \end{minipage}
    \hfill
    \begin{minipage}{0.46\linewidth}
        \centering
        \includegraphics[width=\linewidth]{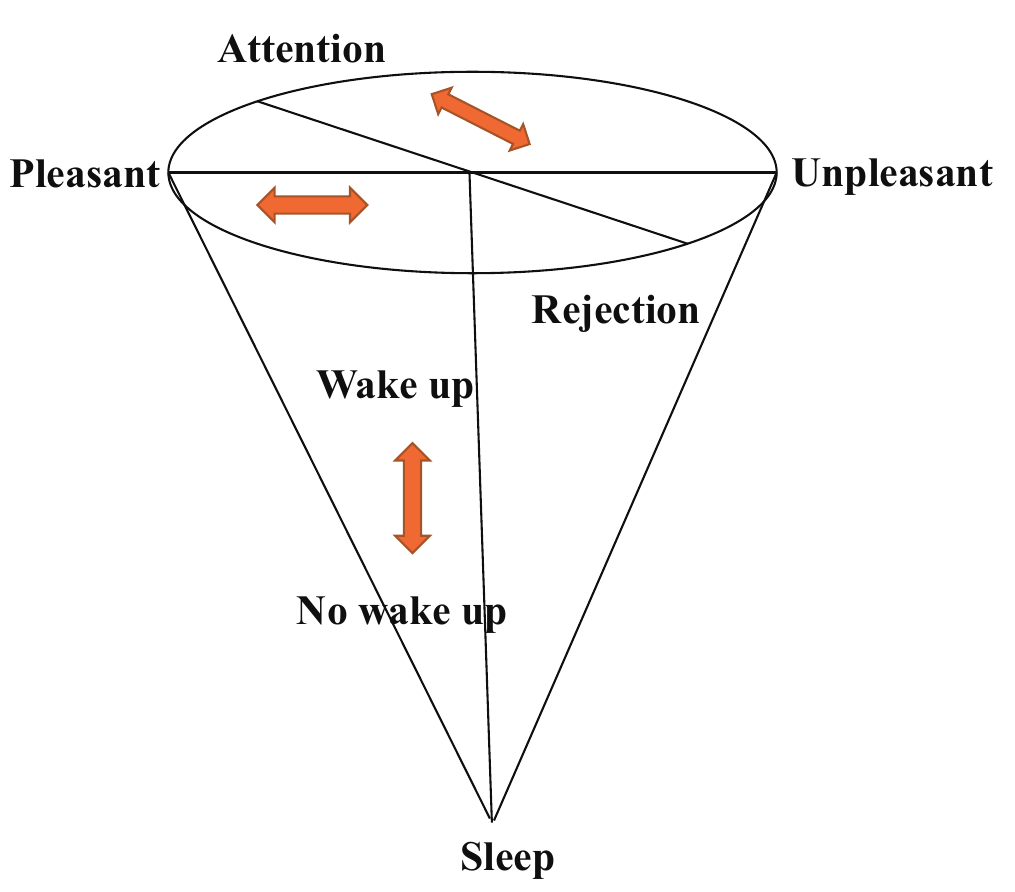}
        \caption{H. Schlosberg's three-dimensional model of emotion (as cited in Engen et al., 1958, slightly modified)}
        \label{fig9}
    \end{minipage}
\end{figure}

Another method is the dimensional emotion space (DES). Observers don't select discrete labels but can display their impressions of each stimulus on several continuous scales, such as pleasant-unpleasant, attention-rejection, simple-complex, and more. Two common scales are valence and arousal. Valence describes the pleasantness of a stimulus, with one end being positive (or pleasant) and the other negative (or unpleasant). For instance, happiness has a positive valence, while disgust has a negative valence. Another dimension is arousal. For example, sadness has a low arousal level, while surprise has a high arousal level. Different emotional labels can be plotted on these two axes, constructing a two-dimensional emotion model \cite{lang1995emotion}, like the emotion circumplex model proposed by Russell et al. \cite{russell1980circumplex}, as shown in Fig. \ref{fig8}. It's a circular model divided into quadrants, displaying the valence and arousal levels of emotional states. The x-axis represents a continuum between pleasant and unpleasant emotions, and the y-axis represents high and low arousal emotions, with the circle's center indicating neutral valence and moderate arousal. Subsequently, the American psychologist Engen et al. \cite{engen1958dimensional} proposed a three-dimensional model describing emotions with three independent dimensions: pleasant-unpleasant, attention-rejection, and high arousal-low arousal, as shown in Fig. \ref{fig9}. Different emotions can be distinguished in the emotional space based on the distinct characteristics of these three dimensions.

\begin{table}[t]
  \caption{Comparison of Two Emotion Description Models}
  \label{tab1}
  \begin{tabular}{cm{5cm}<{\centering}m{5cm}<{\centering}}
    \toprule
    Emotion Model & CES & DES \\
    \midrule
    Description & Emotion Label& Coordinate Point in Cartesian space \\
    Category & Limited & Arbitrary \\
    Examples & Mikels, Plutchik & VAD \\
    Advantages & Simple and easy to understand & Infinite, can describe spontaneous emotions and avoid ambiguity issues \\
    Disadvantages & Single, limited, unable to describe spontaneous emotions & Difficult to complete the conversion from qualitative to quantitative \\
  \bottomrule
\end{tabular}
\end{table}

The CES model and the DES model each have their own merits, as shown in Table \ref{tab1}. While the CES is easy to understand, the limited number of emotion categories fails to capture the complexity and subtlety of emotions fully. Additionally, psychologists have not reached a consensus on how many discrete emotion categories should be included. In contrast, the DES uses continuous two-dimensional, three-dimensional, or higher-dimensional Cartesian space to represent emotions. In theory, each emotion can be represented as a coordinate point in Cartesian space. However, the representation of continuous values is challenging to comprehend, limiting the use of the DES.

\section{Datasets}\label{datas}

The dataset plays a crucial role in training AVCA models and evaluating their effectiveness. Corresponding to the AVCA method, the dataset can be divided into single-modal and unimodal datasets. In unimodal datasets, samples are exclusively from the video modality. In contrast, multimodal datasets include various information beyond the video modality, such as other modal images (e.g., depth maps, infrared images), text, and audio. Refer to Table \ref{tab2} for a detailed comparison of mainstream datasets.

\begin{table}[h]
    \centering
    \caption{Released datasets for AVCA, where ‘\#Vidoes’ means the total number of videos, and 'V', 'A', 'T' in ‘Modals’ represent video data, audio data, text data}
    \label{tab2}
    \begin{tabular}{llrlm{1cm}<{\centering}lm{3cm}<{\centering}}
        \toprule
        Dataset & Ref & \#Videos & Modals & Source & Emotion model & Label detail\\
        \midrule
        IEMOCAP & \cite{busso2008iemocap} & 10,039 & V, A, T & Lab & Ekman & one sentiment category for each video\\
        YouTube & \cite{morency2011towards} & 47 & V, A, T & Internet & Sentiment & one sentiment category for each video\\
        News Rover & \cite{ellis2014we} & 929 & V, A, T & Media & Sentiment  & one sentiment category for each video\\
        ICT-MMMO & \cite{wollmer2013youtube} & 370 & V, A, T & Internet, Media & Sentiment  & one sentiment category for each video\\
        MOUD & \cite{perez2013utterance} & 412 & V, A, T & Internet & Sentiment  & one sentiment category for each video\\
        CH-SIMS & \cite{yu2020ch} & 2,281 & V, A, T & Media & VA & one average VA values for each video\\
        MELD & \cite{poria2018meld} & 13,708 & V, A, T & Media & Ekman & one sentiment category for each video\\
        CMU-MOSI & \cite{zadeh2016multimodal} & 2,199 & V, A, T & Internet & VA & one average VA values for each video\\
        CMU-MOSEI & \cite{zadeh2018multimodal} & 23,453 & V, A, T & Internet & Ekman, VA & one sentiment category, VA values for each video\\
        eNTERFACE’05 & \cite{martin2006enterface} & 1,166 & V, A & Lab & Ekman & one sentiment category for each video\\
        LIRIS-ACCEDE & \cite{baveye2015liris} & 9,800 & V, A & Media & VA & one average VA values for each video\\
        VideoEmotion-8 & \cite{jiang2014predicting} & 1,101 & V, A & Internet & Plutchik & one sentiment category for each video\\
        Ekman-6 & \cite{xu2016heterogeneous} & 1,637 & V, A & Internet & Plutchik & one sentiment category for each video\\
        SEWA & \cite{kossaifi2019sewa} & 2,000 & V, A & Lab & VA & one average VA values for each video\\
        \bottomrule
    \end{tabular}
\end{table}

\subsection{Unimodal Datasets}

The most commonly used datasets in the video unimodal data category are the MMI facial expression database \cite{valstar2010induced} and the AFEW database \cite{mollahosseini2016facial}. The MMI facial expression database comprises data from 61 adults displaying various basic emotions and 25 adults responding to emotional videos. The data are captured from frontal and side angles, resulting in over 2,900 videos and high-resolution still images of 75 subjects. This database encompasses six basic emotions: anger, disgust, fear, happiness, sadness, and surprise. Due to the diverse ways in which subjects express emotions, MMI is considered a challenging database.

\begin{figure}[h]
    \centering
    \begin{minipage}{0.48\linewidth}
        \centering
        \includegraphics[width=\linewidth]{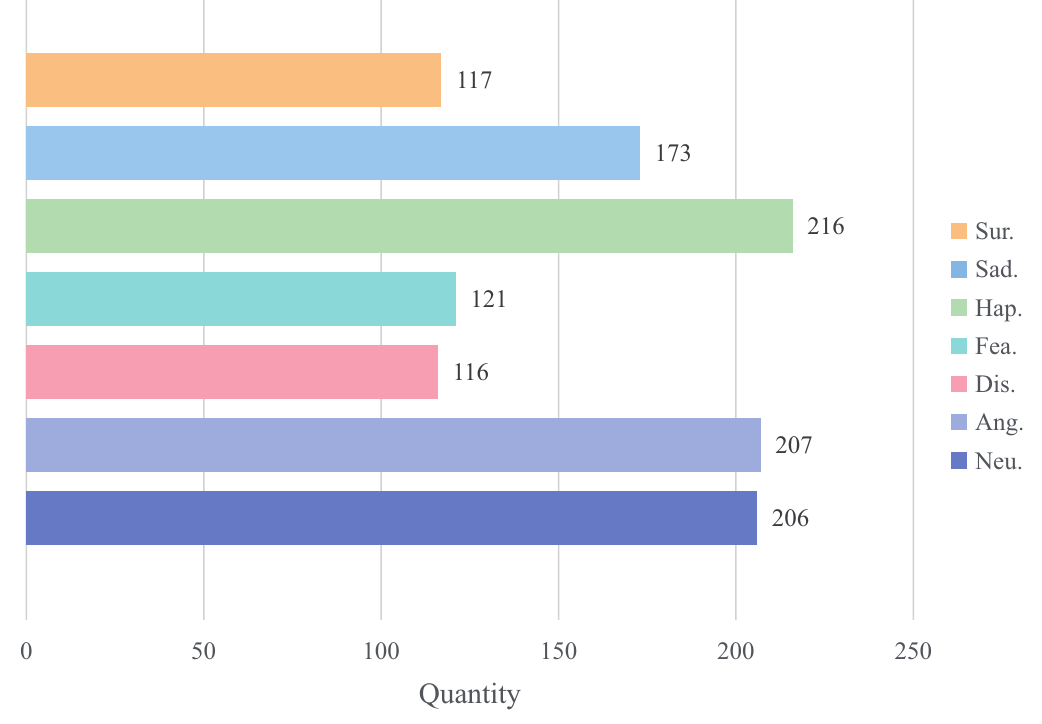} 
        \caption{Label distribution of AFEW dataset}
        \label{fig10}
    \end{minipage}
    \hfill
    \begin{minipage}{0.48\linewidth}
        \centering
        \includegraphics[width=\linewidth]{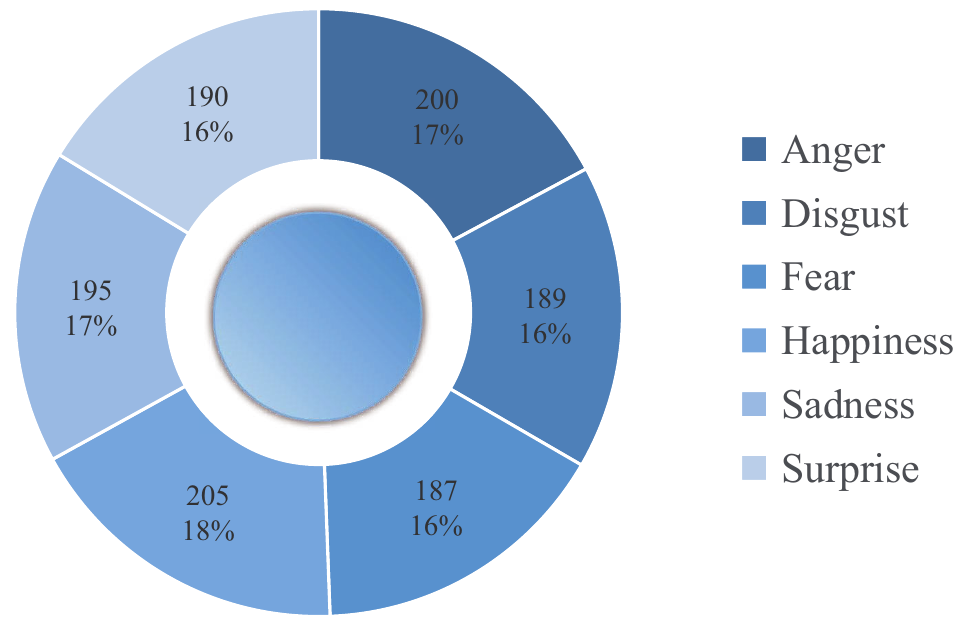}
        \caption{Distribution of emotion labels in the eNTERFACE'05 dataset}
        \label{fig11}
    \end{minipage}
\end{figure}

The AFEW database is currently the most widely used dataset for audio-visual emotion recognition. It is employed in the emotion recognition challenge held by AFEW corporation for the wild challenge. The database comprises video clips extracted from movies that include facial expressions, presenting an environment close to the real world and accurately reflecting human emotional expressions in authentic scenarios. With a total duration of approximately 1.2 hours, it includes a total of 1,809 video segments, covering six basic emotions (anger, disgust, fear, happiness, sadness, surprise) along with neutral emotions, totaling seven types of labels. The specific distribution of data labels is illustrated in Fig. \ref{fig10}.

\subsection{Multimodal Datasets}

Compared with unimodal emotion recognition using speech and facial expressions, research on multimodal emotion recognition is relatively recent. Therefore, the number of multimodal emotion datasets is limited, and most are either semi-open or closed. Many researchers have established their datasets similar to speech emotion datasets. These datasets are primarily categorized into spontaneous and induced types based on the way emotions are triggered. Spontaneous multimodal emotion datasets mainly originate from interviews or variety shows, while induced multimodal emotion datasets are primarily created through performances in a laboratory environment based on given scripts. Some commonly used multimodal emotion recognition databases include the IEMOCAP dataset and the eNTERFACE’05 dataset.

The IEMOCAP is a multimodal emotional dataset recorded collaboratively by Busso et al.. The dataset consists of 12 hours of audiovisual material, encompassing various aspects such as video, audio, and text. This dataset is created by 10 actors based on scripts or improvisational performances. Subsequently, each dialogue is segmented into multiple single sentences. Each sentence must encompass at least one of the emotions: joy, sadness, anger, surprise, fear, or disgust. Moreover, a minimum of three annotators is involved in the annotation process.

eNTERFACE'05 is a multimodal emotion dataset based on speech and facial expressions. Co-created by Martin et al. in 2006, the dataset comprises 1,287 videos involving 42 participants. Testers listened to six English short stories and responded to the specific emotional contexts portrayed in these narratives. Each audio-visual file is labeled with a single emotion, encompassing six basic emotions: happiness, sadness, anger, surprise, fear, and disgust. The emotional label distribution of this dataset is illustrated in Fig. \ref{fig11}.

From the above datasets, it can be observed that the existing datasets are mainly collected through script deduction or laboratory records, lacking genuine emotional records in natural states. Moreover, the types of modalities included in multimodal datasets need to be more comprehensive, with instances of partial modal data loss. Therefore, the trend in future dataset collection should be to increase the number of samples, consider multiple factors (such as region, race, personality, gender, age, etc.), pay attention to environmental changes, and consider adding sources of information. The more factors considered, the more comprehensive the dataset will be, providing researchers with data that will be more beneficial for exploring new methods of video emotion recognition.

\section{Affective Video Content Analysis}\label{avca}

In recent years, affective video content analysis has attracted significant attention and has been widely applied in various areas, including human-computer interaction, personalized video retrieval, and emotional video advertising. Generally, AVCA methods can be categorized into two main types: unimodal methods, including those based on facial expressions and action, and multimodal methods. We categorize mainstream AVCA methods into unimodal and multimodal types for discussion. We provide an overview of these methods in Table \ref{tab3}, introducing representative literature, methodological principles, and the strengths and weaknesses of existing approaches.

\begin{table}[h]
    \centering
    \caption{Results obtained with state-of-the-art video emotion recognition methods}
    \label{tab3}
    \renewcommand{\arraystretch}{2}
    \begin{tabular}{ccp{2cm}p{3cm}p{3cm}}
    
        \hline
        Method & Subissue & Ref & Advantages & Challenges \\
        \hline
        \multirow{4}{*}{Unimodal} & \multirow{2}{*}{FER} & \multirow{2}{*}{\cite{bargal2016emotion, fan2016video, xue2022coarse, hu2019video, savchenko2022classifying, hernandez2023multi}} & Rich training data, easy preprocessing, efficient deployment & Uncertain results, fine-grained identification \\
        & \multirow{2}{*}{AER} & \multirow{2}{*}{\cite{gavrilescu2015recognizing, shen2019emotion, wu2022generalized, santhoshkumar2019deep, liu2021imigue}} &  Rich training data, distinct feature differences & Uncertain results, coarse-grained identification \\
        \hline
        \multirow{5}{*}{Multimodal} & Feature-level & \cite{samadiani2022multiple, keshari2019emotion, cai2019feature, nguyen2018deep, chen2016emotion, ghaleb2017multimodal, qing2023dvc} & Rich feature information & Complex fusion rules, uncertain results \\
        & Decision-Level & \cite{avots2019audiovisual, noroozi2017audio, sahoo2016emotion, kim2017isla, xia2022multimodal, liu2022multi, praveen2022joint, zhu2020multimodal} & Intuitive fusion rules, easy understand & Underutilized multimodal features\\
        & Hybrid & \cite{zhou2020hi, le2022global, amer2018deep, huang2020multimodal, lian2021ctnet, john2022audio, chumachenko2022self} & Rich feature information, take advantage of multimodal features & Complex fusion rules, uncertain results \\
        \hline
        
    \end{tabular}
\end{table}

\subsection{Unimodal AVCA}

\textbf{Facial Emotion Recognition.} The facial emotion recognition (FER) method primarily involves capturing the facial expression features of individuals in videos to identify emotions. Facial expressions are an important means of emotional expression, with each change in the morphology of facial organs and muscles containing crucial information to support affective analysis. According to the research of the American psychologist Mehrabian \cite{mehrabian1974approach}, in the process of communication, facial expressions can convey 55\% of the information, tone of voice conveys 38\%, while the language itself conveys only 7\% of the information. Therefore, facial expressions can convey more emotional information compared to other methods. As a result, many researchers prefer to use facial expressions for emotion recognition.

Bargal et al. \cite{bargal2016emotion} adopted the facial detection method proposed by Chen et al. \cite{chen2016supervised}, using three deep learning networks - VGG13, VGG16, and ResNet - to extract facial expression features from videos. Subsequently, emotion recognition was performed on the features encoded through STST. The results of ablation experiments revealed that the combination of features, specifically using fc5 of VGG13, fc7 of VGG16, and the pool of ResNet, yielded the optimal performance in emotion recognition. Notably, features connected from different networks exhibited superior performance compared to features calculated from the same network. This feature combination achieved an accuracy of 59.42\% on the AFEW dataset, marking a 20.61\% improvement over the baseline method in the EmotiW’16 challenge.

Fan et al. \cite{fan2016video} proposed an emotion recognition system based on video, which adopts a late fusion approach of RNN and C3D. The RNN takes the appearance features extracted by CNN on individual video frames as input and encodes the motion, while C3D models both the appearance and motion of the video simultaneously. By incorporating an audio module, the system achieved a recognition accuracy of 59.02\% without using additional emotional labels for video clips, with an accuracy rate of 53.8\% in the EmotiW’15 challenge. Extensive comparative experiments have shown that the combined RNN and C3D model has a significant advantage in video emotion recognition.

Xue et al. \cite{xue2022coarse} proposed a coarse-to-fine cascaded network with smooth predictions (CFC-SP) for video-based facial emotion recognition. To address the issue of label ambiguity, they initially grouped several similar emotions into rough categories. Subsequently, they employed a cascaded network for preliminary classification, followed by fine classification. Furthermore, they introduced smooth prediction (SP) to enhance performance. The effectiveness of this method was validated through ablation experiments on the Aff-Wild2 dataset \cite{kollias2019expression}, achieving a final f1-score of 46.15\%. Regarding the label ambiguity issue, they identified two possible causes: the inherent ambiguity in expressions, where some expressions are similar and challenging to differentiate, and the diverse interpretations of facial expressions among individuals, leading to label ambiguity and inconsistency. For example, "sadness" and "disgust" are very similar and difficult to distinguish. Hence, they proposed a coarse-to-fine cascaded network to obtain more reliable predictions gradually.

\begin{figure}[h]
    \centering
    \includegraphics[width=0.7\linewidth]{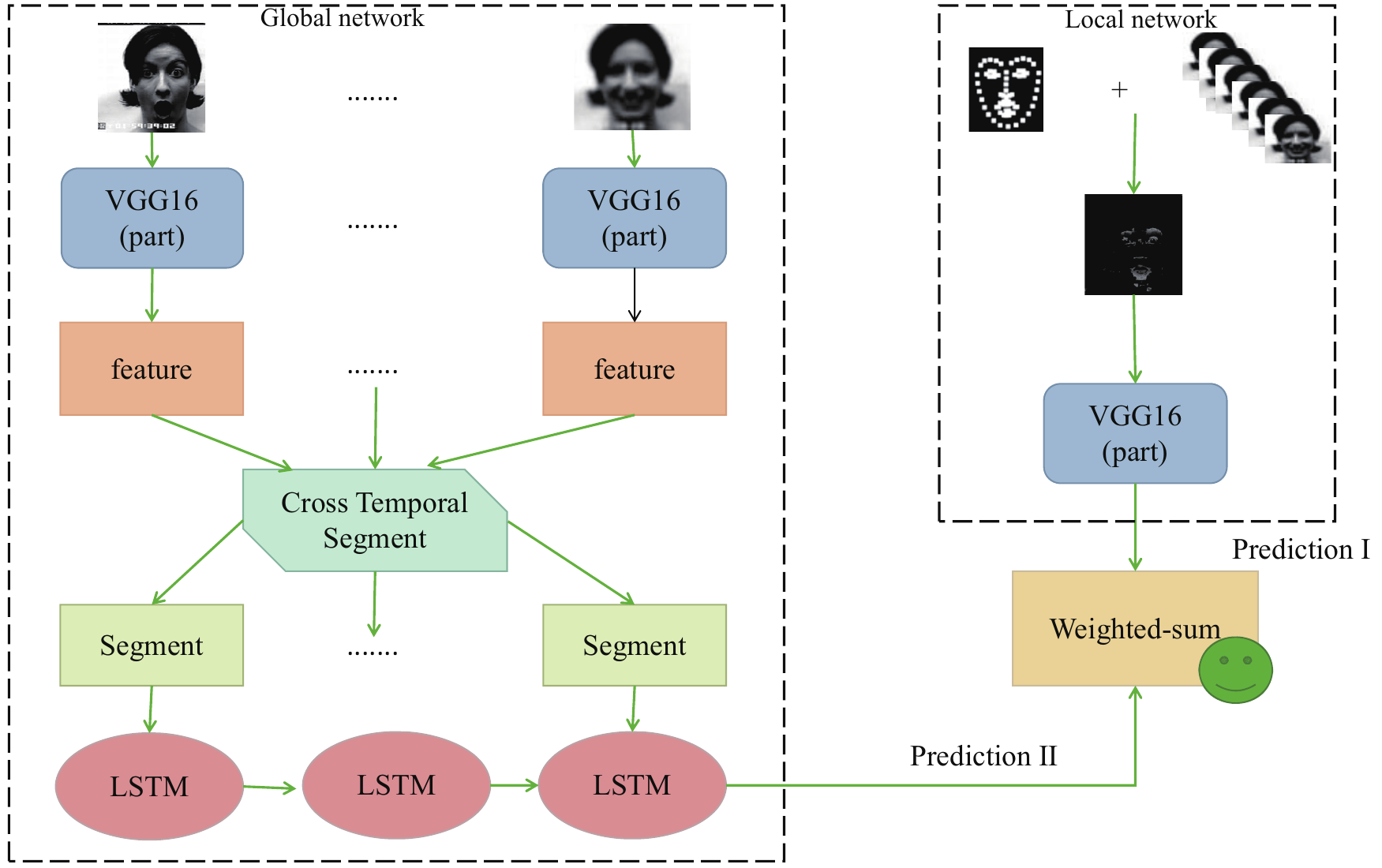}
    \caption{Integration framework diagram (as cited in Hu M et al.,2019)}
    \label{fig12}
\end{figure}

Hu et al. \cite{hu2019video} focused on the emotion recognition of facial expressions in video sequences. They proposed an integrated framework based on Local Enhanced Motion History Images (LEMHI) and a CNN-LSTM cascaded network, as illustrated in Fig. \ref{fig12}. In the local network, they employed a novel approach, LEMHI, to aggregate frames from unrecognized videos into a single frame. This method used detected facial landmarks as attention regions to enhance local values in differential image calculations, effectively capturing the movements of key facial units. Subsequently, this frame was input into a CNN network for prediction. On the other hand, they utilized an improved CNN-LSTM model as a global network, serving as a feature extractor and classifier for facial emotion recognition in video. Finally, a random search weighted sum strategy was employed as a late fusion method for the ultimate prediction. Experimental results on AFEW and MMI datasets yielded accuracies of 51.2\% and 78.4\%, respectively. The findings indicate that the integrated framework of the two networks performs better than the individual networks alone.

\noindent\textbf{Actional Emotion Recognition.} Emotion recognition in videos based on body postures and movements is an emerging research area. Its principle involves using the movement characteristics of individuals in videos to predict the emotions conveyed in the footage. Psychological studies have found that human perception can identify emotional states expressed through bodily movements \cite{luo2020arbee}. When expressing emotions, people exhibit habitual bodily actions involving hands, legs, shoulders, etc. For instance, trembling legs when nervous, shrugging shoulders when helpless, or dancing joyfully when happy. Dutch psychologist Beatrice de Gelder \cite{de2015perception} systematically examined the neural basis of bodily expression processing based on six basic emotion theories, discovering that emotional bodily stimuli significantly activate the amygdala, fusiform gyrus, and superior temporal sulcus, partially overlapping with brain areas involved in facial expression processing. As bodily movements are often subconscious actions, they are seldom deceitful. Hence, there has been a growing interest in recognizing emotions solely through body movements, postures, and gestures in recent years.

\begin{figure}[h]
    \centering
    \includegraphics[width=\linewidth]{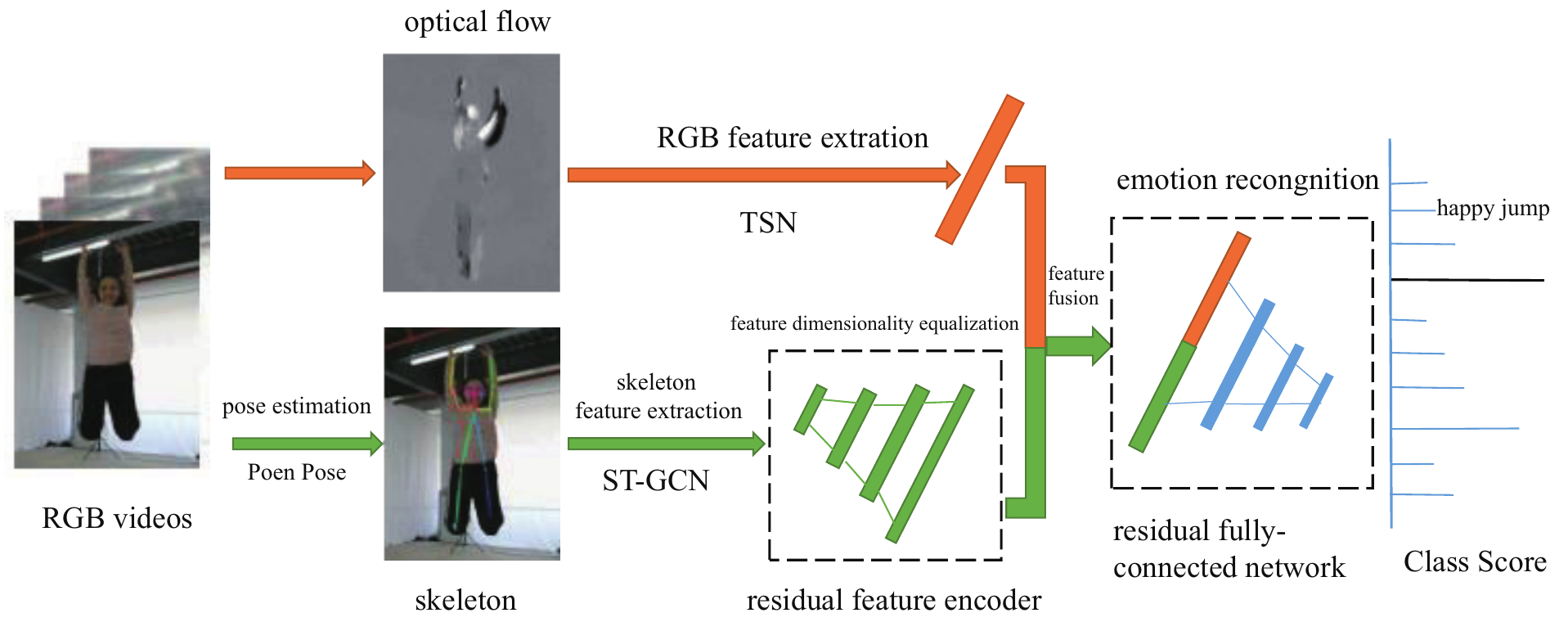}
    \caption{Schematic diagram of the ST-GCN model (as cited in Shen et al.,2019)}
    \label{fig13}
\end{figure}

Gavrilescu et al. \cite{gavrilescu2015recognizing} conducted experiments to verify that incorporating gesture emotion recognition can improve the classification accuracy of facial emotion recognition systems, indicating that gestures and body postures contain emotional information not obtainable from facial expressions. Therefore, studying individuals' gestures and body movements in videos can enhance the accuracy of traditional facial emotion recognition methods. Shen et al. \cite{shen2019emotion} conducted research on a self-collected dataset, which includes six postures: jumping, squatting, throwing, standing, retreating, and turning away. They utilized Temporal Segment Network (TSN) and Spatial-temporal Graph Convolutional Networks (ST-GCN) to extract behavioral and posture features and trained the emotion recognition model using an improved ResNet network. Ultimately, they achieved a recognition accuracy of 53.57\% on the self-collected dataset. Specifically, ST-GCN addressed the issue of traditional GCN's inability to model relative position changes between key nodes, proving highly effective in extracting skeletal posture features. The network framework is illustrated in Fig. \ref{fig13}.

In AICA studies, basic emotions are insufficient to describe the complex and diverse emotional states. New postures and gestures not included in the collected gesture samples may appear during testing. Based on this, Wu et al. \cite{wu2022generalized} proposed a new mechanism to address this issue by considering each emotional category as a set of multiple gesture categories. This approach aims to utilize gesture information for emotion recognition better. They introduced the Generalized Zero-shot Learning (GZSL) framework to identify visible and invisible body gesture categories using semantic information, and made emotion predictions based on the relationship between gestures and emotions. Ultimately, they achieved an accuracy of 67.85\% on the publicly available MASR \cite{psaltis2016multimodal} dataset.

\subsection{Multimodal AVCA}

\textbf{Feature-level Fusion,} also known as early fusion \cite{sarkar2014feature}, is a technique that uses feature concatenation as a fusion method. The design process involves extracting various modal data, constructing corresponding modal features, and designing cross-modal feature mapping rules. Samadian et al. \cite{samadiani2022multiple} proposed a video emotion recognition system, VERMFF, which combines audio and video modalities. They use an SR classifier to classify seven basic emotions and employ a kernel SR based on quality metrics to fuse multiple features. Additionally, they consider challenges such as head poses and lighting variations in outdoor-shot videos and propose a feature extraction method to address them. Ultimately, they achieved an accuracy of 54.39\% on the AFEW dataset for the seven-classification task. Adhikari et al. \cite{keshari2019emotion} introduced a feature-level fusion-based audio-visual multimodal emotion recognition system. They fuse features learned from audio-visual models and connect them to a dense layer meta-classifier. The fusion process involves simple linear, concatenation, or element-wise addition. They achieved a classification accuracy of 71.5\% on the eNTERACE’05 dataset for the seven-classification task.

\begin{figure}[h]
    \centering
    \includegraphics[width=\linewidth]{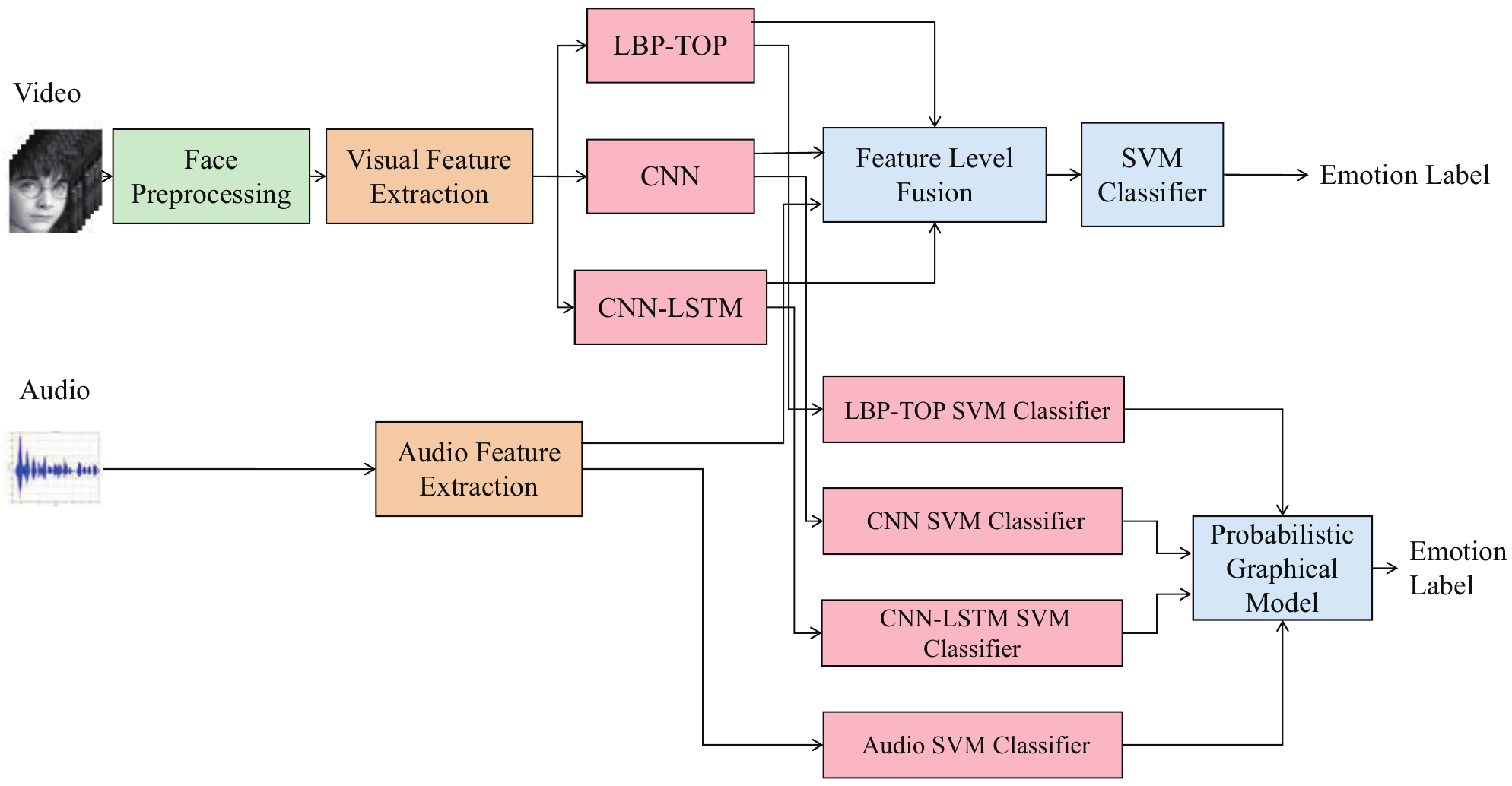}
    \caption{Flow chart of feature-level fusion and model-level fusion frameworks (as cited in Cai et al.,2019)}
    \label{fig14}
\end{figure}

Cai et al. \cite{cai2019feature} proposed two new audio-visual fusion methods. One is the feature-level fusion method, combining audio features with three different visual features: LBP-top-based, CNN-based, and CNN-BLSTM features. These are connected to form a joint feature vector through feature normalization, and each video clip is then inputted into a linear SVM for emotion recognition. The other is the model-level fusion method, explicitly addressing differences in time scale, temporal shift, and measurement from different signals. They independently extract audio and visual information for emotion recognition, and then combine these unimodal recognition results through a probabilistic framework (Bayesian network). In the inference process, the final decision is made by maximizing the posterior probability of all measured values. Experimental results on the AFEW database indicate that both proposed fusion methods significantly outperform baseline unimodal recognition methods and perform comparably or even better than state-of-the-art methods. Fig. \ref{fig14} illustrates the proposed audio-visual feature-level fusion and model-level fusion frameworks.

In contrast to the commonly used audio and video fusion frameworks, Nguyen et al. \cite{nguyen2018deep} proposed a novel emotion recognition framework. This framework integrates facial expressions, postures, body movements, and sound. This framework employed a cascaded 3D Convolutional Network (C3Ds) and Deep Belief Networks (DBNs) to introduce new depth-temporal features, effectively modeling spatiotemporal information in videos and audios for emotion recognition. Subsequently, they introduced a novel feature-level fusion method based on bilinear pooling theory to integrate visual and audio feature vectors. This fusion strategy allows all elements of component vectors to interact effectively, capturing the complexity and inherent correlations between component patterns. Extensive experiments conducted on the eNTERFACE’05 dataset demonstrate that this method can enhance the performance of multimodal emotion recognition, significantly outperforming existing approaches.

\noindent\textbf{Decision-level Fusion,} also known as late fusion \cite{lee2005toward}, is a technique that integrates single-modal decisions through specific fusion rules. The method design process includes inputting signals of different modalities into the corresponding models for feature extraction, using their respective classifiers for emotion recognition, and integrating the predictive results of each modality according to fusion rules. Common fusion methods include maximum value method, minimum value method, product method, sum method, average value method, voting method, Bayesian decision theory, Adaboost algorithm, and DS evidence theory.

Avots et al. \cite{avots2019audiovisual} conducted emotion recognition based on audio-visual information. For the video part, key frames of the video were selected for facial emotion recognition, and MFCC coefficients were extracted for the audio part. The decision-level feature fusion method used is as follows: 6 score values were set for each audio and video prediction, corresponding to the prediction accuracy of a specific category. The sum of all probabilities is 1, and the maximum value represents the predicted label. The probabilities are summed and normalized separately to obtain the final prediction. The system achieved an accuracy of 69.30\% in the six-classification task of the RML dataset \cite{dimou2014rml}. Noroozi et al. \cite{noroozi2017audio} proposed a multimodal emotion recognition system based on audio-visual clues. For the visual part, they first calculated the geometric relationships of facial landmarks and then summarized each emotional video into a set of simplified keyframes, which can distinguish emotions visually. Finally, the confidence outputs of classifiers for all modalities were used to define a new feature space for decision-level fusion for the final emotion label prediction. This system achieved an accuracy of 63.56\% in the six-classification task of the eNTERACE’05 dataset.

\begin{figure}[h]
    \centering
    \includegraphics[width=\linewidth]{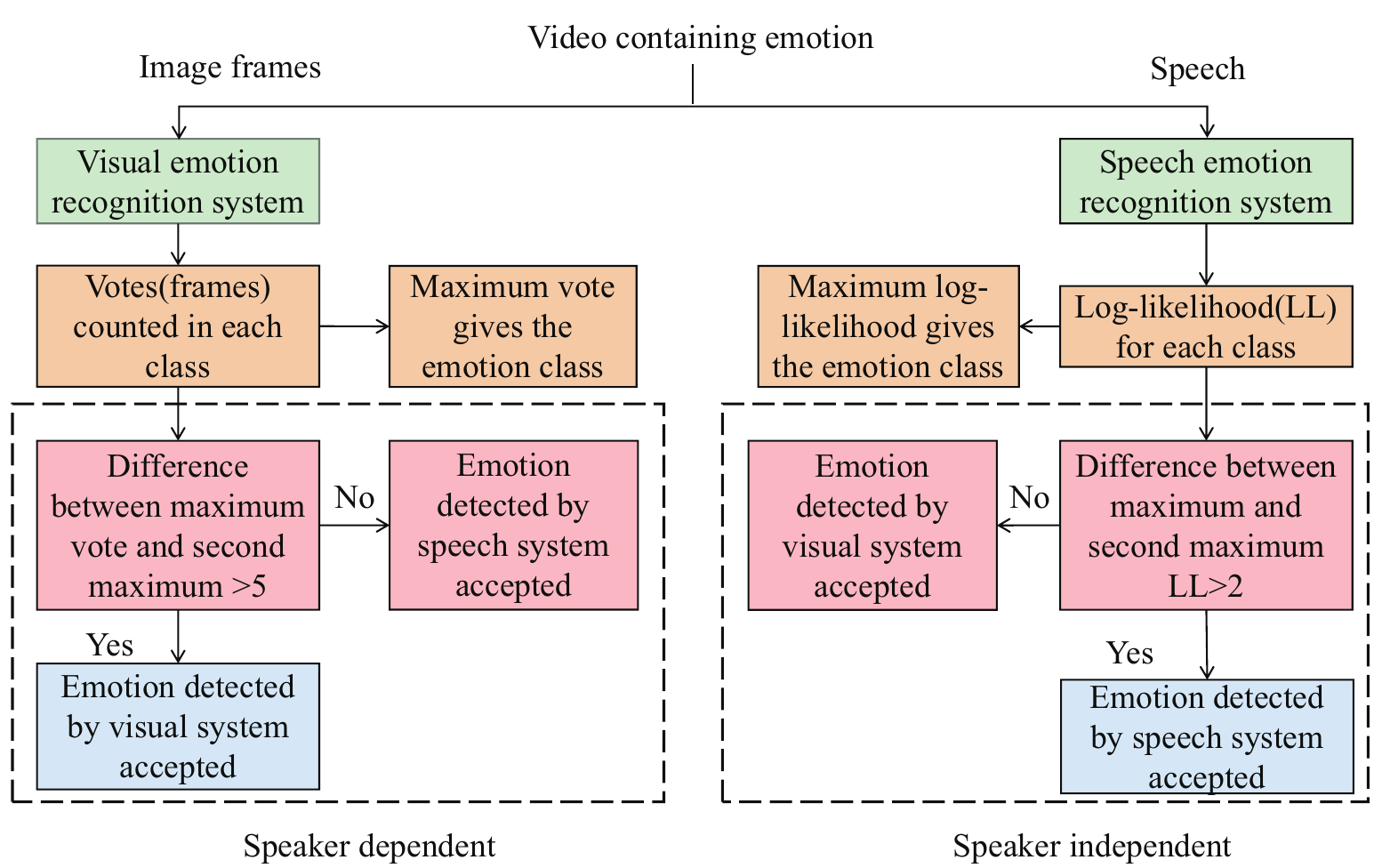}
    \caption{Decision rules for accepting any of the outcomes between audiovisual outcomes (as cited in Sahoo et al.,2016)}
    \label{fig15}
\end{figure}

Sahoo et al. \cite{sahoo2016emotion} proposed a multimodal emotion recognition method using facial images and voice data. The method employs a rule-based decision-level fusion approach, and the decision rules are illustrated in Fig. \ref{fig15}. The method achieved good results on the eNTERACE’05 dataset, with an accuracy of 81\% in situations related to the subjects and 54\% in situations independent of the subjects. However, this work was conducted in a laboratory environment without any ambient noise, using an emotion database recorded in the laboratory for validation. The effectiveness of the method in the presence of various environmental anomalies and its applicability to unknown genuine emotions and non-behavioral emotions remain unverified. Additionally, the decision rules set here are based on unimodal performance in the database. Applying the same rules with identical thresholds on different datasets may not be effective.

For videos, emotional signals exist only in certain segments rather than the entire video signal. Emotion recognition methods based on deep learning directly process input video signals, but they fail to capture the regions containing emotional information in the signal effectively. Moreover, they do not make optimal use of the characteristics of emotional information, significantly limiting the efficiency and performance of the algorithm. In response to this, Zhalehpour et al. \cite{zhalehpour2016multimodal} proposed a fully automatic multimodal emotion recognition system based on three new peak frame selection methods. These methods include MAX DIST, DEND-CLUSTER, and EIFS. Regarding the decision-level fusion of video and audio modalities, they tested several methods for obtaining probabilities and ultimately chose the weighted product rule \cite{wang2012kernel}, achieving an optimal emotion recognition accuracy of 78.26\% on the eNTERACE’05 dataset.

\begin{figure}[h]
    \centering
    \includegraphics[width=\linewidth]{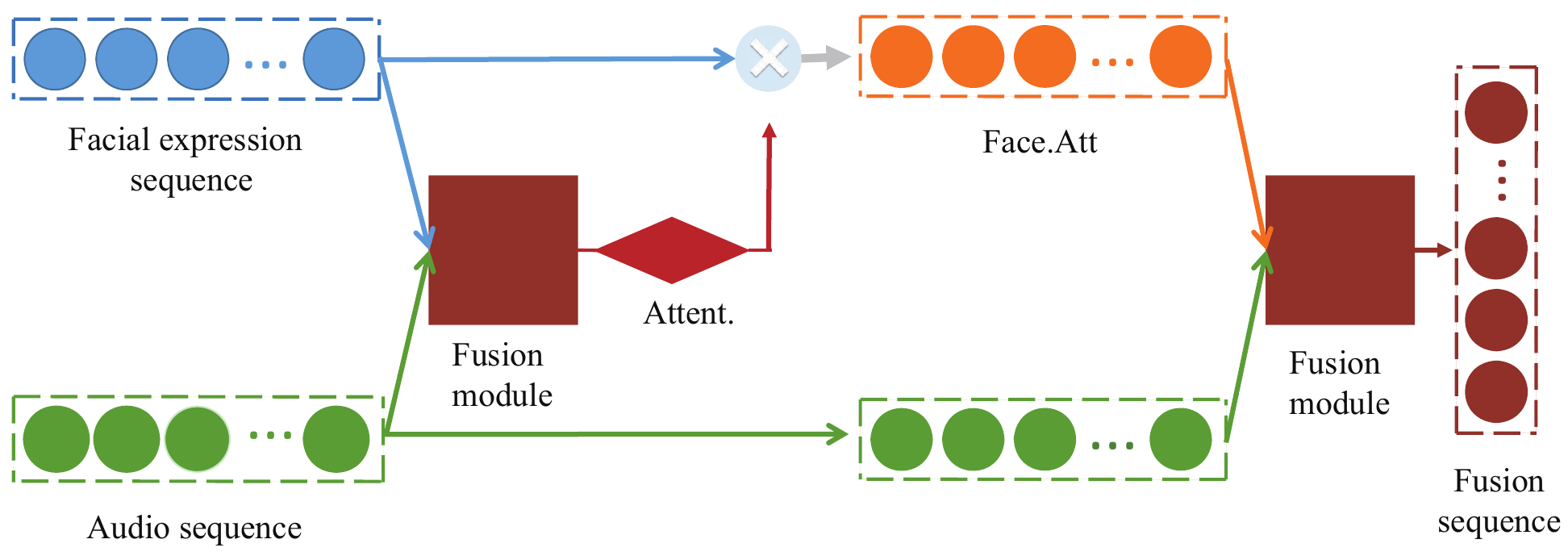}
    \caption{Structure diagram of the fusion model proposed by Liu D (as cited in Liu et al.,2022)}
    \label{fig16}
\end{figure}

Liu et al. \cite{liu2022multi} proposed a multimodal emotion recognition model based on an attention mechanism, as illustrated in Fig. \ref{fig16}. Firstly, facial features and audio features are fused to obtain integrated features. Subsequently, attention to facial features is highlighted using the fused features, and facial features are weighted. Finally, through attention analysis, facial and audio features are integrated to obtain the ultimate fused features. This method, through attention analysis of fused features, reveals relationships between features, assigning more weight to noise-free and highly discernible features while reducing the weight of noisy features. Ultimately, an 81.18\% recognition accuracy is achieved on the RML dataset.

Multiple modalities convey different valence and arousal information, and their complementary relationships need to be effectively captured. However, most advanced audio-visual fusion methods rely on recurrent networks or traditional attention mechanisms, and cannot effectively utilize the complementarity of audio-visual patterns. Based on this, Praveen et al. \cite{praveen2022joint} proposed an audio-visual emotion recognition model with joint cross-modal attention fusion to utilize the complementary relationships between modalities effectively. This model relies on a fusion mechanism based on cross-modal attention to encode inter-modal information while preserving intra-modal features. It calculates cross-modal attention weights based on the correlation between joint feature representation and individual modalities. Deploying the joint A-V feature representation into the cross-modal attention module significantly improves system performance by simultaneously utilizing intra-modal and inter-modal relationships. On the Affwild2 dataset, the consistency correlation coefficients for valence and arousal are 0.374 and 0.363, respectively, showing a significant improvement compared to the baseline of the third challenge in the 2022 Affective Behavior Analysis in-the-Wild competition.

In previous research on multimodal data fusion, tensor-based representations were predominantly used. When the input is transformed into a tensor, the dimensions and computational complexity grow exponentially. In light of this, Zhu et al. \cite{zhu2020multimodal} proposed a multimodal fusion approach employing a low-rank weight tensor with an attention mechanism. Specifically, in the self-attention module, a novel output vector calculation method, namely, proportionally weighted self-attention, was utilized. By obtaining an unimodal representation through an unimodal feature extraction network, this representation was fed into the fusion module. The self-attention mechanism was employed to generate an unimodal representation with new weights, which was then used to output the final classification result. Experimental studies indicate that this approach outperforms methods solely using low-rank tensor representations in multimodal fusion, enhancing both model efficiency and reducing computational complexity.

\begin{figure}[h]
    \centering
    \includegraphics[width=0.7\linewidth]{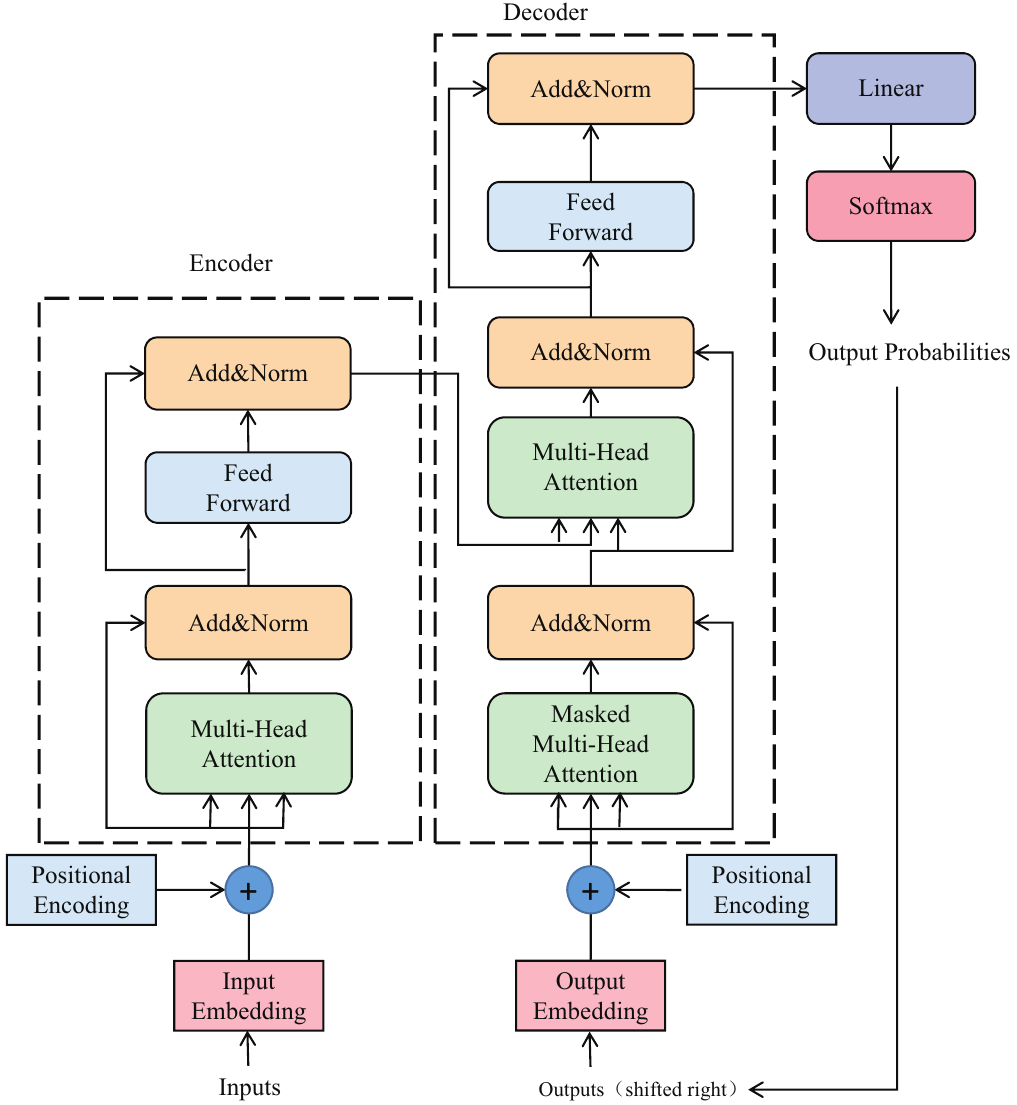}
    \caption{Transformer network structure diagram (as cited in Vaswani et al.,2017, slightly modified)}
    \label{fig17}
\end{figure}

\noindent\textbf{Hybrid Fusion.} Feature-level fusion enhances the richness of features, although it can improve recognition performance, it does not consider the differences between features. Some features cannot be integrated, resulting in the inability to model complex relationships, and high-dimensional feature sets are prone to data sparsity issues \cite{wu2014survey}. Decision-level fusion, while easy to implement, assumes that the modalities involved in fusion are mutually independent and cannot capture the interrelation between different modalities \cite{zhang2017learning}. Therefore, both feature-level and decision-level fusion methods struggle to capture deeper cross-modal information and fail to fully exploit the correlations between different modalities, making breakthrough progress challenging. To address these issues, researchers have proposed various more effective multimodal fusion approaches. Based on the attention mechanism of Transformer \cite{vaswani2017attention}, it can adaptively determine which information to extract from the data and generate a robust and effective fusion strategy. Transformer adaptively integrates useful information related to object queries, spatial and contextual relationships for feature fusion. Multi-head attention generates multimodal emotional intermediate representations from a common semantic feature space after encoding multimodal information. Additionally, it can effectively learn long-term dependencies with a self-attention mechanism. The network architecture diagram of Transformer is shown in Fig. \ref{fig17}.

Lian et al. \cite{lian2021ctnet} proposed a model-level multimodal fusion strategy based on Transformer. On the one hand, it captures the temporal dependencies between unimodal features through the Transformer-based unimodal structure. On the other hand, it learns cross-modal interactions on misaligned multimodal features through the Transformer-based cross-modal structure. It effectively models intra-modal and cross-modal interactions. Experiments on the AVEC 2017 database show that the Transformer-based model-level fusion outperforms other fusion strategies. Higher accuracy is achieved on the IEMOCAP and MELD datasets. Multimodal fusion enhances emotion recognition performance due to the complementary nature of different modalities. Compared to decision-level fusion and feature-level fusion, model-level fusion better utilizes the advantages of deep neural networks.

John et al. \cite{john2022audio} proposed adopting a new Transformer-based model to enhance the performance of audio-visual emotion recognition in videos. The model consists of three Transformer branches, referred to as multimodal Transformers. These three branches respectively perform audio self-attention, video self-attention, and audio-video cross-modal attention. The self-attention branches are used to identify the most relevant information in audio and video inputs, while the cross-modal attention branch is employed to recognize the most relevant audio-video interaction information. Through ablation experiments, the researchers verified that the performance benefits from the relevant information provided by these three branches. Additionally, a new time embedding scheme called block embedding was introduced, involving the incorporation of temporal information from multiple frames in the video into visual features.

Chumachenko et al. \cite{chumachenko2022self} proposed three modal fusion methods. The first is Late Transformer Fusion, where features learned from two branches are fused with a Transformer block. Specifically, two Transformers are used at the output of each branch, with one modality being fused into the other. The outputs of these Transformer blocks are further concatenated and passed to the final prediction layer. The second is Mid Transformer Fusion, which involves fusion at the mid-feature layer using Transformer blocks similar to the ones mentioned above. Specifically, after the first stage of feature extraction, i.e., after two convolutional layers, a Transformer block is used for fusion in each branch. The third is based on Attention Fusion, where they propose a fusion method solely based on dot-product similarity, forming the attention mechanism in the Transformer block. This method focuses on modality-independent features, a concept not previously introduced. Experimental validation on the MOSEI dataset shows that, among the three fusion methods, the Mid Attention Fusion method achieves the highest classification accuracy, reaching 68.76\%. It is followed by the Mid Transformer Fusion method, with an accuracy of 66.57\%.

\section{Experimental Analysis}\label{eval}

\subsection{Evaluation Criteria}

Each dataset has its evaluation method. In the dimensional emotion, annotations are typically divided into the following four groups: High Arousal and High Valence (HAHV), High Arousal and Low Valence (HALV), Low Arousal and High Valence (LAHV), Low Arousal and Low Valence (LALV). Consistency Concordance Correlation (CCC) is a statistical metric used to measure the consistency among multiple observers when assessing the same measured values. It can comprehensively consider the accuracy and consistency of the measured values, thus playing a crucial role in evaluating the reliability of measurement data. The CCC ranges from -1 to 1, with values closer to 1 indicating higher consistency among observers. The calculation of this coefficient takes into account the mean deviation and the linear relationship of the data, providing a more comprehensive reflection of the degree of consistency among observers. The following is the corresponding calculation formula:

\begin{equation}
    \rho_c=\frac{2\rho\sigma_x\sigma_y}{\sigma_x^2+\sigma_y^2+(\mu_x-\mu_y)^2}
\end{equation}

In the discrete dimension, annotations are usually categorized into different types of emotions, generally measuring the model's accuracy in analyzing different emotion labels. The most common evaluation criteria include accuracy (Acc.), recall, precision, and F1-score, as well as confusion matrices and ROC curves. Here are the corresponding calculation formulas:

\begin{equation}
    Accuracy=\frac{TP+TN}{TP+TN+FP+FN} 
\end{equation}

\begin{equation}
    Recall=\frac{TP}{TP+FN} 
\end{equation}

\begin{equation}
    Precision=\frac{TP}{TP+FP} 
\end{equation}

\begin{equation}
    F1-score=\frac{2\ast Precision\ast Recall }{Precision+Recall}  
\end{equation}

In this formula, TP represents the number of samples that are actually true and predicted as true, TN represents the number of samples that are actually false and predicted as false, FP represents the number of samples that are actually false but predicted as true, and FN represents the number of samples that are actually true but predicted as false.

\subsection{Experimental comparison}

This section introduces recent advancements in both unimodal and multimodal video emotion recognition methods, providing a comparison in Table \ref{tab4}. The table lists various video emotion recognition methods, the modalities utilized, datasets used for testing, and the experimental results obtained.

\begin{table}[h]
    \caption{Results obtained with state-of-the-art video emotion recognition methods}
    \label{tab4}
    \begin{tabular}{lllllr}
    
        \toprule
        \multirow{2}{*}{Ref} & \multirow{2}{*}{Modal} & \multirow{2}{*}{Subissue} & \multirow{2}{*}{Datadset} & \multicolumn{2}{c}{Performance} \\
        & & & &m easure & \% \\
        \midrule
        Kahou \cite{kahou2016emonets} & V & FER & AFEW & Acc. & 47.67\\
        Chao \cite{chao2015long} & V & FER & AV+EC2015 \cite{ringeval2015av+} & CCC & 53.80\\
        Wang \cite{wang2019multi} & V & FER & AFEW & Acc. & 58.65\\
        Du \cite{du2019spatio} & V & FER & Recola \cite{ringeval2013introducing} & CCC & 66.65\\
        Daoudi \cite{daoudi2017emotion} & V & AER & P-BME \cite{hicheur2013combined} & Acc. & 50.74\\
        Wei \cite{wei2020multimodal} & V & AER & Self-collection & Acc. & 53.57\\
        Nigam \cite{nigam2022emotion} & V & AER & Kinetics400 \cite{kay2017kinetics} & Acc. & 83.00\\
        Zhao \cite{zhao2020end} & V, A & Feature-level & Ekman-6 & Acc. & 55.30\\
        Wang \cite{wang2017emotion} & V, A & Feature-level & AFEW & Acc. & 58.50\\
        Njoku \cite{njoku2022deep} & V, A & Feature-level & RAVDESS\cite{livingstone2018ryerson} & Acc. & 58.33\\
        Liu \cite{liu2022multi} & V, A & Feature-level & RML & Acc. & 81.18\\
        Nguyen \cite{nguyen2018deep} & V, A & Feature-level & eNTERFACE‘05 & Acc. & 83.34\\
        Sahoo \cite{sahoo2016emotion} & V, A & Decision-level & eNTERFACE‘05 & Acc. & 76.00\\
        Poria \cite{poria2018meld} & V, A & Decision-level & YouTube & Acc. & 71.21\\
        Xia \cite{xia2022multimodal} & V, A & Decision-level & IEMOCAP & Acc. & 87.40\\
        Sun \cite{sun2021multimodal} & V, A, T & Decision-level & MuSe-CaR \cite{stappen2021multimodal} & CCC & 66.49\\
        Franceschini \cite{franceschini2022multimodal} & V, A & Decision-level & RAVDESS & Acc. & 91.40\\
        Mocanu \cite{mocanu2022audio} & V, A, T & Decision-level & RAVDESS & Acc. & 87.89\\
        Ren \cite{ren2021interactive} & V, A, T & Decision-level & IEMOCAP & Acc. & 65.50\\
        Zhu \cite{zhu2020multimodal} & V, A, T & Decision-level & IEMOCAP & Acc. & 71.90\\
        Zhang \cite{zhang2019deep} & V, A & Decision-level & AFEW & Acc. & 61.87\\
        Qi \cite{qi2021feature} & V, A, T & Decision-level & CMU-MOSEI & Acc. & 76.70\\
        Zhuang \cite{zhuang2022transformer} & V, A, T & Hybrid & CMU-MOSEI & Acc. & 73.80\\
        Guo \cite{guo2022er} & V, A, T & Hybrid & CMU-MOSEI & Acc. & 84.30\\
        Zhang \cite{zhang2022transformer} & V, A & Hybrid & Aff-Wild2 & F1 & 49.90\\
        Lv \cite{lv2021progressive} & V, A, T & Hybrid & IEMOCAP & Acc. & 85.05\\
        Shen \cite{shen2021mmtrans} & V, A, T & Hybrid & IEMOCAP & Acc. & 86.80\\
        Chaudhari \cite{chaudhari2023facial} & V, A, T & Hybrid & RAVDESS & Acc. & 87.60\\
        \bottomrule
        
    \end{tabular}
\end{table}

\section{Future Directions}\label{futd}

\begin{figure}[h]
    \centering
    \includegraphics[width=\linewidth]{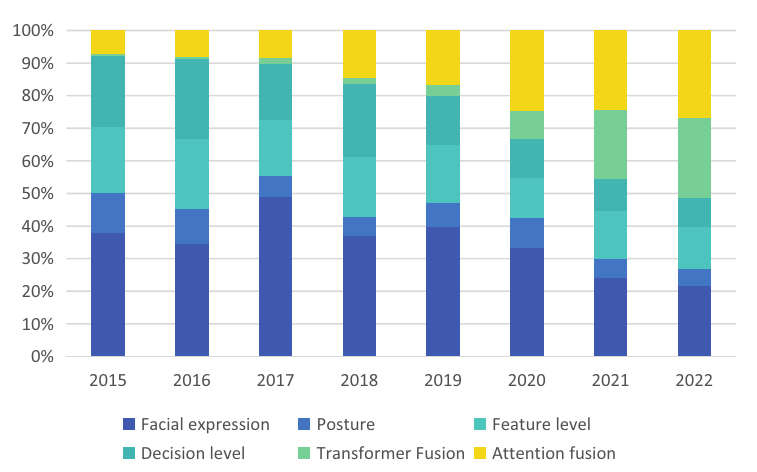}
    \caption{Percentage stacked histogram of various video emotion recognition methods}
    \label{fig18}
\end{figure}

Through an extensive survey of published literature on video emotion recognition from 2015 to 2022, this study finds that research on emotion recognition in video scenes is gradually gaining momentum, attracting continuous attention from many researchers. Fig. \ref{fig18} presents the proportional analysis of publications using different methods. From the figure, it can be observed that researchers currently favor attention-based multimodal fusion methods and are entering a period of rapid development. At the same time, the research focus is shifting gradually from unimodal to multimodal video emotion recognition. As evident from the preceding text, one of the driving factors behind this trend is that multimodal data typically complement each other, addressing their respective shortcomings [81], thereby improving the effectiveness of video emotion recognition and becoming a research hotspot. Furthermore, fusion methods based on attention mechanisms can dynamically generate weights, effectively addressing differences in modality quality caused by different samples in practice, thus possessing higher robustness and becoming a focus of current research in multimodal fusion. In summary, the trend in the field of video emotion recognition is moving towards multimodality, and finding superior multimodal fusion methods has become a key focus of current research.

\subsection{Video Feature Extraction}

Due to the emotional information conveyed by video content, accurately analyzing video content can significantly improve the performance of AVCA. In the model introduced in the Section \ref{avca}, part of the performance improvement comes from a more advanced video feature extraction module. Although in AVCA, temporal features are more valuable than spatial features, how to balance the two for better performance remains unclear. The introduction of an attention mechanism partially addresses this issue, with a general emphasis on temporal features and a stronger focus on noteworthy spatial features. The advanced performance of models incorporating attention mechanisms confirms the importance of judiciously utilizing both temporal and spatial features. Additionally, compared to spatial feature extraction, there is still room for improvement in temporal feature extraction techniques, drawing inspiration from advanced spatial feature extraction methods. For instance, introducing an attention mechanism for temporal features allows the model to automatically focus on time intervals in the video with richer emotional information, achieving a balance of spatiotemporal features. Correspondingly, research on three-dimensional convolution models still lags behind two-dimensional convolution models. In current research on three-dimensional convolutions, modifications to two-dimensional techniques still dominate, and innovative three-dimensional convolution models are still lacking. Taking model input as an example, current video models still require complex preprocessing of videos to receive data, while image models can directly take images as input. Therefore, there is a need for more innovative models specifically designed for videos to propel the rapid development of video feature extraction.

\subsection{Multimodal Feature Extraction}

The expression of human emotions is diverse. In addition to speech intonation and facial expressions, the modalities of emotional expression also include gestures, physiological signals, and text. Therefore, emotional recognition is fundamentally a problem of extracting multimodal emotional features. Currently, mainstream research on multimodal emotion recognition is based on speech and facial modalities, but there is a lack of in-depth study on other modal features. Thus, the integrated utilization of more modal information is an important direction for future emotion recognition, and incorporating additional modalities helps to analyze a person's emotional state more accurately. Various modal features have complementary and different importance in expressing spatial information. They are mutually independent yet interrelated, so how to leverage the complementarity and correlation between multimodal features is a question that needs consideration. By adopting effective fusion methods, multimodal features can be integrated to enhance the accuracy and robustness of emotion recognition. Additionally, exploring methods that combine difficult-to-fake physiological signals and easily collectible non-physiological signals is an area for future development in multimodal video emotion recognition. Furthermore, for most existing video emotion recognition methods, the utilized number of frames is limited, and they cannot fully utilize the video information, making it challenging to capture emotional information regions within the video signal effectively. Therefore, one of the future development directions is to devise methods that can utilize long-term video information and effectively capture the emotional information contained therein.

\subsection{Multimodal Feature Fusion}

Currently, the focus of research on multimodal fusion methods lies in considering how to balance different modalities to utilize the fusion features of various modalities better. Regarding the advantages and disadvantages of early and late fusion, hybrid fusion achieves relatively high accuracy. However, due to the need for model structure and fusion rules designed for different tasks and data, it lacks generality, posing challenges for widespread application in AVCA. Therefore, there is a need for further research on the trade-off between model optimization and generality. Designing more reasonable and versatile fusion rules while ensuring that the model's performance does not degrade is crucial and is a significant direction for future development.

Furthermore, existing late fusion methods often assume stability in the importance of different modalities for all samples, assigning equal or fixed weights to each modality. In practice, the importance of modalities often varies among different samples, and fusion rules should not overlook this issue. Thus, models should adaptively make decisions based on the input of multimodal features, designing appropriate fusion weights for each sample. Introducing attention mechanisms, where attention modules provide higher weights to important modalities, is one method to address this issue. Additionally, Han et al. \cite{han2022trusted} proposed a method using confidence for late fusion, adaptively integrating decisions for different modalities based on uncertainty estimation using evidence theory for each sample. Both approaches have shown significant performance improvements compared to existing methods, simultaneously enhancing the interpretability of fusion rules. Therefore, adaptive fusion rules will be a highly worthwhile direction for future multimodal research.

\subsection{Trusted Recognition Results}

The success of deep learning involves three factors: increased computing power, deep complex models, and sufficient data support. However, these factors all point out a significant drawback of deep learning, which is the lack of interpretability. This deficiency manifests in the interpretation of the learning process of deep learning models and the results of model recognition. This limitation necessitates careful consideration of decisions in practical applications, requiring not only the evaluation of the model's performance but also consideration of whether to use the model's results as the final decision. The evaluation of model results has become a crucial research direction in addressing this issue.

The study on credibility was first proposed by Guo et al. \cite{guo2017calibration}, who analyzed the correctness of multiple models in computer vision and natural language processing domains under different datasets. They pointed out an overall overconfidence tendency in existing deep learning models' confidence. Correction methods for the confidence of deep learning models have been widely studied, with main methods including Temperature Scaling, isotonic regression, Mix-n-Match, etc. Han et al. \cite{han2022trusted} combined modal confidence with multimodal decision fusion, using the confidence of each modal decision as fusion weights, achieving adaptive fusion. Introducing the concept of confidence further enhanced the model's performance and decision credibility, making fused decisions more interpretable. Credible research on deep learning decisions will be the most important research direction in the future of the deep learning field, enabling model decisions to be interpretable.

\subsection{Group Emotion Clustering}

Identifying the dominant emotions in videos is too general, while predicting personalized emotions for each user is too specific. Since certain user groups or small communities share similar tastes or interests and have similar backgrounds, they are more likely to react similarly to the same video. Therefore, predicting emotions for these user groups or small communities would be more meaningful. Analyzing user profiles provided by each individual and categorizing users into different groups based on gender, background, taste, interests, etc., may provide a feasible solution.

Currently, research on group emotions mainly focuses on identifying the emotions conveyed by video creators. Emotional video analysis for creator groups has not been explored. Group emotion recognition plays a crucial role in areas such as public opinion analysis. For instance, for the same group of people, if one person expresses a positive opinion about a certain event, others are more likely to express positive opinions.

\subsection{Novel and Real-world AVCA-based Applications}

With the improvement of multimodal AVCA methods, the model's performance has significantly increased. Therefore, we anticipate the arrival of the era of emotional intelligence, where there will be more applications based on emotion recognition. For example, in voice assistants, having emotion recognition capabilities can provide users with services that better meet their needs. In online shopping recommendations, intelligent consumer services such as customer image interaction can offer a better experience for customers. In sentiment analysis, AVCA models can quickly model public opinions on a large number of videos based on emotional knowledge. The literature \cite{thushara2016multimodal} has preliminarily implemented a video emotion recognition system that captures emotional features from input audio and facial image data, providing emotion results applicable in areas such as call centers, humanoid robots, and robotic pets. Amali et al. \cite{amali2018semantic} proposed an emotion-related video recommendation system that identifies emotions from videos watched by users now and in the past, recommending a video queue that better aligns with the user's emotions. In the education field, teaching assistants with rich emotions can help users learn and understand better. Of course, there are more exciting applications on the horizon.

\subsection{Large-scale AVCA Dataset}

The essential foundation of deep learning lies in data. Existing datasets mostly consist of clips from movies or records derived in laboratories, lacking the complexity of the real world. In the application of models, it becomes challenging to handle the intricate factors of the real world. Constructing a dataset that reflects real-world conditions with well-annotated data is advantageous for advancing AVCA research. Publicly available multimodal emotion datasets are relatively scarce compared to unimodal datasets, and some modalities may be missing or damaged \cite{garcia2023building}. Existing multimodal emotion databases are predominantly based on Western languages, overlooking regional and racial differences. Establishing a rich, multimodal, linguistically diverse, and openly accessible dataset in the field of multimodal AVCA will contribute to further research in AVCA, and creating a large-scale benchmark dataset is an important trend. With the development of internet technology, the widespread use of smartphones as mobile internet clients has led to the rapid growth of short videos. Most mobile internet users browse short videos due to their low production costs and strong entertainment appeal, catering to modern lifestyles. Establishing a large-scale short video dataset is necessary and crucial, as short videos increasingly dominate public opinion dissemination, leading to a growing demand for AVCA citations and driving the research forward.
Please note that the translation is focused on maintaining the academic and technical context while improving clarity and readability.

\section{Conclusion}

This article aims to comprehensively investigate the research progress in affective video content analysis (AVCA) over the past decade. Obviously, it cannot cover all the literature on AVCA, and we focus on a representative subset of the latest methods. We summarize and compare widely used emotion representation models, available datasets, and representative work in video feature extraction, model design, and multimodal analysis. Finally, we discuss some open issues and potential research directions in the field of AVCA. Despite the significant progress achieved with multimodal deep learning-based AVCA methods, there is still a need for designing an effective, efficient, robust, trustworthy, and interpretable AVCA algorithm. With the profound understanding of emotional arousal in neuroscience, the establishment of emotional measurement in psychology, and the rapid development of novel deep learning network architectures in machine learning, we believe that AVCA will continue to be an active and promising research topic.


\bibliography{sn-bibliography}


\begin{thebibliography}{144}
\ifx \bisbn   \undefined \def \bisbn  #1{ISBN #1}\fi
\ifx \binits  \undefined \def \binits#1{#1}\fi
\ifx \bauthor  \undefined \def \bauthor#1{#1}\fi
\ifx \batitle  \undefined \def \batitle#1{#1}\fi
\ifx \bjtitle  \undefined \def \bjtitle#1{#1}\fi
\ifx \bvolume  \undefined \def \bvolume#1{\textbf{#1}}\fi
\ifx \byear  \undefined \def \byear#1{#1}\fi
\ifx \bissue  \undefined \def \bissue#1{#1}\fi
\ifx \bfpage  \undefined \def \bfpage#1{#1}\fi
\ifx \blpage  \undefined \def \blpage #1{#1}\fi
\ifx \burl  \undefined \def \burl#1{\textsf{#1}}\fi
\ifx \doiurl  \undefined \def \doiurl#1{\url{https://doi.org/#1}}\fi
\ifx \betal  \undefined \def \betal{\textit{et al.}}\fi
\ifx \binstitute  \undefined \def \binstitute#1{#1}\fi
\ifx \binstitutionaled  \undefined \def \binstitutionaled#1{#1}\fi
\ifx \bctitle  \undefined \def \bctitle#1{#1}\fi
\ifx \beditor  \undefined \def \beditor#1{#1}\fi
\ifx \bpublisher  \undefined \def \bpublisher#1{#1}\fi
\ifx \bbtitle  \undefined \def \bbtitle#1{#1}\fi
\ifx \bedition  \undefined \def \bedition#1{#1}\fi
\ifx \bseriesno  \undefined \def \bseriesno#1{#1}\fi
\ifx \blocation  \undefined \def \blocation#1{#1}\fi
\ifx \bsertitle  \undefined \def \bsertitle#1{#1}\fi
\ifx \bsnm \undefined \def \bsnm#1{#1}\fi
\ifx \bsuffix \undefined \def \bsuffix#1{#1}\fi
\ifx \bparticle \undefined \def \bparticle#1{#1}\fi
\ifx \barticle \undefined \def \barticle#1{#1}\fi
\bibcommenthead
\ifx \bconfdate \undefined \def \bconfdate #1{#1}\fi
\ifx \botherref \undefined \def \botherref #1{#1}\fi
\ifx \url \undefined \def \url#1{\textsf{#1}}\fi
\ifx \bchapter \undefined \def \bchapter#1{#1}\fi
\ifx \bbook \undefined \def \bbook#1{#1}\fi
\ifx \bcomment \undefined \def \bcomment#1{#1}\fi
\ifx \oauthor \undefined \def \oauthor#1{#1}\fi
\ifx \citeauthoryear \undefined \def \citeauthoryear#1{#1}\fi
\ifx \endbibitem  \undefined \def \endbibitem {}\fi
\ifx \bconflocation  \undefined \def \bconflocation#1{#1}\fi
\ifx \arxivurl  \undefined \def \arxivurl#1{\textsf{#1}}\fi
\csname PreBibitemsHook\endcsname

\bibitem[\protect\citeauthoryear{Tajadura-Jim{\'e}nez}{2008}]{tajadura2008embodied}
\begin{botherref}
\oauthor{\bsnm{Tajadura-Jim{\'e}nez}, \binits{A.}}:
Embodied Psychoacoustics: Spatial and Multisensory Determinants of Auditory-induced Emotion,
Chalmers University of Technology Gothenburg
(2008)
\end{botherref}
\endbibitem

\bibitem[\protect\citeauthoryear{Asutay et~al.}{2012}]{asutay2012emoacoustics}
\begin{barticle}
\bauthor{\bsnm{Asutay}, \binits{E.}},
\bauthor{\bsnm{V{\"a}stfj{\"a}ll}, \binits{D.}},
\bauthor{\bsnm{Tajadura-Jimenez}, \binits{A.}},
\bauthor{\bsnm{Genell}, \binits{A.}},
\bauthor{\bsnm{Bergman}, \binits{P.}},
\bauthor{\bsnm{Kleiner}, \binits{M.}}:
\batitle{Emoacoustics: A study of the psychoacoustical and psychological dimensions of emotional sound design}.
\bjtitle{Journal of the Audio Engineering Society}
\bvolume{60}(\bissue{1/2}),
\bfpage{21}--\blpage{28}
(\byear{2012})
\end{barticle}
\endbibitem

\bibitem[\protect\citeauthoryear{Globerson et~al.}{2013}]{globerson2013psychoacoustic}
\begin{barticle}
\bauthor{\bsnm{Globerson}, \binits{E.}},
\bauthor{\bsnm{Amir}, \binits{N.}},
\bauthor{\bsnm{Golan}, \binits{O.}},
\bauthor{\bsnm{Kishon-Rabin}, \binits{L.}},
\bauthor{\bsnm{Lavidor}, \binits{M.}}:
\batitle{Psychoacoustic abilities as predictors of vocal emotion recognition}.
\bjtitle{Attention, Perception, \& Psychophysics}
\bvolume{75},
\bfpage{1799}--\blpage{1810}
(\byear{2013})
\end{barticle}
\endbibitem

\bibitem[\protect\citeauthoryear{Gross}{2002}]{gross2002emotion}
\begin{barticle}
\bauthor{\bsnm{Gross}, \binits{J.J.}}:
\batitle{Emotion regulation: Affective, cognitive, and social consequences}.
\bjtitle{Psychophysiology}
\bvolume{39}(\bissue{3}),
\bfpage{281}--\blpage{291}
(\byear{2002})
\end{barticle}
\endbibitem

\bibitem[\protect\citeauthoryear{Fischer et~al.}{}]{fischer2016social}
\begin{botherref}
\oauthor{\bsnm{Fischer}, \binits{A.H.}},
\oauthor{\bsnm{Manstead}, \binits{A.S.}}, et al.:
Social functions of emotion and emotion regulation
\end{botherref}
\endbibitem

\bibitem[\protect\citeauthoryear{Arandjelovic et~al.}{2016}]{arandjelovic2016netvlad}
\begin{bchapter}
\bauthor{\bsnm{Arandjelovic}, \binits{R.}},
\bauthor{\bsnm{Gronat}, \binits{P.}},
\bauthor{\bsnm{Torii}, \binits{A.}},
\bauthor{\bsnm{Pajdla}, \binits{T.}},
\bauthor{\bsnm{Sivic}, \binits{J.}}:
\bctitle{Netvlad: Cnn architecture for weakly supervised place recognition}.
In: \bbtitle{Proceedings of the IEEE Conference on Computer Vision and Pattern Recognition},
pp. \bfpage{5297}--\blpage{5307}
(\byear{2016})
\end{bchapter}
\endbibitem

\bibitem[\protect\citeauthoryear{Glenn et~al.}{2023}]{glenn2023emotion}
\begin{barticle}
\bauthor{\bsnm{Glenn}, \binits{A.}},
\bauthor{\bsnm{LaCasse}, \binits{P.}},
\bauthor{\bsnm{Cox}, \binits{B.}}:
\batitle{Emotion classification of indonesian tweets using bidirectional lstm}.
\bjtitle{Neural Computing and Applications}
\bvolume{35}(\bissue{13}),
\bfpage{9567}--\blpage{9578}
(\byear{2023})
\end{barticle}
\endbibitem

\bibitem[\protect\citeauthoryear{Zhao et~al.}{2018}]{zhao2018hsa}
\begin{bchapter}
\bauthor{\bsnm{Zhao}, \binits{B.}},
\bauthor{\bsnm{Li}, \binits{X.}},
\bauthor{\bsnm{Lu}, \binits{X.}}:
\bctitle{Hsa-rnn: Hierarchical structure-adaptive rnn for video summarization}.
In: \bbtitle{Proceedings of the IEEE Conference on Computer Vision and Pattern Recognition},
pp. \bfpage{7405}--\blpage{7414}
(\byear{2018})
\end{bchapter}
\endbibitem

\bibitem[\protect\citeauthoryear{Donahue et~al.}{2015}]{donahue2015long}
\begin{bchapter}
\bauthor{\bsnm{Donahue}, \binits{J.}},
\bauthor{\bsnm{Anne~Hendricks}, \binits{L.}},
\bauthor{\bsnm{Guadarrama}, \binits{S.}},
\bauthor{\bsnm{Rohrbach}, \binits{M.}},
\bauthor{\bsnm{Venugopalan}, \binits{S.}},
\bauthor{\bsnm{Saenko}, \binits{K.}},
\bauthor{\bsnm{Darrell}, \binits{T.}}:
\bctitle{Long-term recurrent convolutional networks for visual recognition and description}.
In: \bbtitle{Proceedings of the IEEE Conference on Computer Vision and Pattern Recognition},
pp. \bfpage{2625}--\blpage{2634}
(\byear{2015})
\end{bchapter}
\endbibitem

\bibitem[\protect\citeauthoryear{Carreira and Zisserman}{2017}]{carreira2017quo}
\begin{bchapter}
\bauthor{\bsnm{Carreira}, \binits{J.}},
\bauthor{\bsnm{Zisserman}, \binits{A.}}:
\bctitle{Quo vadis, action recognition? a new model and the kinetics dataset}.
In: \bbtitle{Proceedings of the IEEE Conference on Computer Vision and Pattern Recognition},
pp. \bfpage{6299}--\blpage{6308}
(\byear{2017})
\end{bchapter}
\endbibitem

\bibitem[\protect\citeauthoryear{Shaver et~al.}{1992}]{shaver1992cross}
\begin{botherref}
\oauthor{\bsnm{Shaver}, \binits{P.R.}},
\oauthor{\bsnm{Wu}, \binits{S.}},
\oauthor{\bsnm{Schwartz}, \binits{J.C.}}:
Cross-cultural similarities and differences in emotion and its representation.
(1992)
\end{botherref}
\endbibitem

\bibitem[\protect\citeauthoryear{Fischer et~al.}{2003}]{fischer2003social}
\begin{barticle}
\bauthor{\bsnm{Fischer}, \binits{A.H.}},
\bauthor{\bsnm{Manstead}, \binits{A.S.}},
\bauthor{\bsnm{Zaalberg}, \binits{R.}}:
\batitle{Social influences on the emotion process}.
\bjtitle{European review of social psychology}
\bvolume{14}(\bissue{1}),
\bfpage{171}--\blpage{201}
(\byear{2003})
\end{barticle}
\endbibitem

\bibitem[\protect\citeauthoryear{Plutchik}{1982}]{plutchik1982psychoevolutionary}
\begin{botherref}
\oauthor{\bsnm{Plutchik}, \binits{R.}}:
A psychoevolutionary theory of emotions.
Sage Publications
(1982)
\end{botherref}
\endbibitem

\bibitem[\protect\citeauthoryear{Ekman}{1992}]{ekman1992argument}
\begin{barticle}
\bauthor{\bsnm{Ekman}, \binits{P.}}:
\batitle{An argument for basic emotions}.
\bjtitle{Cognition \& emotion}
\bvolume{6}(\bissue{3-4}),
\bfpage{169}--\blpage{200}
(\byear{1992})
\end{barticle}
\endbibitem

\bibitem[\protect\citeauthoryear{Mikels et~al.}{2005}]{mikels2005emotional}
\begin{barticle}
\bauthor{\bsnm{Mikels}, \binits{J.A.}},
\bauthor{\bsnm{Fredrickson}, \binits{B.L.}},
\bauthor{\bsnm{Larkin}, \binits{G.R.}},
\bauthor{\bsnm{Lindberg}, \binits{C.M.}},
\bauthor{\bsnm{Maglio}, \binits{S.J.}},
\bauthor{\bsnm{Reuter-Lorenz}, \binits{P.A.}}:
\batitle{Emotional category data on images from the international affective picture system}.
\bjtitle{Behavior research methods}
\bvolume{37},
\bfpage{626}--\blpage{630}
(\byear{2005})
\end{barticle}
\endbibitem

\bibitem[\protect\citeauthoryear{Schlosberg}{1954}]{schlosberg1954three}
\begin{barticle}
\bauthor{\bsnm{Schlosberg}, \binits{H.}}:
\batitle{Three dimensions of emotion.}
\bjtitle{Psychological review}
\bvolume{61}(\bissue{2}),
\bfpage{81}
(\byear{1954})
\end{barticle}
\endbibitem

\bibitem[\protect\citeauthoryear{Lee and Park}{2011}]{lee2011fuzzy}
\begin{barticle}
\bauthor{\bsnm{Lee}, \binits{J.}},
\bauthor{\bsnm{Park}, \binits{E.}}:
\batitle{Fuzzy similarity-based emotional classification of color images}.
\bjtitle{IEEE Transactions on Multimedia}
\bvolume{13}(\bissue{5}),
\bfpage{1031}--\blpage{1039}
(\byear{2011})
\end{barticle}
\endbibitem

\bibitem[\protect\citeauthoryear{Cai et~al.}{2019}]{cai2019feature}
\begin{bchapter}
\bauthor{\bsnm{Cai}, \binits{J.}},
\bauthor{\bsnm{Meng}, \binits{Z.}},
\bauthor{\bsnm{Khan}, \binits{A.S.}},
\bauthor{\bsnm{Li}, \binits{Z.}},
\bauthor{\bsnm{O’Reilly}, \binits{J.}},
\bauthor{\bsnm{Han}, \binits{S.}},
\bauthor{\bsnm{Liu}, \binits{P.}},
\bauthor{\bsnm{Chen}, \binits{M.}},
\bauthor{\bsnm{Tong}, \binits{Y.}}:
\bctitle{Feature-level and model-level audiovisual fusion for emotion recognition in the wild}.
In: \bbtitle{2019 IEEE Conference on Multimedia Information Processing and Retrieval (MIPR)},
pp. \bfpage{443}--\blpage{448}
(\byear{2019}).
\bcomment{IEEE}
\end{bchapter}
\endbibitem

\bibitem[\protect\citeauthoryear{Sahoo and Routray}{2016}]{sahoo2016emotion}
\begin{bchapter}
\bauthor{\bsnm{Sahoo}, \binits{S.}},
\bauthor{\bsnm{Routray}, \binits{A.}}:
\bctitle{Emotion recognition from audio-visual data using rule based decision level fusion}.
In: \bbtitle{2016 IEEE Students’ Technology Symposium (TechSym)},
pp. \bfpage{7}--\blpage{12}
(\byear{2016}).
\bcomment{IEEE}
\end{bchapter}
\endbibitem

\bibitem[\protect\citeauthoryear{Lian et~al.}{2021}]{lian2021ctnet}
\begin{barticle}
\bauthor{\bsnm{Lian}, \binits{Z.}},
\bauthor{\bsnm{Liu}, \binits{B.}},
\bauthor{\bsnm{Tao}, \binits{J.}}:
\batitle{Ctnet: Conversational transformer network for emotion recognition}.
\bjtitle{IEEE/ACM Transactions on Audio, Speech, and Language Processing}
\bvolume{29},
\bfpage{985}--\blpage{1000}
(\byear{2021})
\end{barticle}
\endbibitem

\bibitem[\protect\citeauthoryear{Han et~al.}{2022}]{han2022trusted}
\begin{barticle}
\bauthor{\bsnm{Han}, \binits{Z.}},
\bauthor{\bsnm{Zhang}, \binits{C.}},
\bauthor{\bsnm{Fu}, \binits{H.}},
\bauthor{\bsnm{Zhou}, \binits{J.T.}}:
\batitle{Trusted multi-view classification with dynamic evidential fusion}.
\bjtitle{IEEE transactions on pattern analysis and machine intelligence}
\bvolume{45}(\bissue{2}),
\bfpage{2551}--\blpage{2566}
(\byear{2022})
\end{barticle}
\endbibitem

\bibitem[\protect\citeauthoryear{Williamson}{1979}]{williamson1979speech}
\begin{botherref}
\oauthor{\bsnm{Williamson}, \binits{J.D.}}:
Speech analyzer for analyzing frequency perturbations in a speech pattern to determine the emotional state of a person.
Google Patents.
US Patent 4,142,067
(1979)
\end{botherref}
\endbibitem

\bibitem[\protect\citeauthoryear{Cahn}{1990}]{cahn1990generation}
\begin{barticle}
\bauthor{\bsnm{Cahn}, \binits{J.E.}}:
\batitle{The generation of affect in synthesized speech}.
\bjtitle{Journal of the American Voice I/O Society}
\bvolume{8}(\bissue{1}),
\bfpage{1}--\blpage{1}
(\byear{1990})
\end{barticle}
\endbibitem

\bibitem[\protect\citeauthoryear{Kobayashi and Hara}{1992}]{kobayashi1992recognition}
\begin{bchapter}
\bauthor{\bsnm{Kobayashi}, \binits{H.}},
\bauthor{\bsnm{Hara}, \binits{F.}}:
\bctitle{Recognition of six basic facial expression and their strength by neural network}.
In: \bbtitle{[1992] Proceedings IEEE International Workshop on Robot and Human Communication},
pp. \bfpage{381}--\blpage{386}
(\byear{1992}).
\bcomment{IEEE}
\end{bchapter}
\endbibitem

\bibitem[\protect\citeauthoryear{Salovey and Mayer}{1990}]{salovey1990emotional}
\begin{barticle}
\bauthor{\bsnm{Salovey}, \binits{P.}},
\bauthor{\bsnm{Mayer}, \binits{J.D.}}:
\batitle{Emotional intelligence}.
\bjtitle{Imagination, cognition and personality}
\bvolume{9}(\bissue{3}),
\bfpage{185}--\blpage{211}
(\byear{1990})
\end{barticle}
\endbibitem

\bibitem[\protect\citeauthoryear{Picard}{2000}]{picard2000affective}
\begin{botherref}
\oauthor{\bsnm{Picard}, \binits{R.W.}}:
Affective Computing,
MIT press
(2000)
\end{botherref}
\endbibitem

\bibitem[\protect\citeauthoryear{Yanulevskaya et~al.}{2008}]{yanulevskaya2008emotional}
\begin{bchapter}
\bauthor{\bsnm{Yanulevskaya}, \binits{V.}},
\bauthor{\bsnm{Gemert}, \binits{J.C.}},
\bauthor{\bsnm{Roth}, \binits{K.}},
\bauthor{\bsnm{Herbold}, \binits{A.-K.}},
\bauthor{\bsnm{Sebe}, \binits{N.}},
\bauthor{\bsnm{Geusebroek}, \binits{J.-M.}}:
\bctitle{Emotional valence categorization using holistic image features}.
In: \bbtitle{2008 15th IEEE International Conference on Image Processing},
pp. \bfpage{101}--\blpage{104}
(\byear{2008}).
\bcomment{IEEE}
\end{bchapter}
\endbibitem

\bibitem[\protect\citeauthoryear{Machajdik and Hanbury}{2010}]{machajdik2010affective}
\begin{bchapter}
\bauthor{\bsnm{Machajdik}, \binits{J.}},
\bauthor{\bsnm{Hanbury}, \binits{A.}}:
\bctitle{Affective image classification using features inspired by psychology and art theory}.
In: \bbtitle{Proceedings of the 18th ACM International Conference on Multimedia},
pp. \bfpage{83}--\blpage{92}
(\byear{2010})
\end{bchapter}
\endbibitem

\bibitem[\protect\citeauthoryear{Borth et~al.}{2013}]{borth2013large}
\begin{bchapter}
\bauthor{\bsnm{Borth}, \binits{D.}},
\bauthor{\bsnm{Ji}, \binits{R.}},
\bauthor{\bsnm{Chen}, \binits{T.}},
\bauthor{\bsnm{Breuel}, \binits{T.}},
\bauthor{\bsnm{Chang}, \binits{S.-F.}}:
\bctitle{Large-scale visual sentiment ontology and detectors using adjective noun pairs}.
In: \bbtitle{Proceedings of the 21st ACM International Conference on Multimedia},
pp. \bfpage{223}--\blpage{232}
(\byear{2013})
\end{bchapter}
\endbibitem

\bibitem[\protect\citeauthoryear{Zhao et~al.}{2016}]{zhao2016predicting}
\begin{bchapter}
\bauthor{\bsnm{Zhao}, \binits{S.}},
\bauthor{\bsnm{Yao}, \binits{H.}},
\bauthor{\bsnm{Gao}, \binits{Y.}},
\bauthor{\bsnm{Ji}, \binits{R.}},
\bauthor{\bsnm{Xie}, \binits{W.}},
\bauthor{\bsnm{Jiang}, \binits{X.}},
\bauthor{\bsnm{Chua}, \binits{T.-S.}}:
\bctitle{Predicting personalized emotion perceptions of social images}.
In: \bbtitle{Proceedings of the 24th ACM International Conference on Multimedia},
pp. \bfpage{1385}--\blpage{1394}
(\byear{2016})
\end{bchapter}
\endbibitem

\bibitem[\protect\citeauthoryear{Yang et~al.}{2013}]{yang2013user}
\begin{bchapter}
\bauthor{\bsnm{Yang}, \binits{Y.}},
\bauthor{\bsnm{Cui}, \binits{P.}},
\bauthor{\bsnm{Zhu}, \binits{W.}},
\bauthor{\bsnm{Yang}, \binits{S.}}:
\bctitle{User interest and social influence based emotion prediction for individuals}.
In: \bbtitle{Proceedings of the 21st ACM International Conference on Multimedia},
pp. \bfpage{785}--\blpage{788}
(\byear{2013})
\end{bchapter}
\endbibitem

\bibitem[\protect\citeauthoryear{Peng et~al.}{2015}]{peng2015mixed}
\begin{bchapter}
\bauthor{\bsnm{Peng}, \binits{K.-C.}},
\bauthor{\bsnm{Chen}, \binits{T.}},
\bauthor{\bsnm{Sadovnik}, \binits{A.}},
\bauthor{\bsnm{Gallagher}, \binits{A.C.}}:
\bctitle{A mixed bag of emotions: Model, predict, and transfer emotion distributions}.
In: \bbtitle{Proceedings of the IEEE Conference on Computer Vision and Pattern Recognition},
pp. \bfpage{860}--\blpage{868}
(\byear{2015})
\end{bchapter}
\endbibitem

\bibitem[\protect\citeauthoryear{Zhao et~al.}{2015}]{zhao2015predicting}
\begin{bchapter}
\bauthor{\bsnm{Zhao}, \binits{S.}},
\bauthor{\bsnm{Yao}, \binits{H.}},
\bauthor{\bsnm{Jiang}, \binits{X.}},
\bauthor{\bsnm{Sun}, \binits{X.}}:
\bctitle{Predicting discrete probability distribution of image emotions}.
In: \bbtitle{2015 IEEE International Conference on Image Processing (ICIP)},
pp. \bfpage{2459}--\blpage{2463}
(\byear{2015}).
\bcomment{IEEE}
\end{bchapter}
\endbibitem

\bibitem[\protect\citeauthoryear{Zhao et~al.}{2018}]{zhao2018emotiongan}
\begin{bchapter}
\bauthor{\bsnm{Zhao}, \binits{S.}},
\bauthor{\bsnm{Zhao}, \binits{X.}},
\bauthor{\bsnm{Ding}, \binits{G.}},
\bauthor{\bsnm{Keutzer}, \binits{K.}}:
\bctitle{Emotiongan: Unsupervised domain adaptation for learning discrete probability distributions of image emotions}.
In: \bbtitle{Proceedings of the 26th ACM International Conference on Multimedia},
pp. \bfpage{1319}--\blpage{1327}
(\byear{2018})
\end{bchapter}
\endbibitem

\bibitem[\protect\citeauthoryear{Zhao et~al.}{2019}]{zhao2019cycleemotiongan}
\begin{bchapter}
\bauthor{\bsnm{Zhao}, \binits{S.}},
\bauthor{\bsnm{Lin}, \binits{C.}},
\bauthor{\bsnm{Xu}, \binits{P.}},
\bauthor{\bsnm{Zhao}, \binits{S.}},
\bauthor{\bsnm{Guo}, \binits{Y.}},
\bauthor{\bsnm{Krishna}, \binits{R.}},
\bauthor{\bsnm{Ding}, \binits{G.}},
\bauthor{\bsnm{Keutzer}, \binits{K.}}:
\bctitle{Cycleemotiongan: Emotional semantic consistency preserved cyclegan for adapting image emotions}.
In: \bbtitle{Proceedings of the AAAI Conference on Artificial Intelligence},
vol. \bseriesno{33},
pp. \bfpage{2620}--\blpage{2627}
(\byear{2019})
\end{bchapter}
\endbibitem

\bibitem[\protect\citeauthoryear{Zhan et~al.}{2019}]{zhan2019zero}
\begin{bchapter}
\bauthor{\bsnm{Zhan}, \binits{C.}},
\bauthor{\bsnm{She}, \binits{D.}},
\bauthor{\bsnm{Zhao}, \binits{S.}},
\bauthor{\bsnm{Cheng}, \binits{M.-M.}},
\bauthor{\bsnm{Yang}, \binits{J.}}:
\bctitle{Zero-shot emotion recognition via affective structural embedding}.
In: \bbtitle{Proceedings of the IEEE/CVF International Conference on Computer Vision},
pp. \bfpage{1151}--\blpage{1160}
(\byear{2019})
\end{bchapter}
\endbibitem

\bibitem[\protect\citeauthoryear{Bargal et~al.}{2016}]{bargal2016emotion}
\begin{bchapter}
\bauthor{\bsnm{Bargal}, \binits{S.A.}},
\bauthor{\bsnm{Barsoum}, \binits{E.}},
\bauthor{\bsnm{Ferrer}, \binits{C.C.}},
\bauthor{\bsnm{Zhang}, \binits{C.}}:
\bctitle{Emotion recognition in the wild from videos using images}.
In: \bbtitle{Proceedings of the 18th ACM International Conference on Multimodal Interaction},
pp. \bfpage{433}--\blpage{436}
(\byear{2016})
\end{bchapter}
\endbibitem

\bibitem[\protect\citeauthoryear{Hu et~al.}{2019}]{hu2019video}
\begin{barticle}
\bauthor{\bsnm{Hu}, \binits{M.}},
\bauthor{\bsnm{Wang}, \binits{H.}},
\bauthor{\bsnm{Wang}, \binits{X.}},
\bauthor{\bsnm{Yang}, \binits{J.}},
\bauthor{\bsnm{Wang}, \binits{R.}}:
\batitle{Video facial emotion recognition based on local enhanced motion history image and cnn-ctslstm networks}.
\bjtitle{Journal of Visual Communication and Image Representation}
\bvolume{59},
\bfpage{176}--\blpage{185}
(\byear{2019})
\end{barticle}
\endbibitem

\bibitem[\protect\citeauthoryear{Keshari and Palaniswamy}{2019}]{keshari2019emotion}
\begin{bchapter}
\bauthor{\bsnm{Keshari}, \binits{T.}},
\bauthor{\bsnm{Palaniswamy}, \binits{S.}}:
\bctitle{Emotion recognition using feature-level fusion of facial expressions and body gestures}.
In: \bbtitle{2019 International Conference on Communication and Electronics Systems (ICCES)},
pp. \bfpage{1184}--\blpage{1189}
(\byear{2019}).
\bcomment{IEEE}
\end{bchapter}
\endbibitem

\bibitem[\protect\citeauthoryear{Samadiani et~al.}{2022}]{samadiani2022multiple}
\begin{barticle}
\bauthor{\bsnm{Samadiani}, \binits{N.}},
\bauthor{\bsnm{Huang}, \binits{G.}},
\bauthor{\bsnm{Luo}, \binits{W.}},
\bauthor{\bsnm{Chi}, \binits{C.-H.}},
\bauthor{\bsnm{Shu}, \binits{Y.}},
\bauthor{\bsnm{Wang}, \binits{R.}},
\bauthor{\bsnm{Kocaturk}, \binits{T.}}:
\batitle{A multiple feature fusion framework for video emotion recognition in the wild}.
\bjtitle{Concurrency and Computation: Practice and Experience}
\bvolume{34}(\bissue{8}),
\bfpage{5764}
(\byear{2022})
\end{barticle}
\endbibitem

\bibitem[\protect\citeauthoryear{Noroozi et~al.}{2017}]{noroozi2017audio}
\begin{barticle}
\bauthor{\bsnm{Noroozi}, \binits{F.}},
\bauthor{\bsnm{Marjanovic}, \binits{M.}},
\bauthor{\bsnm{Njegus}, \binits{A.}},
\bauthor{\bsnm{Escalera}, \binits{S.}},
\bauthor{\bsnm{Anbarjafari}, \binits{G.}}:
\batitle{Audio-visual emotion recognition in video clips}.
\bjtitle{IEEE Transactions on Affective Computing}
\bvolume{10}(\bissue{1}),
\bfpage{60}--\blpage{75}
(\byear{2017})
\end{barticle}
\endbibitem

\bibitem[\protect\citeauthoryear{Avots et~al.}{2019}]{avots2019audiovisual}
\begin{barticle}
\bauthor{\bsnm{Avots}, \binits{E.}},
\bauthor{\bsnm{Sapi{\'n}ski}, \binits{T.}},
\bauthor{\bsnm{Bachmann}, \binits{M.}},
\bauthor{\bsnm{Kami{\'n}ska}, \binits{D.}}:
\batitle{Audiovisual emotion recognition in wild}.
\bjtitle{Machine Vision and Applications}
\bvolume{30}(\bissue{5}),
\bfpage{975}--\blpage{985}
(\byear{2019})
\end{barticle}
\endbibitem

\bibitem[\protect\citeauthoryear{Zhou et~al.}{2020}]{zhou2020hi}
\begin{barticle}
\bauthor{\bsnm{Zhou}, \binits{T.}},
\bauthor{\bsnm{Fu}, \binits{H.}},
\bauthor{\bsnm{Chen}, \binits{G.}},
\bauthor{\bsnm{Shen}, \binits{J.}},
\bauthor{\bsnm{Shao}, \binits{L.}}:
\batitle{Hi-net: hybrid-fusion network for multi-modal mr image synthesis}.
\bjtitle{IEEE transactions on medical imaging}
\bvolume{39}(\bissue{9}),
\bfpage{2772}--\blpage{2781}
(\byear{2020})
\end{barticle}
\endbibitem

\bibitem[\protect\citeauthoryear{Giachanou and Crestani}{2016}]{giachanou2016like}
\begin{barticle}
\bauthor{\bsnm{Giachanou}, \binits{A.}},
\bauthor{\bsnm{Crestani}, \binits{F.}}:
\batitle{Like it or not: A survey of twitter sentiment analysis methods}.
\bjtitle{ACM Computing Surveys (CSUR)}
\bvolume{49}(\bissue{2}),
\bfpage{1}--\blpage{41}
(\byear{2016})
\end{barticle}
\endbibitem

\bibitem[\protect\citeauthoryear{Zhang et~al.}{2018}]{zhang2018deep}
\begin{barticle}
\bauthor{\bsnm{Zhang}, \binits{L.}},
\bauthor{\bsnm{Wang}, \binits{S.}},
\bauthor{\bsnm{Liu}, \binits{B.}}:
\batitle{Deep learning for sentiment analysis: A survey}.
\bjtitle{Wiley Interdisciplinary Reviews: Data Mining and Knowledge Discovery}
\bvolume{8}(\bissue{4}),
\bfpage{1253}
(\byear{2018})
\end{barticle}
\endbibitem

\bibitem[\protect\citeauthoryear{Schuller et~al.}{2012}]{schuller2012automatic}
\begin{bchapter}
\bauthor{\bsnm{Schuller}, \binits{B.}},
\bauthor{\bsnm{Hantke}, \binits{S.}},
\bauthor{\bsnm{Weninger}, \binits{F.}},
\bauthor{\bsnm{Han}, \binits{W.}},
\bauthor{\bsnm{Zhang}, \binits{Z.}},
\bauthor{\bsnm{Narayanan}, \binits{S.}}:
\bctitle{Automatic recognition of emotion evoked by general sound events}.
In: \bbtitle{2012 IEEE International Conference on Acoustics, Speech and Signal Processing (ICASSP)},
pp. \bfpage{341}--\blpage{344}
(\byear{2012}).
\bcomment{IEEE}
\end{bchapter}
\endbibitem

\bibitem[\protect\citeauthoryear{Schuller et~al.}{2010}]{schuller2010mister}
\begin{barticle}
\bauthor{\bsnm{Schuller}, \binits{B.}},
\bauthor{\bsnm{Hage}, \binits{C.}},
\bauthor{\bsnm{Schuller}, \binits{D.}},
\bauthor{\bsnm{Rigoll}, \binits{G.}}:
\batitle{‘mister dj, cheer me up!’: Musical and textual features for automatic mood classification}.
\bjtitle{Journal of New Music Research}
\bvolume{39}(\bissue{1}),
\bfpage{13}--\blpage{34}
(\byear{2010})
\end{barticle}
\endbibitem

\bibitem[\protect\citeauthoryear{Yang and Chen}{2012}]{yang2012machine}
\begin{barticle}
\bauthor{\bsnm{Yang}, \binits{Y.-H.}},
\bauthor{\bsnm{Chen}, \binits{H.H.}}:
\batitle{Machine recognition of music emotion: A review}.
\bjtitle{ACM Transactions on Intelligent Systems and Technology (TIST)}
\bvolume{3}(\bissue{3}),
\bfpage{1}--\blpage{30}
(\byear{2012})
\end{barticle}
\endbibitem

\bibitem[\protect\citeauthoryear{Zhao et~al.}{2021}]{zhao2021affective}
\begin{barticle}
\bauthor{\bsnm{Zhao}, \binits{S.}},
\bauthor{\bsnm{Yao}, \binits{X.}},
\bauthor{\bsnm{Yang}, \binits{J.}},
\bauthor{\bsnm{Jia}, \binits{G.}},
\bauthor{\bsnm{Ding}, \binits{G.}},
\bauthor{\bsnm{Chua}, \binits{T.-S.}},
\bauthor{\bsnm{Schuller}, \binits{B.W.}},
\bauthor{\bsnm{Keutzer}, \binits{K.}}:
\batitle{Affective image content analysis: Two decades review and new perspectives}.
\bjtitle{IEEE Transactions on Pattern Analysis and Machine Intelligence}
\bvolume{44}(\bissue{10}),
\bfpage{6729}--\blpage{6751}
(\byear{2021})
\end{barticle}
\endbibitem

\bibitem[\protect\citeauthoryear{Pantic and Patras}{2006}]{pantic2006dynamics}
\begin{barticle}
\bauthor{\bsnm{Pantic}, \binits{M.}},
\bauthor{\bsnm{Patras}, \binits{I.}}:
\batitle{Dynamics of facial expression: recognition of facial actions and their temporal segments from face profile image sequences}.
\bjtitle{IEEE Transactions on Systems, Man, and Cybernetics, Part B (Cybernetics)}
\bvolume{36}(\bissue{2}),
\bfpage{433}--\blpage{449}
(\byear{2006})
\end{barticle}
\endbibitem

\bibitem[\protect\citeauthoryear{Sariyanidi et~al.}{2014}]{sariyanidi2014automatic}
\begin{barticle}
\bauthor{\bsnm{Sariyanidi}, \binits{E.}},
\bauthor{\bsnm{Gunes}, \binits{H.}},
\bauthor{\bsnm{Cavallaro}, \binits{A.}}:
\batitle{Automatic analysis of facial affect: A survey of registration, representation, and recognition}.
\bjtitle{IEEE transactions on pattern analysis and machine intelligence}
\bvolume{37}(\bissue{6}),
\bfpage{1113}--\blpage{1133}
(\byear{2014})
\end{barticle}
\endbibitem

\bibitem[\protect\citeauthoryear{Hassan et~al.}{2019}]{hassan2019automatic}
\begin{barticle}
\bauthor{\bsnm{Hassan}, \binits{T.}},
\bauthor{\bsnm{Seu{\ss}}, \binits{D.}},
\bauthor{\bsnm{Wollenberg}, \binits{J.}},
\bauthor{\bsnm{Weitz}, \binits{K.}},
\bauthor{\bsnm{Kunz}, \binits{M.}},
\bauthor{\bsnm{Lautenbacher}, \binits{S.}},
\bauthor{\bsnm{Garbas}, \binits{J.-U.}},
\bauthor{\bsnm{Schmid}, \binits{U.}}:
\batitle{Automatic detection of pain from facial expressions: a survey}.
\bjtitle{IEEE transactions on pattern analysis and machine intelligence}
\bvolume{43}(\bissue{6}),
\bfpage{1815}--\blpage{1831}
(\byear{2019})
\end{barticle}
\endbibitem

\bibitem[\protect\citeauthoryear{Alarcao and Fonseca}{2017}]{alarcao2017emotions}
\begin{barticle}
\bauthor{\bsnm{Alarcao}, \binits{S.M.}},
\bauthor{\bsnm{Fonseca}, \binits{M.J.}}:
\batitle{Emotions recognition using eeg signals: A survey}.
\bjtitle{IEEE Transactions on Affective Computing}
\bvolume{10}(\bissue{3}),
\bfpage{374}--\blpage{393}
(\byear{2017})
\end{barticle}
\endbibitem

\bibitem[\protect\citeauthoryear{Zhao et~al.}{2019}]{zhao2019personalized}
\begin{barticle}
\bauthor{\bsnm{Zhao}, \binits{S.}},
\bauthor{\bsnm{Gholaminejad}, \binits{A.}},
\bauthor{\bsnm{Ding}, \binits{G.}},
\bauthor{\bsnm{Gao}, \binits{Y.}},
\bauthor{\bsnm{Han}, \binits{J.}},
\bauthor{\bsnm{Keutzer}, \binits{K.}}:
\batitle{Personalized emotion recognition by personality-aware high-order learning of physiological signals}.
\bjtitle{ACM Transactions on Multimedia Computing, Communications, and Applications (TOMM)}
\bvolume{15}(\bissue{1s}),
\bfpage{1}--\blpage{18}
(\byear{2019})
\end{barticle}
\endbibitem

\bibitem[\protect\citeauthoryear{Minsky}{1988}]{minsky1988society}
\begin{botherref}
\oauthor{\bsnm{Minsky}, \binits{M.}}:
The Society of Mind,
Simon and Schuster
(1988)
\end{botherref}
\endbibitem

\bibitem[\protect\citeauthoryear{Parrott}{2001}]{parrott2001emotions}
\begin{botherref}
\oauthor{\bsnm{Parrott}, \binits{W.G.}}:
Emotions in Social Psychology: Essential Readings,
psychology press
(2001)
\end{botherref}
\endbibitem

\bibitem[\protect\citeauthoryear{Lang}{1995}]{lang1995emotion}
\begin{barticle}
\bauthor{\bsnm{Lang}, \binits{P.J.}}:
\batitle{The emotion probe: Studies of motivation and attention.}
\bjtitle{American psychologist}
\bvolume{50}(\bissue{5}),
\bfpage{372}
(\byear{1995})
\end{barticle}
\endbibitem

\bibitem[\protect\citeauthoryear{Russell}{1980}]{russell1980circumplex}
\begin{barticle}
\bauthor{\bsnm{Russell}, \binits{J.A.}}:
\batitle{A circumplex model of affect.}
\bjtitle{Journal of personality and social psychology}
\bvolume{39}(\bissue{6}),
\bfpage{1161}
(\byear{1980})
\end{barticle}
\endbibitem

\bibitem[\protect\citeauthoryear{Engen et~al.}{1958}]{engen1958dimensional}
\begin{barticle}
\bauthor{\bsnm{Engen}, \binits{T.}},
\bauthor{\bsnm{Levy}, \binits{N.}},
\bauthor{\bsnm{Schlosberg}, \binits{H.}}:
\batitle{The dimensional analysis of a new series of facial expressions.}
\bjtitle{Journal of Experimental Psychology}
\bvolume{55}(\bissue{5}),
\bfpage{454}
(\byear{1958})
\end{barticle}
\endbibitem

\bibitem[\protect\citeauthoryear{Busso et~al.}{2008}]{busso2008iemocap}
\begin{barticle}
\bauthor{\bsnm{Busso}, \binits{C.}},
\bauthor{\bsnm{Bulut}, \binits{M.}},
\bauthor{\bsnm{Lee}, \binits{C.-C.}},
\bauthor{\bsnm{Kazemzadeh}, \binits{A.}},
\bauthor{\bsnm{Mower}, \binits{E.}},
\bauthor{\bsnm{Kim}, \binits{S.}},
\bauthor{\bsnm{Chang}, \binits{J.N.}},
\bauthor{\bsnm{Lee}, \binits{S.}},
\bauthor{\bsnm{Narayanan}, \binits{S.S.}}:
\batitle{Iemocap: Interactive emotional dyadic motion capture database}.
\bjtitle{Language resources and evaluation}
\bvolume{42},
\bfpage{335}--\blpage{359}
(\byear{2008})
\end{barticle}
\endbibitem

\bibitem[\protect\citeauthoryear{Morency et~al.}{2011}]{morency2011towards}
\begin{bchapter}
\bauthor{\bsnm{Morency}, \binits{L.-P.}},
\bauthor{\bsnm{Mihalcea}, \binits{R.}},
\bauthor{\bsnm{Doshi}, \binits{P.}}:
\bctitle{Towards multimodal sentiment analysis: Harvesting opinions from the web}.
In: \bbtitle{Proc. 13th Int. Conf. Multimodal Interfaces(ICMI)},
pp. \bfpage{169}--\blpage{176}
(\byear{2011})
\end{bchapter}
\endbibitem

\bibitem[\protect\citeauthoryear{Ellis et~al.}{2014}]{ellis2014we}
\begin{bchapter}
\bauthor{\bsnm{Ellis}, \binits{J.G.}},
\bauthor{\bsnm{Jou}, \binits{B.}},
\bauthor{\bsnm{Chang}, \binits{S.-F.}}:
\bctitle{Why we watch the news: a dataset for exploring sentiment in broadcast video news}.
In: \bbtitle{Proceedings of the 16th International Conference on Multimodal Interaction},
pp. \bfpage{104}--\blpage{111}
(\byear{2014})
\end{bchapter}
\endbibitem

\bibitem[\protect\citeauthoryear{W{\"o}llmer et~al.}{2013}]{wollmer2013youtube}
\begin{barticle}
\bauthor{\bsnm{W{\"o}llmer}, \binits{M.}},
\bauthor{\bsnm{Weninger}, \binits{F.}},
\bauthor{\bsnm{Knaup}, \binits{T.}},
\bauthor{\bsnm{Schuller}, \binits{B.}},
\bauthor{\bsnm{Sun}, \binits{C.}},
\bauthor{\bsnm{Sagae}, \binits{K.}},
\bauthor{\bsnm{Morency}, \binits{L.-P.}}:
\batitle{Youtube movie reviews: Sentiment analysis in an audio-visual context}.
\bjtitle{IEEE Intelligent Systems}
\bvolume{28}(\bissue{3}),
\bfpage{46}--\blpage{53}
(\byear{2013})
\end{barticle}
\endbibitem

\bibitem[\protect\citeauthoryear{P{\'e}rez-Rosas et~al.}{2013}]{perez2013utterance}
\begin{bchapter}
\bauthor{\bsnm{P{\'e}rez-Rosas}, \binits{V.}},
\bauthor{\bsnm{Mihalcea}, \binits{R.}},
\bauthor{\bsnm{Morency}, \binits{L.-P.}}:
\bctitle{Utterance-level multimodal sentiment analysis}.
In: \bbtitle{Proceedings of the 51st Annual Meeting of the Association for Computational Linguistics (Volume 1: Long Papers)},
pp. \bfpage{973}--\blpage{982}
(\byear{2013})
\end{bchapter}
\endbibitem

\bibitem[\protect\citeauthoryear{Yu et~al.}{2020}]{yu2020ch}
\begin{bchapter}
\bauthor{\bsnm{Yu}, \binits{W.}},
\bauthor{\bsnm{Xu}, \binits{H.}},
\bauthor{\bsnm{Meng}, \binits{F.}},
\bauthor{\bsnm{Zhu}, \binits{Y.}},
\bauthor{\bsnm{Ma}, \binits{Y.}},
\bauthor{\bsnm{Wu}, \binits{J.}},
\bauthor{\bsnm{Zou}, \binits{J.}},
\bauthor{\bsnm{Yang}, \binits{K.}}:
\bctitle{Ch-sims: A chinese multimodal sentiment analysis dataset with fine-grained annotation of modality}.
In: \bbtitle{Proceedings of the 58th Annual Meeting of the Association for Computational Linguistics},
pp. \bfpage{3718}--\blpage{3727}
(\byear{2020})
\end{bchapter}
\endbibitem

\bibitem[\protect\citeauthoryear{Poria et~al.}{2018}]{poria2018meld}
\begin{botherref}
\oauthor{\bsnm{Poria}, \binits{S.}},
\oauthor{\bsnm{Hazarika}, \binits{D.}},
\oauthor{\bsnm{Majumder}, \binits{N.}},
\oauthor{\bsnm{Naik}, \binits{G.}},
\oauthor{\bsnm{Cambria}, \binits{E.}},
\oauthor{\bsnm{Mihalcea}, \binits{R.}}:
Meld: A multimodal multi-party dataset for emotion recognition in conversations.
arXiv preprint arXiv:1810.02508
(2018)
\end{botherref}
\endbibitem

\bibitem[\protect\citeauthoryear{Zadeh et~al.}{2016}]{zadeh2016multimodal}
\begin{barticle}
\bauthor{\bsnm{Zadeh}, \binits{A.}},
\bauthor{\bsnm{Zellers}, \binits{R.}},
\bauthor{\bsnm{Pincus}, \binits{E.}},
\bauthor{\bsnm{Morency}, \binits{L.-P.}}:
\batitle{Multimodal sentiment intensity analysis in videos: Facial gestures and verbal messages}.
\bjtitle{IEEE Intelligent Systems}
\bvolume{31}(\bissue{6}),
\bfpage{82}--\blpage{88}
(\byear{2016})
\end{barticle}
\endbibitem

\bibitem[\protect\citeauthoryear{Zadeh et~al.}{2018}]{zadeh2018multimodal}
\begin{bchapter}
\bauthor{\bsnm{Zadeh}, \binits{A.B.}},
\bauthor{\bsnm{Liang}, \binits{P.P.}},
\bauthor{\bsnm{Poria}, \binits{S.}},
\bauthor{\bsnm{Cambria}, \binits{E.}},
\bauthor{\bsnm{Morency}, \binits{L.-P.}}:
\bctitle{Multimodal language analysis in the wild: Cmu-mosei dataset and interpretable dynamic fusion graph}.
In: \bbtitle{Proceedings of the 56th Annual Meeting of the Association for Computational Linguistics (Volume 1: Long Papers)},
pp. \bfpage{2236}--\blpage{2246}
(\byear{2018})
\end{bchapter}
\endbibitem

\bibitem[\protect\citeauthoryear{Martin et~al.}{2006}]{martin2006enterface}
\begin{bchapter}
\bauthor{\bsnm{Martin}, \binits{O.}},
\bauthor{\bsnm{Kotsia}, \binits{I.}},
\bauthor{\bsnm{Macq}, \binits{B.}},
\bauthor{\bsnm{Pitas}, \binits{I.}}:
\bctitle{The enterface'05 audio-visual emotion database}.
In: \bbtitle{22nd International Conference on Data Engineering Workshops (ICDEW'06)},
pp. \bfpage{8}--\blpage{8}
(\byear{2006}).
\bcomment{IEEE}
\end{bchapter}
\endbibitem

\bibitem[\protect\citeauthoryear{Baveye et~al.}{2015}]{baveye2015liris}
\begin{barticle}
\bauthor{\bsnm{Baveye}, \binits{Y.}},
\bauthor{\bsnm{Dellandrea}, \binits{E.}},
\bauthor{\bsnm{Chamaret}, \binits{C.}},
\bauthor{\bsnm{Chen}, \binits{L.}}:
\batitle{Liris-accede: A video database for affective content analysis}.
\bjtitle{IEEE Transactions on Affective Computing}
\bvolume{6}(\bissue{1}),
\bfpage{43}--\blpage{55}
(\byear{2015})
\end{barticle}
\endbibitem

\bibitem[\protect\citeauthoryear{Jiang et~al.}{2014}]{jiang2014predicting}
\begin{bchapter}
\bauthor{\bsnm{Jiang}, \binits{Y.-G.}},
\bauthor{\bsnm{Xu}, \binits{B.}},
\bauthor{\bsnm{Xue}, \binits{X.}}:
\bctitle{Predicting emotions in user-generated videos}.
In: \bbtitle{Proceedings of the AAAI Conference on Artificial Intelligence},
vol. \bseriesno{28}
(\byear{2014})
\end{bchapter}
\endbibitem

\bibitem[\protect\citeauthoryear{Xu et~al.}{2016}]{xu2016heterogeneous}
\begin{barticle}
\bauthor{\bsnm{Xu}, \binits{B.}},
\bauthor{\bsnm{Fu}, \binits{Y.}},
\bauthor{\bsnm{Jiang}, \binits{Y.-G.}},
\bauthor{\bsnm{Li}, \binits{B.}},
\bauthor{\bsnm{Sigal}, \binits{L.}}:
\batitle{Heterogeneous knowledge transfer in video emotion recognition, attribution and summarization}.
\bjtitle{IEEE Transactions on Affective Computing}
\bvolume{9}(\bissue{2}),
\bfpage{255}--\blpage{270}
(\byear{2016})
\end{barticle}
\endbibitem

\bibitem[\protect\citeauthoryear{Kossaifi et~al.}{2019}]{kossaifi2019sewa}
\begin{barticle}
\bauthor{\bsnm{Kossaifi}, \binits{J.}},
\bauthor{\bsnm{Walecki}, \binits{R.}},
\bauthor{\bsnm{Panagakis}, \binits{Y.}},
\bauthor{\bsnm{Shen}, \binits{J.}},
\bauthor{\bsnm{Schmitt}, \binits{M.}},
\bauthor{\bsnm{Ringeval}, \binits{F.}},
\bauthor{\bsnm{Han}, \binits{J.}},
\bauthor{\bsnm{Pandit}, \binits{V.}},
\bauthor{\bsnm{Toisoul}, \binits{A.}},
\bauthor{\bsnm{Schuller}, \binits{B.}}, \betal:
\batitle{Sewa db: A rich database for audio-visual emotion and sentiment research in the wild}.
\bjtitle{IEEE transactions on pattern analysis and machine intelligence}
\bvolume{43}(\bissue{3}),
\bfpage{1022}--\blpage{1040}
(\byear{2019})
\end{barticle}
\endbibitem

\bibitem[\protect\citeauthoryear{Valstar et~al.}{2010}]{valstar2010induced}
\begin{bchapter}
\bauthor{\bsnm{Valstar}, \binits{M.}},
\bauthor{\bsnm{Pantic}, \binits{M.}}, \betal:
\bctitle{Induced disgust, happiness and surprise: an addition to the mmi facial expression database}.
In: \bbtitle{Proc. 3rd Intern. Workshop on EMOTION (satellite of LREC): Corpora for Research on Emotion and Affect},
vol. \bseriesno{10},
p. \bfpage{65}
(\byear{2010}).
\bcomment{Paris, France.}
\end{bchapter}
\endbibitem

\bibitem[\protect\citeauthoryear{Mollahosseini et~al.}{2016}]{mollahosseini2016facial}
\begin{bchapter}
\bauthor{\bsnm{Mollahosseini}, \binits{A.}},
\bauthor{\bsnm{Hasani}, \binits{B.}},
\bauthor{\bsnm{Salvador}, \binits{M.J.}},
\bauthor{\bsnm{Abdollahi}, \binits{H.}},
\bauthor{\bsnm{Chan}, \binits{D.}},
\bauthor{\bsnm{Mahoor}, \binits{M.H.}}:
\bctitle{Facial expression recognition from world wild web}.
In: \bbtitle{Proceedings of the IEEE Conference on Computer Vision and Pattern Recognition Workshops},
pp. \bfpage{58}--\blpage{65}
(\byear{2016})
\end{bchapter}
\endbibitem

\bibitem[\protect\citeauthoryear{Fan et~al.}{2016}]{fan2016video}
\begin{bchapter}
\bauthor{\bsnm{Fan}, \binits{Y.}},
\bauthor{\bsnm{Lu}, \binits{X.}},
\bauthor{\bsnm{Li}, \binits{D.}},
\bauthor{\bsnm{Liu}, \binits{Y.}}:
\bctitle{Video-based emotion recognition using cnn-rnn and c3d hybrid networks}.
In: \bbtitle{Proceedings of the 18th ACM International Conference on Multimodal Interaction},
pp. \bfpage{445}--\blpage{450}
(\byear{2016})
\end{bchapter}
\endbibitem

\bibitem[\protect\citeauthoryear{Xue et~al.}{2022}]{xue2022coarse}
\begin{bchapter}
\bauthor{\bsnm{Xue}, \binits{F.}},
\bauthor{\bsnm{Tan}, \binits{Z.}},
\bauthor{\bsnm{Zhu}, \binits{Y.}},
\bauthor{\bsnm{Ma}, \binits{Z.}},
\bauthor{\bsnm{Guo}, \binits{G.}}:
\bctitle{Coarse-to-fine cascaded networks with smooth predicting for video facial expression recognition}.
In: \bbtitle{Proceedings of the IEEE/CVF Conference on Computer Vision and Pattern Recognition},
pp. \bfpage{2412}--\blpage{2418}
(\byear{2022})
\end{bchapter}
\endbibitem

\bibitem[\protect\citeauthoryear{Savchenko et~al.}{2022}]{savchenko2022classifying}
\begin{barticle}
\bauthor{\bsnm{Savchenko}, \binits{A.V.}},
\bauthor{\bsnm{Savchenko}, \binits{L.V.}},
\bauthor{\bsnm{Makarov}, \binits{I.}}:
\batitle{Classifying emotions and engagement in online learning based on a single facial expression recognition neural network}.
\bjtitle{IEEE Transactions on Affective Computing}
\bvolume{13}(\bissue{4}),
\bfpage{2132}--\blpage{2143}
(\byear{2022})
\end{barticle}
\endbibitem

\bibitem[\protect\citeauthoryear{Hernandez-Luquin and Escalante}{2023}]{hernandez2023multi}
\begin{barticle}
\bauthor{\bsnm{Hernandez-Luquin}, \binits{F.}},
\bauthor{\bsnm{Escalante}, \binits{H.J.}}:
\batitle{Multi-branch deep radial basis function networks for facial emotion recognition}.
\bjtitle{Neural Computing and Applications}
\bvolume{35}(\bissue{25}),
\bfpage{18131}--\blpage{18145}
(\byear{2023})
\end{barticle}
\endbibitem

\bibitem[\protect\citeauthoryear{Gavrilescu}{2015}]{gavrilescu2015recognizing}
\begin{bchapter}
\bauthor{\bsnm{Gavrilescu}, \binits{M.}}:
\bctitle{Recognizing emotions from videos by studying facial expressions, body postures and hand gestures}.
In: \bbtitle{2015 23rd Telecommunications Forum Telfor (TELFOR)},
pp. \bfpage{720}--\blpage{723}
(\byear{2015}).
\bcomment{IEEE}
\end{bchapter}
\endbibitem

\bibitem[\protect\citeauthoryear{Shen et~al.}{2019}]{shen2019emotion}
\begin{bchapter}
\bauthor{\bsnm{Shen}, \binits{Z.}},
\bauthor{\bsnm{Cheng}, \binits{J.}},
\bauthor{\bsnm{Hu}, \binits{X.}},
\bauthor{\bsnm{Dong}, \binits{Q.}}:
\bctitle{Emotion recognition based on multi-view body gestures}.
In: \bbtitle{2019 Ieee International Conference on Image Processing (icip)},
pp. \bfpage{3317}--\blpage{3321}
(\byear{2019}).
\bcomment{IEEE}
\end{bchapter}
\endbibitem

\bibitem[\protect\citeauthoryear{Wu et~al.}{2022}]{wu2022generalized}
\begin{botherref}
\oauthor{\bsnm{Wu}, \binits{J.}},
\oauthor{\bsnm{Zhang}, \binits{Y.}},
\oauthor{\bsnm{Sun}, \binits{S.}},
\oauthor{\bsnm{Li}, \binits{Q.}},
\oauthor{\bsnm{Zhao}, \binits{X.}}:
Generalized zero-shot emotion recognition from body gestures.
Applied Intelligence,
1--19
(2022)
\end{botherref}
\endbibitem

\bibitem[\protect\citeauthoryear{Santhoshkumar and Geetha}{2019}]{santhoshkumar2019deep}
\begin{barticle}
\bauthor{\bsnm{Santhoshkumar}, \binits{R.}},
\bauthor{\bsnm{Geetha}, \binits{M.K.}}:
\batitle{Deep learning approach for emotion recognition from human body movements with feedforward deep convolution neural networks}.
\bjtitle{Procedia Computer Science}
\bvolume{152},
\bfpage{158}--\blpage{165}
(\byear{2019})
\end{barticle}
\endbibitem

\bibitem[\protect\citeauthoryear{Liu et~al.}{2021}]{liu2021imigue}
\begin{bchapter}
\bauthor{\bsnm{Liu}, \binits{X.}},
\bauthor{\bsnm{Shi}, \binits{H.}},
\bauthor{\bsnm{Chen}, \binits{H.}},
\bauthor{\bsnm{Yu}, \binits{Z.}},
\bauthor{\bsnm{Li}, \binits{X.}},
\bauthor{\bsnm{Zhao}, \binits{G.}}:
\bctitle{imigue: An identity-free video dataset for micro-gesture understanding and emotion analysis}.
In: \bbtitle{Proceedings of the IEEE/CVF Conference on Computer Vision and Pattern Recognition},
pp. \bfpage{10631}--\blpage{10642}
(\byear{2021})
\end{bchapter}
\endbibitem

\bibitem[\protect\citeauthoryear{Nguyen et~al.}{2018}]{nguyen2018deep}
\begin{barticle}
\bauthor{\bsnm{Nguyen}, \binits{D.}},
\bauthor{\bsnm{Nguyen}, \binits{K.}},
\bauthor{\bsnm{Sridharan}, \binits{S.}},
\bauthor{\bsnm{Dean}, \binits{D.}},
\bauthor{\bsnm{Fookes}, \binits{C.}}:
\batitle{Deep spatio-temporal feature fusion with compact bilinear pooling for multimodal emotion recognition}.
\bjtitle{Computer vision and image understanding}
\bvolume{174},
\bfpage{33}--\blpage{42}
(\byear{2018})
\end{barticle}
\endbibitem

\bibitem[\protect\citeauthoryear{Chen et~al.}{2016}]{chen2016emotion}
\begin{bchapter}
\bauthor{\bsnm{Chen}, \binits{C.}},
\bauthor{\bsnm{Wu}, \binits{Z.}},
\bauthor{\bsnm{Jiang}, \binits{Y.-G.}}:
\bctitle{Emotion in context: Deep semantic feature fusion for video emotion recognition}.
In: \bbtitle{Proceedings of the 24th ACM International Conference on Multimedia},
pp. \bfpage{127}--\blpage{131}
(\byear{2016})
\end{bchapter}
\endbibitem

\bibitem[\protect\citeauthoryear{Ghaleb et~al.}{2017}]{ghaleb2017multimodal}
\begin{bchapter}
\bauthor{\bsnm{Ghaleb}, \binits{E.}},
\bauthor{\bsnm{Popa}, \binits{M.}},
\bauthor{\bsnm{Hortal}, \binits{E.}},
\bauthor{\bsnm{Asteriadis}, \binits{S.}}:
\bctitle{Multimodal fusion based on information gain for emotion recognition in the wild}.
In: \bbtitle{2017 Intelligent Systems Conference (IntelliSys)},
pp. \bfpage{814}--\blpage{823}
(\byear{2017}).
\bcomment{IEEE}
\end{bchapter}
\endbibitem

\bibitem[\protect\citeauthoryear{Qing et~al.}{2023}]{qing2023dvc}
\begin{botherref}
\oauthor{\bsnm{Qing}, \binits{L.}},
\oauthor{\bsnm{Wen}, \binits{H.}},
\oauthor{\bsnm{Chen}, \binits{H.}},
\oauthor{\bsnm{Jin}, \binits{R.}},
\oauthor{\bsnm{Cheng}, \binits{Y.}},
\oauthor{\bsnm{Peng}, \binits{Y.}}:
Dvc-net: a new dual-view context-aware network for emotion recognition in the wild.
Neural Computing and Applications,
1--13
(2023)
\end{botherref}
\endbibitem

\bibitem[\protect\citeauthoryear{Kim and Provost}{2017}]{kim2017isla}
\begin{barticle}
\bauthor{\bsnm{Kim}, \binits{Y.}},
\bauthor{\bsnm{Provost}, \binits{E.M.}}:
\batitle{Isla: Temporal segmentation and labeling for audio-visual emotion recognition}.
\bjtitle{IEEE Transactions on affective computing}
\bvolume{10}(\bissue{2}),
\bfpage{196}--\blpage{208}
(\byear{2017})
\end{barticle}
\endbibitem

\bibitem[\protect\citeauthoryear{Xia et~al.}{2022}]{xia2022multimodal}
\begin{barticle}
\bauthor{\bsnm{Xia}, \binits{X.}},
\bauthor{\bsnm{Zhao}, \binits{Y.}},
\bauthor{\bsnm{Jiang}, \binits{D.}}:
\batitle{Multimodal interaction enhanced representation learning for video emotion recognition}.
\bjtitle{Frontiers in Neuroscience}
\bvolume{16},
\bfpage{1086380}
(\byear{2022})
\end{barticle}
\endbibitem

\bibitem[\protect\citeauthoryear{Liu et~al.}{2022}]{liu2022multi}
\begin{barticle}
\bauthor{\bsnm{Liu}, \binits{D.}},
\bauthor{\bsnm{Chen}, \binits{L.}},
\bauthor{\bsnm{Wang}, \binits{L.}},
\bauthor{\bsnm{Wang}, \binits{Z.}}:
\batitle{A multi-modal emotion fusion classification method combined expression and speech based on attention mechanism}.
\bjtitle{Multimedia Tools and Applications}
\bvolume{81}(\bissue{29}),
\bfpage{41677}--\blpage{41695}
(\byear{2022})
\end{barticle}
\endbibitem

\bibitem[\protect\citeauthoryear{Praveen et~al.}{2022}]{praveen2022joint}
\begin{bchapter}
\bauthor{\bsnm{Praveen}, \binits{R.G.}},
\bauthor{\bsnm{Melo}, \binits{W.C.}},
\bauthor{\bsnm{Ullah}, \binits{N.}},
\bauthor{\bsnm{Aslam}, \binits{H.}},
\bauthor{\bsnm{Zeeshan}, \binits{O.}},
\bauthor{\bsnm{Denorme}, \binits{T.}},
\bauthor{\bsnm{Pedersoli}, \binits{M.}},
\bauthor{\bsnm{Koerich}, \binits{A.L.}},
\bauthor{\bsnm{Bacon}, \binits{S.}},
\bauthor{\bsnm{Cardinal}, \binits{P.}}, \betal:
\bctitle{A joint cross-attention model for audio-visual fusion in dimensional emotion recognition}.
In: \bbtitle{Proceedings of the IEEE/CVF Conference on Computer Vision and Pattern Recognition},
pp. \bfpage{2486}--\blpage{2495}
(\byear{2022})
\end{bchapter}
\endbibitem

\bibitem[\protect\citeauthoryear{Zhu et~al.}{2020}]{zhu2020multimodal}
\begin{barticle}
\bauthor{\bsnm{Zhu}, \binits{H.}},
\bauthor{\bsnm{Wang}, \binits{Z.}},
\bauthor{\bsnm{Shi}, \binits{Y.}},
\bauthor{\bsnm{Hua}, \binits{Y.}},
\bauthor{\bsnm{Xu}, \binits{G.}},
\bauthor{\bsnm{Deng}, \binits{L.}}:
\batitle{Multimodal fusion method based on self-attention mechanism}.
\bjtitle{Wireless Communications and Mobile Computing}
\bvolume{2020},
\bfpage{1}--\blpage{8}
(\byear{2020})
\end{barticle}
\endbibitem

\bibitem[\protect\citeauthoryear{Le et~al.}{2022}]{le2022global}
\begin{barticle}
\bauthor{\bsnm{Le}, \binits{N.}},
\bauthor{\bsnm{Nguyen}, \binits{K.}},
\bauthor{\bsnm{Nguyen}, \binits{A.}},
\bauthor{\bsnm{Le}, \binits{B.}}:
\batitle{Global-local attention for emotion recognition}.
\bjtitle{Neural Computing and Applications}
\bvolume{34}(\bissue{24}),
\bfpage{21625}--\blpage{21639}
(\byear{2022})
\end{barticle}
\endbibitem

\bibitem[\protect\citeauthoryear{Amer et~al.}{2018}]{amer2018deep}
\begin{barticle}
\bauthor{\bsnm{Amer}, \binits{M.R.}},
\bauthor{\bsnm{Shields}, \binits{T.}},
\bauthor{\bsnm{Siddiquie}, \binits{B.}},
\bauthor{\bsnm{Tamrakar}, \binits{A.}},
\bauthor{\bsnm{Divakaran}, \binits{A.}},
\bauthor{\bsnm{Chai}, \binits{S.}}:
\batitle{Deep multimodal fusion: A hybrid approach}.
\bjtitle{International Journal of Computer Vision}
\bvolume{126},
\bfpage{440}--\blpage{456}
(\byear{2018})
\end{barticle}
\endbibitem

\bibitem[\protect\citeauthoryear{Huang et~al.}{2020}]{huang2020multimodal}
\begin{bchapter}
\bauthor{\bsnm{Huang}, \binits{J.}},
\bauthor{\bsnm{Tao}, \binits{J.}},
\bauthor{\bsnm{Liu}, \binits{B.}},
\bauthor{\bsnm{Lian}, \binits{Z.}},
\bauthor{\bsnm{Niu}, \binits{M.}}:
\bctitle{Multimodal transformer fusion for continuous emotion recognition}.
In: \bbtitle{ICASSP 2020-2020 IEEE International Conference on Acoustics, Speech and Signal Processing (ICASSP)},
pp. \bfpage{3507}--\blpage{3511}
(\byear{2020}).
\bcomment{IEEE}
\end{bchapter}
\endbibitem

\bibitem[\protect\citeauthoryear{John and Kawanishi}{2022}]{john2022audio}
\begin{bchapter}
\bauthor{\bsnm{John}, \binits{V.}},
\bauthor{\bsnm{Kawanishi}, \binits{Y.}}:
\bctitle{Audio and video-based emotion recognition using multimodal transformers}.
In: \bbtitle{2022 26th International Conference on Pattern Recognition (ICPR)},
pp. \bfpage{2582}--\blpage{2588}
(\byear{2022}).
\bcomment{IEEE}
\end{bchapter}
\endbibitem

\bibitem[\protect\citeauthoryear{Chumachenko et~al.}{2022}]{chumachenko2022self}
\begin{bchapter}
\bauthor{\bsnm{Chumachenko}, \binits{K.}},
\bauthor{\bsnm{Iosifidis}, \binits{A.}},
\bauthor{\bsnm{Gabbouj}, \binits{M.}}:
\bctitle{Self-attention fusion for audiovisual emotion recognition with incomplete data}.
In: \bbtitle{2022 26th International Conference on Pattern Recognition (ICPR)},
pp. \bfpage{2822}--\blpage{2828}
(\byear{2022}).
\bcomment{IEEE}
\end{bchapter}
\endbibitem

\bibitem[\protect\citeauthoryear{Mehrabian and Russell}{1974}]{mehrabian1974approach}
\begin{botherref}
\oauthor{\bsnm{Mehrabian}, \binits{A.}},
\oauthor{\bsnm{Russell}, \binits{J.A.}}:
An Approach to Environmental Psychology.,
the MIT Press
(1974)
\end{botherref}
\endbibitem

\bibitem[\protect\citeauthoryear{Chen et~al.}{2016}]{chen2016supervised}
\begin{bchapter}
\bauthor{\bsnm{Chen}, \binits{D.}},
\bauthor{\bsnm{Hua}, \binits{G.}},
\bauthor{\bsnm{Wen}, \binits{F.}},
\bauthor{\bsnm{Sun}, \binits{J.}}:
\bctitle{Supervised transformer network for efficient face detection}.
In: \bbtitle{Computer Vision--ECCV 2016: 14th European Conference, Amsterdam, The Netherlands, October 11-14, 2016, Proceedings, Part V 14},
pp. \bfpage{122}--\blpage{138}
(\byear{2016}).
\bcomment{Springer}
\end{bchapter}
\endbibitem

\bibitem[\protect\citeauthoryear{Kollias and Zafeiriou}{2019}]{kollias2019expression}
\begin{botherref}
\oauthor{\bsnm{Kollias}, \binits{D.}},
\oauthor{\bsnm{Zafeiriou}, \binits{S.}}:
Expression, affect, action unit recognition: Aff-wild2, multi-task learning and arcface.
arXiv preprint arXiv:1910.04855
(2019)
\end{botherref}
\endbibitem

\bibitem[\protect\citeauthoryear{Luo et~al.}{2020}]{luo2020arbee}
\begin{barticle}
\bauthor{\bsnm{Luo}, \binits{Y.}},
\bauthor{\bsnm{Ye}, \binits{J.}},
\bauthor{\bsnm{Adams}, \binits{R.B.}},
\bauthor{\bsnm{Li}, \binits{J.}},
\bauthor{\bsnm{Newman}, \binits{M.G.}},
\bauthor{\bsnm{Wang}, \binits{J.Z.}}:
\batitle{Arbee: Towards automated recognition of bodily expression of emotion in the wild}.
\bjtitle{International journal of computer vision}
\bvolume{128},
\bfpage{1}--\blpage{25}
(\byear{2020})
\end{barticle}
\endbibitem

\bibitem[\protect\citeauthoryear{De~Gelder et~al.}{2015}]{de2015perception}
\begin{barticle}
\bauthor{\bsnm{De~Gelder}, \binits{B.}},
\bauthor{\bsnm{Borst}, \binits{A.W.}},
\bauthor{\bsnm{Watson}, \binits{R.}}:
\batitle{The perception of emotion in body expressions}.
\bjtitle{Wiley Interdisciplinary Reviews: Cognitive Science}
\bvolume{6}(\bissue{2}),
\bfpage{149}--\blpage{158}
(\byear{2015})
\end{barticle}
\endbibitem

\bibitem[\protect\citeauthoryear{Psaltis et~al.}{2016}]{psaltis2016multimodal}
\begin{bchapter}
\bauthor{\bsnm{Psaltis}, \binits{A.}},
\bauthor{\bsnm{Kaza}, \binits{K.}},
\bauthor{\bsnm{Stefanidis}, \binits{K.}},
\bauthor{\bsnm{Thermos}, \binits{S.}},
\bauthor{\bsnm{Apostolakis}, \binits{K.C.}},
\bauthor{\bsnm{Dimitropoulos}, \binits{K.}},
\bauthor{\bsnm{Daras}, \binits{P.}}:
\bctitle{Multimodal affective state recognition in serious games applications}.
In: \bbtitle{2016 Ieee International Conference on Imaging Systems and Techniques (ist)},
pp. \bfpage{435}--\blpage{439}
(\byear{2016}).
\bcomment{IEEE}
\end{bchapter}
\endbibitem

\bibitem[\protect\citeauthoryear{Sarkar et~al.}{2014}]{sarkar2014feature}
\begin{bchapter}
\bauthor{\bsnm{Sarkar}, \binits{C.}},
\bauthor{\bsnm{Bhatia}, \binits{S.}},
\bauthor{\bsnm{Agarwal}, \binits{A.}},
\bauthor{\bsnm{Li}, \binits{J.}}:
\bctitle{Feature analysis for computational personality recognition using youtube personality data set}.
In: \bbtitle{Proceedings of the 2014 ACM Multi Media on Workshop on Computational Personality Recognition},
pp. \bfpage{11}--\blpage{14}
(\byear{2014})
\end{bchapter}
\endbibitem

\bibitem[\protect\citeauthoryear{Lee and Narayanan}{2005}]{lee2005toward}
\begin{barticle}
\bauthor{\bsnm{Lee}, \binits{C.M.}},
\bauthor{\bsnm{Narayanan}, \binits{S.S.}}:
\batitle{Toward detecting emotions in spoken dialogs}.
\bjtitle{IEEE transactions on speech and audio processing}
\bvolume{13}(\bissue{2}),
\bfpage{293}--\blpage{303}
(\byear{2005})
\end{barticle}
\endbibitem

\bibitem[\protect\citeauthoryear{Dimou et~al.}{2014}]{dimou2014rml}
\begin{botherref}
\oauthor{\bsnm{Dimou}, \binits{A.}},
\oauthor{\bsnm{Vander~Sande}, \binits{M.}},
\oauthor{\bsnm{Colpaert}, \binits{P.}},
\oauthor{\bsnm{Verborgh}, \binits{R.}},
\oauthor{\bsnm{Mannens}, \binits{E.}},
\oauthor{\bsnm{Walle}, \binits{R.}}:
Rml: A generic language for integrated rdf mappings of heterogeneous data.
Ldow
\textbf{1184}
(2014)
\end{botherref}
\endbibitem

\bibitem[\protect\citeauthoryear{Zhalehpour et~al.}{2016}]{zhalehpour2016multimodal}
\begin{barticle}
\bauthor{\bsnm{Zhalehpour}, \binits{S.}},
\bauthor{\bsnm{Akhtar}, \binits{Z.}},
\bauthor{\bsnm{Eroglu~Erdem}, \binits{C.}}:
\batitle{Multimodal emotion recognition based on peak frame selection from video}.
\bjtitle{Signal, Image and Video Processing}
\bvolume{10},
\bfpage{827}--\blpage{834}
(\byear{2016})
\end{barticle}
\endbibitem

\bibitem[\protect\citeauthoryear{Wang et~al.}{2012}]{wang2012kernel}
\begin{barticle}
\bauthor{\bsnm{Wang}, \binits{Y.}},
\bauthor{\bsnm{Guan}, \binits{L.}},
\bauthor{\bsnm{Venetsanopoulos}, \binits{A.N.}}:
\batitle{Kernel cross-modal factor analysis for information fusion with application to bimodal emotion recognition}.
\bjtitle{IEEE Transactions on Multimedia}
\bvolume{14}(\bissue{3}),
\bfpage{597}--\blpage{607}
(\byear{2012})
\end{barticle}
\endbibitem

\bibitem[\protect\citeauthoryear{Wu et~al.}{2014}]{wu2014survey}
\begin{barticle}
\bauthor{\bsnm{Wu}, \binits{C.-H.}},
\bauthor{\bsnm{Lin}, \binits{J.-C.}},
\bauthor{\bsnm{Wei}, \binits{W.-L.}}:
\batitle{Survey on audiovisual emotion recognition: databases, features, and data fusion strategies}.
\bjtitle{APSIPA transactions on signal and information processing}
\bvolume{3},
\bfpage{12}
(\byear{2014})
\end{barticle}
\endbibitem

\bibitem[\protect\citeauthoryear{Zhang et~al.}{2017}]{zhang2017learning}
\begin{barticle}
\bauthor{\bsnm{Zhang}, \binits{S.}},
\bauthor{\bsnm{Zhang}, \binits{S.}},
\bauthor{\bsnm{Huang}, \binits{T.}},
\bauthor{\bsnm{Gao}, \binits{W.}},
\bauthor{\bsnm{Tian}, \binits{Q.}}:
\batitle{Learning affective features with a hybrid deep model for audio--visual emotion recognition}.
\bjtitle{IEEE Transactions on Circuits and Systems for Video Technology}
\bvolume{28}(\bissue{10}),
\bfpage{3030}--\blpage{3043}
(\byear{2017})
\end{barticle}
\endbibitem

\bibitem[\protect\citeauthoryear{Vaswani et~al.}{2017}]{vaswani2017attention}
\begin{botherref}
\oauthor{\bsnm{Vaswani}, \binits{A.}},
\oauthor{\bsnm{Shazeer}, \binits{N.}},
\oauthor{\bsnm{Parmar}, \binits{N.}},
\oauthor{\bsnm{Uszkoreit}, \binits{J.}},
\oauthor{\bsnm{Jones}, \binits{L.}},
\oauthor{\bsnm{Gomez}, \binits{A.N.}},
\oauthor{\bsnm{Kaiser}, \binits{{\L}.}},
\oauthor{\bsnm{Polosukhin}, \binits{I.}}:
Attention is all you need.
Advances in neural information processing systems
\textbf{30}
(2017)
\end{botherref}
\endbibitem

\bibitem[\protect\citeauthoryear{Kahou et~al.}{2016}]{kahou2016emonets}
\begin{barticle}
\bauthor{\bsnm{Kahou}, \binits{S.E.}},
\bauthor{\bsnm{Bouthillier}, \binits{X.}},
\bauthor{\bsnm{Lamblin}, \binits{P.}},
\bauthor{\bsnm{Gulcehre}, \binits{C.}},
\bauthor{\bsnm{Michalski}, \binits{V.}},
\bauthor{\bsnm{Konda}, \binits{K.}},
\bauthor{\bsnm{Jean}, \binits{S.}},
\bauthor{\bsnm{Froumenty}, \binits{P.}},
\bauthor{\bsnm{Dauphin}, \binits{Y.}},
\bauthor{\bsnm{Boulanger-Lewandowski}, \binits{N.}}, \betal:
\batitle{Emonets: Multimodal deep learning approaches for emotion recognition in video}.
\bjtitle{Journal on Multimodal User Interfaces}
\bvolume{10},
\bfpage{99}--\blpage{111}
(\byear{2016})
\end{barticle}
\endbibitem

\bibitem[\protect\citeauthoryear{Chao et~al.}{2015}]{chao2015long}
\begin{bchapter}
\bauthor{\bsnm{Chao}, \binits{L.}},
\bauthor{\bsnm{Tao}, \binits{J.}},
\bauthor{\bsnm{Yang}, \binits{M.}},
\bauthor{\bsnm{Li}, \binits{Y.}},
\bauthor{\bsnm{Wen}, \binits{Z.}}:
\bctitle{Long short term memory recurrent neural network based multimodal dimensional emotion recognition}.
In: \bbtitle{Proceedings of the 5th International Workshop on Audio/visual Emotion Challenge},
pp. \bfpage{65}--\blpage{72}
(\byear{2015})
\end{bchapter}
\endbibitem

\bibitem[\protect\citeauthoryear{Ringeval et~al.}{2015}]{ringeval2015av+}
\begin{bchapter}
\bauthor{\bsnm{Ringeval}, \binits{F.}},
\bauthor{\bsnm{Schuller}, \binits{B.}},
\bauthor{\bsnm{Valstar}, \binits{M.}},
\bauthor{\bsnm{Jaiswal}, \binits{S.}},
\bauthor{\bsnm{Marchi}, \binits{E.}},
\bauthor{\bsnm{Lalanne}, \binits{D.}},
\bauthor{\bsnm{Cowie}, \binits{R.}},
\bauthor{\bsnm{Pantic}, \binits{M.}}:
\bctitle{Av+ ec 2015: The first affect recognition challenge bridging across audio, video, and physiological data}.
In: \bbtitle{Proceedings of the 5th International Workshop on Audio/visual Emotion Challenge},
pp. \bfpage{3}--\blpage{8}
(\byear{2015})
\end{bchapter}
\endbibitem

\bibitem[\protect\citeauthoryear{Wang et~al.}{2019}]{wang2019multi}
\begin{bchapter}
\bauthor{\bsnm{Wang}, \binits{Y.}},
\bauthor{\bsnm{Wu}, \binits{J.}},
\bauthor{\bsnm{Hoashi}, \binits{K.}}:
\bctitle{Multi-attention fusion network for video-based emotion recognition}.
In: \bbtitle{2019 International Conference on Multimodal Interaction},
pp. \bfpage{595}--\blpage{601}
(\byear{2019})
\end{bchapter}
\endbibitem

\bibitem[\protect\citeauthoryear{Du et~al.}{2019}]{du2019spatio}
\begin{barticle}
\bauthor{\bsnm{Du}, \binits{Z.}},
\bauthor{\bsnm{Wu}, \binits{S.}},
\bauthor{\bsnm{Huang}, \binits{D.}},
\bauthor{\bsnm{Li}, \binits{W.}},
\bauthor{\bsnm{Wang}, \binits{Y.}}:
\batitle{Spatio-temporal encoder-decoder fully convolutional network for video-based dimensional emotion recognition}.
\bjtitle{IEEE Transactions on Affective Computing}
\bvolume{12}(\bissue{3}),
\bfpage{565}--\blpage{578}
(\byear{2019})
\end{barticle}
\endbibitem

\bibitem[\protect\citeauthoryear{Ringeval et~al.}{2013}]{ringeval2013introducing}
\begin{bchapter}
\bauthor{\bsnm{Ringeval}, \binits{F.}},
\bauthor{\bsnm{Sonderegger}, \binits{A.}},
\bauthor{\bsnm{Sauer}, \binits{J.}},
\bauthor{\bsnm{Lalanne}, \binits{D.}}:
\bctitle{Introducing the recola multimodal corpus of remote collaborative and affective interactions}.
In: \bbtitle{2013 10th IEEE International Conference and Workshops on Automatic Face and Gesture Recognition (FG)},
pp. \bfpage{1}--\blpage{8}
(\byear{2013}).
\bcomment{IEEE}
\end{bchapter}
\endbibitem

\bibitem[\protect\citeauthoryear{Daoudi et~al.}{2017}]{daoudi2017emotion}
\begin{bchapter}
\bauthor{\bsnm{Daoudi}, \binits{M.}},
\bauthor{\bsnm{Berretti}, \binits{S.}},
\bauthor{\bsnm{Pala}, \binits{P.}},
\bauthor{\bsnm{Delevoye}, \binits{Y.}},
\bauthor{\bsnm{Del~Bimbo}, \binits{A.}}:
\bctitle{Emotion recognition by body movement representation on the manifold of symmetric positive definite matrices}.
In: \bbtitle{Image Analysis and Processing-ICIAP 2017: 19th International Conference, Catania, Italy, September 11-15, 2017, Proceedings, Part I 19},
pp. \bfpage{550}--\blpage{560}
(\byear{2017}).
\bcomment{Springer}
\end{bchapter}
\endbibitem

\bibitem[\protect\citeauthoryear{Hicheur et~al.}{2013}]{hicheur2013combined}
\begin{botherref}
\oauthor{\bsnm{Hicheur}, \binits{H.}},
\oauthor{\bsnm{Kadone}, \binits{H.}},
\oauthor{\bsnm{Gr{\`e}zes}, \binits{J.}},
\oauthor{\bsnm{Berthoz}, \binits{A.}}:
The combined role of motion-related cues and upper body posture for the expression of emotions during human walking.
Modeling, simulation and optimization of bipedal walking,
71--85
(2013)
\end{botherref}
\endbibitem

\bibitem[\protect\citeauthoryear{Wei et~al.}{2020}]{wei2020multimodal}
\begin{bchapter}
\bauthor{\bsnm{Wei}, \binits{G.}},
\bauthor{\bsnm{Jian}, \binits{L.}},
\bauthor{\bsnm{Mo}, \binits{S.}}:
\bctitle{Multimodal (audio, facial and gesture) based emotion recognition challenge}.
In: \bbtitle{2020 15th IEEE International Conference on Automatic Face and Gesture Recognition (FG 2020)},
pp. \bfpage{908}--\blpage{911}
(\byear{2020}).
\bcomment{IEEE}
\end{bchapter}
\endbibitem

\bibitem[\protect\citeauthoryear{Nigam and Dutta}{2022}]{nigam2022emotion}
\begin{botherref}
\oauthor{\bsnm{Nigam}, \binits{N.}},
\oauthor{\bsnm{Dutta}, \binits{T.}}:
Emotion and gesture guided action recognition in videos using supervised deep networks.
IEEE Transactions on Computational Social Systems
(2022)
\end{botherref}
\endbibitem

\bibitem[\protect\citeauthoryear{Kay et~al.}{2017}]{kay2017kinetics}
\begin{botherref}
\oauthor{\bsnm{Kay}, \binits{W.}},
\oauthor{\bsnm{Carreira}, \binits{J.}},
\oauthor{\bsnm{Simonyan}, \binits{K.}},
\oauthor{\bsnm{Zhang}, \binits{B.}},
\oauthor{\bsnm{Hillier}, \binits{C.}},
\oauthor{\bsnm{Vijayanarasimhan}, \binits{S.}},
\oauthor{\bsnm{Viola}, \binits{F.}},
\oauthor{\bsnm{Green}, \binits{T.}},
\oauthor{\bsnm{Back}, \binits{T.}},
\oauthor{\bsnm{Natsev}, \binits{P.}}, et al.:
The kinetics human action video dataset.
arXiv preprint arXiv:1705.06950
(2017)
\end{botherref}
\endbibitem

\bibitem[\protect\citeauthoryear{Zhao et~al.}{2020}]{zhao2020end}
\begin{bchapter}
\bauthor{\bsnm{Zhao}, \binits{S.}},
\bauthor{\bsnm{Ma}, \binits{Y.}},
\bauthor{\bsnm{Gu}, \binits{Y.}},
\bauthor{\bsnm{Yang}, \binits{J.}},
\bauthor{\bsnm{Xing}, \binits{T.}},
\bauthor{\bsnm{Xu}, \binits{P.}},
\bauthor{\bsnm{Hu}, \binits{R.}},
\bauthor{\bsnm{Chai}, \binits{H.}},
\bauthor{\bsnm{Keutzer}, \binits{K.}}:
\bctitle{An end-to-end visual-audio attention network for emotion recognition in user-generated videos}.
In: \bbtitle{Proceedings of the AAAI Conference on Artificial Intelligence},
vol. \bseriesno{34},
pp. \bfpage{303}--\blpage{311}
(\byear{2020})
\end{bchapter}
\endbibitem

\bibitem[\protect\citeauthoryear{Wang et~al.}{2017}]{wang2017emotion}
\begin{bchapter}
\bauthor{\bsnm{Wang}, \binits{S.}},
\bauthor{\bsnm{Wang}, \binits{W.}},
\bauthor{\bsnm{Zhao}, \binits{J.}},
\bauthor{\bsnm{Chen}, \binits{S.}},
\bauthor{\bsnm{Jin}, \binits{Q.}},
\bauthor{\bsnm{Zhang}, \binits{S.}},
\bauthor{\bsnm{Qin}, \binits{Y.}}:
\bctitle{Emotion recognition with multimodal features and temporal models}.
In: \bbtitle{Proceedings of the 19th ACM International Conference on Multimodal Interaction},
pp. \bfpage{598}--\blpage{602}
(\byear{2017})
\end{bchapter}
\endbibitem

\bibitem[\protect\citeauthoryear{Njoku et~al.}{2022}]{njoku2022deep}
\begin{barticle}
\bauthor{\bsnm{Njoku}, \binits{J.N.}},
\bauthor{\bsnm{Caliwag}, \binits{A.C.}},
\bauthor{\bsnm{Lim}, \binits{W.}},
\bauthor{\bsnm{Kim}, \binits{S.}},
\bauthor{\bsnm{Hwang}, \binits{H.}},
\bauthor{\bsnm{Jung}, \binits{J.}}:
\batitle{Deep learning based data fusion methods for multimodal emotion recognition}.
\bjtitle{The Journal of Korean Institute of Communications and Information Sciences}
\bvolume{47}(\bissue{1}),
\bfpage{79}--\blpage{87}
(\byear{2022})
\end{barticle}
\endbibitem

\bibitem[\protect\citeauthoryear{Livingstone and Russo}{2018}]{livingstone2018ryerson}
\begin{barticle}
\bauthor{\bsnm{Livingstone}, \binits{S.R.}},
\bauthor{\bsnm{Russo}, \binits{F.A.}}:
\batitle{The ryerson audio-visual database of emotional speech and song (ravdess): A dynamic, multimodal set of facial and vocal expressions in north american english}.
\bjtitle{PloS one}
\bvolume{13}(\bissue{5}),
\bfpage{0196391}
(\byear{2018})
\end{barticle}
\endbibitem

\bibitem[\protect\citeauthoryear{Sun et~al.}{2021}]{sun2021multimodal}
\begin{bchapter}
\bauthor{\bsnm{Sun}, \binits{L.}},
\bauthor{\bsnm{Xu}, \binits{M.}},
\bauthor{\bsnm{Lian}, \binits{Z.}},
\bauthor{\bsnm{Liu}, \binits{B.}},
\bauthor{\bsnm{Tao}, \binits{J.}},
\bauthor{\bsnm{Wang}, \binits{M.}},
\bauthor{\bsnm{Cheng}, \binits{Y.}}:
\bctitle{Multimodal emotion recognition and sentiment analysis via attention enhanced recurrent model}.
In: \bbtitle{Proceedings of the 2nd on Multimodal Sentiment Analysis Challenge},
pp. \bfpage{15}--\blpage{20}
(\byear{2021})
\end{bchapter}
\endbibitem

\bibitem[\protect\citeauthoryear{Stappen et~al.}{2021}]{stappen2021multimodal}
\begin{botherref}
\oauthor{\bsnm{Stappen}, \binits{L.}},
\oauthor{\bsnm{Baird}, \binits{A.}},
\oauthor{\bsnm{Schumann}, \binits{L.}},
\oauthor{\bsnm{Bjorn}, \binits{S.}}:
The multimodal sentiment analysis in car reviews (muse-car) dataset: Collection, insights and improvements.
IEEE Transactions on Affective Computing
(2021)
\end{botherref}
\endbibitem

\bibitem[\protect\citeauthoryear{Franceschini et~al.}{2022}]{franceschini2022multimodal}
\begin{bchapter}
\bauthor{\bsnm{Franceschini}, \binits{R.}},
\bauthor{\bsnm{Fini}, \binits{E.}},
\bauthor{\bsnm{Beyan}, \binits{C.}},
\bauthor{\bsnm{Conti}, \binits{A.}},
\bauthor{\bsnm{Arrigoni}, \binits{F.}},
\bauthor{\bsnm{Ricci}, \binits{E.}}:
\bctitle{Multimodal emotion recognition with modality-pairwise unsupervised contrastive loss}.
In: \bbtitle{2022 26th International Conference on Pattern Recognition (ICPR)},
pp. \bfpage{2589}--\blpage{2596}
(\byear{2022}).
\bcomment{IEEE}
\end{bchapter}
\endbibitem

\bibitem[\protect\citeauthoryear{Mocanu and Tapu}{2022}]{mocanu2022audio}
\begin{bchapter}
\bauthor{\bsnm{Mocanu}, \binits{B.}},
\bauthor{\bsnm{Tapu}, \binits{R.}}:
\bctitle{Audio-video fusion with double attention for multimodal emotion recognition}.
In: \bbtitle{2022 IEEE 14th Image, Video, and Multidimensional Signal Processing Workshop (IVMSP)},
pp. \bfpage{1}--\blpage{5}
(\byear{2022}).
\bcomment{IEEE}
\end{bchapter}
\endbibitem

\bibitem[\protect\citeauthoryear{Ren et~al.}{2021}]{ren2021interactive}
\begin{barticle}
\bauthor{\bsnm{Ren}, \binits{M.}},
\bauthor{\bsnm{Huang}, \binits{X.}},
\bauthor{\bsnm{Shi}, \binits{X.}},
\bauthor{\bsnm{Nie}, \binits{W.}}:
\batitle{Interactive multimodal attention network for emotion recognition in conversation}.
\bjtitle{IEEE Signal Processing Letters}
\bvolume{28},
\bfpage{1046}--\blpage{1050}
(\byear{2021})
\end{barticle}
\endbibitem

\bibitem[\protect\citeauthoryear{Zhang et~al.}{2019}]{zhang2019deep}
\begin{bchapter}
\bauthor{\bsnm{Zhang}, \binits{Y.}},
\bauthor{\bsnm{Wang}, \binits{Z.-R.}},
\bauthor{\bsnm{Du}, \binits{J.}}:
\bctitle{Deep fusion: An attention guided factorized bilinear pooling for audio-video emotion recognition}.
In: \bbtitle{2019 International Joint Conference on Neural Networks (IJCNN)},
pp. \bfpage{1}--\blpage{8}
(\byear{2019}).
\bcomment{IEEE}
\end{bchapter}
\endbibitem

\bibitem[\protect\citeauthoryear{Qi et~al.}{2021}]{qi2021feature}
\begin{barticle}
\bauthor{\bsnm{Qi}, \binits{Q.}},
\bauthor{\bsnm{Lin}, \binits{L.}},
\bauthor{\bsnm{Zhang}, \binits{R.}}:
\batitle{Feature extraction network with attention mechanism for data enhancement and recombination fusion for multimodal sentiment analysis}.
\bjtitle{Information}
\bvolume{12}(\bissue{9}),
\bfpage{342}
(\byear{2021})
\end{barticle}
\endbibitem

\bibitem[\protect\citeauthoryear{Zhuang et~al.}{2022}]{zhuang2022transformer}
\begin{barticle}
\bauthor{\bsnm{Zhuang}, \binits{X.}},
\bauthor{\bsnm{Liu}, \binits{F.}},
\bauthor{\bsnm{Hou}, \binits{J.}},
\bauthor{\bsnm{Hao}, \binits{J.}},
\bauthor{\bsnm{Cai}, \binits{X.}}:
\batitle{Transformer-based interactive multi-modal attention network for video sentiment detection}.
\bjtitle{Neural Processing Letters}
\bvolume{54}(\bissue{3}),
\bfpage{1943}--\blpage{1960}
(\byear{2022})
\end{barticle}
\endbibitem

\bibitem[\protect\citeauthoryear{Guo et~al.}{2022}]{guo2022er}
\begin{bchapter}
\bauthor{\bsnm{Guo}, \binits{X.}},
\bauthor{\bsnm{Wang}, \binits{Y.}},
\bauthor{\bsnm{Miao}, \binits{Z.}},
\bauthor{\bsnm{Yang}, \binits{X.}},
\bauthor{\bsnm{Guo}, \binits{J.}},
\bauthor{\bsnm{Hou}, \binits{X.}},
\bauthor{\bsnm{Zao}, \binits{F.}}:
\bctitle{Er-mrl: Emotion recognition based on multimodal representation learning}.
In: \bbtitle{2022 12th International Conference on Information Science and Technology (ICIST)},
pp. \bfpage{421}--\blpage{428}
(\byear{2022}).
\bcomment{IEEE}
\end{bchapter}
\endbibitem

\bibitem[\protect\citeauthoryear{Zhang et~al.}{2022}]{zhang2022transformer}
\begin{bchapter}
\bauthor{\bsnm{Zhang}, \binits{W.}},
\bauthor{\bsnm{Qiu}, \binits{F.}},
\bauthor{\bsnm{Wang}, \binits{S.}},
\bauthor{\bsnm{Zeng}, \binits{H.}},
\bauthor{\bsnm{Zhang}, \binits{Z.}},
\bauthor{\bsnm{An}, \binits{R.}},
\bauthor{\bsnm{Ma}, \binits{B.}},
\bauthor{\bsnm{Ding}, \binits{Y.}}:
\bctitle{Transformer-based multimodal information fusion for facial expression analysis}.
In: \bbtitle{Proceedings of the IEEE/CVF Conference on Computer Vision and Pattern Recognition},
pp. \bfpage{2428}--\blpage{2437}
(\byear{2022})
\end{bchapter}
\endbibitem

\bibitem[\protect\citeauthoryear{Lv et~al.}{2021}]{lv2021progressive}
\begin{bchapter}
\bauthor{\bsnm{Lv}, \binits{F.}},
\bauthor{\bsnm{Chen}, \binits{X.}},
\bauthor{\bsnm{Huang}, \binits{Y.}},
\bauthor{\bsnm{Duan}, \binits{L.}},
\bauthor{\bsnm{Lin}, \binits{G.}}:
\bctitle{Progressive modality reinforcement for human multimodal emotion recognition from unaligned multimodal sequences}.
In: \bbtitle{Proceedings of the IEEE/CVF Conference on Computer Vision and Pattern Recognition},
pp. \bfpage{2554}--\blpage{2562}
(\byear{2021})
\end{bchapter}
\endbibitem

\bibitem[\protect\citeauthoryear{Shen et~al.}{2021}]{shen2021mmtrans}
\begin{bchapter}
\bauthor{\bsnm{Shen}, \binits{J.}},
\bauthor{\bsnm{Zheng}, \binits{J.}},
\bauthor{\bsnm{Wang}, \binits{X.}}:
\bctitle{Mmtrans-mt: A framework for multimodal emotion recognition using multitask learning}.
In: \bbtitle{2021 13th International Conference on Advanced Computational Intelligence (ICACI)},
pp. \bfpage{52}--\blpage{59}
(\byear{2021}).
\bcomment{IEEE}
\end{bchapter}
\endbibitem

\bibitem[\protect\citeauthoryear{Chaudhari et~al.}{2023}]{chaudhari2023facial}
\begin{barticle}
\bauthor{\bsnm{Chaudhari}, \binits{A.}},
\bauthor{\bsnm{Bhatt}, \binits{C.}},
\bauthor{\bsnm{Krishna}, \binits{A.}},
\bauthor{\bsnm{Travieso-Gonz{\'a}lez}, \binits{C.M.}}:
\batitle{Facial emotion recognition with inter-modality-attention-transformer-based self-supervised learning}.
\bjtitle{Electronics}
\bvolume{12}(\bissue{2}),
\bfpage{288}
(\byear{2023})
\end{barticle}
\endbibitem

\bibitem[\protect\citeauthoryear{Guo et~al.}{2017}]{guo2017calibration}
\begin{bchapter}
\bauthor{\bsnm{Guo}, \binits{C.}},
\bauthor{\bsnm{Pleiss}, \binits{G.}},
\bauthor{\bsnm{Sun}, \binits{Y.}},
\bauthor{\bsnm{Weinberger}, \binits{K.Q.}}:
\bctitle{On calibration of modern neural networks}.
In: \bbtitle{International Conference on Machine Learning},
pp. \bfpage{1321}--\blpage{1330}
(\byear{2017}).
\bcomment{PMLR}
\end{bchapter}
\endbibitem

\bibitem[\protect\citeauthoryear{Thushara and Veni}{2016}]{thushara2016multimodal}
\begin{bchapter}
\bauthor{\bsnm{Thushara}, \binits{S.}},
\bauthor{\bsnm{Veni}, \binits{S.}}:
\bctitle{A multimodal emotion recognition system from video}.
In: \bbtitle{2016 International Conference on Circuit, Power and Computing Technologies (ICCPCT)},
pp. \bfpage{1}--\blpage{5}
(\byear{2016}).
\bcomment{IEEE}
\end{bchapter}
\endbibitem

\bibitem[\protect\citeauthoryear{Amali et~al.}{2018}]{amali2018semantic}
\begin{bchapter}
\bauthor{\bsnm{Amali}, \binits{D.N.}},
\bauthor{\bsnm{Barakbah}, \binits{A.R.}},
\bauthor{\bsnm{Besari}, \binits{A.R.A.}},
\bauthor{\bsnm{Agata}, \binits{D.}}:
\bctitle{Semantic video recommendation system based on video viewers impression from emotion detection}.
In: \bbtitle{2018 International Electronics Symposium on Knowledge Creation and Intelligent Computing (ies-kcic)},
pp. \bfpage{176}--\blpage{183}
(\byear{2018}).
\bcomment{IEEE}
\end{bchapter}
\endbibitem

\bibitem[\protect\citeauthoryear{Garcia-Garcia et~al.}{2023}]{garcia2023building}
\begin{barticle}
\bauthor{\bsnm{Garcia-Garcia}, \binits{J.M.}},
\bauthor{\bsnm{Lozano}, \binits{M.D.}},
\bauthor{\bsnm{Penichet}, \binits{V.M.}},
\bauthor{\bsnm{Law}, \binits{E.L.-C.}}:
\batitle{Building a three-level multimodal emotion recognition framework}.
\bjtitle{Multimedia Tools and Applications}
\bvolume{82}(\bissue{1}),
\bfpage{239}--\blpage{269}
(\byear{2023})
\end{barticle}
\endbibitem

\end{thebibliography}

\end{document}